\documentclass[11pt]{article}

\usepackage[final]{acl}

\usepackage{times}
\usepackage{latexsym}

\usepackage[T1]{fontenc}

\usepackage[utf8]{inputenc}

\usepackage{microtype}

\usepackage{inconsolata}

\usepackage{graphicx}
\usepackage{placeins}
\usepackage{caption}

\usepackage{tcolorbox}
\tcbuselibrary{skins, breakable}
\newtcolorbox{promptbox}[2][]{
  enhanced,
  breakable,
  colback=gray!6,
  colframe=black!40,
  fonttitle=\bfseries\small,
  title={#2},
  left=4pt, right=4pt, top=4pt, bottom=4pt,
  #1
}
\usepackage[utf8]{inputenc}
\usepackage[T2A,T1]{fontenc}   
\usepackage[russian,english]{babel}
\usepackage{CJKutf8}

\definecolor{darkblue}{rgb}{0, 0, 0.5}
\hypersetup{colorlinks=true, citecolor=darkblue, linkcolor=darkblue, urlcolor=darkblue}

\usepackage{soul}
\usepackage{xcolor}
\newcommand{\ra}[1]{\renewcommand{\arraystretch}{#1}}

\newcommand{\txttodo}[1]{{\small \color{red!70!black}\textbf{[TODO: #1]}}}
\usepackage{graphicx}
\usepackage{makecell}
\usepackage{multirow}
\usepackage{arydshln}
\usepackage{booktabs}
\usepackage{amsmath, amssymb}
\usepackage[noabbrev,capitalize,nameinlink]{cleveref}
\usepackage{subcaption}

\usepackage{float}
\usepackage{stfloats}

\definecolor{darkgreen}{rgb}{0,0.5,0}
\definecolor{darkred}{rgb}{0.7,0,0}

\definecolor{deltagray}{gray}{0.45}

\newcommand{\github}{\raisebox{-0.25cm}{\includegraphics[width=0.7cm]{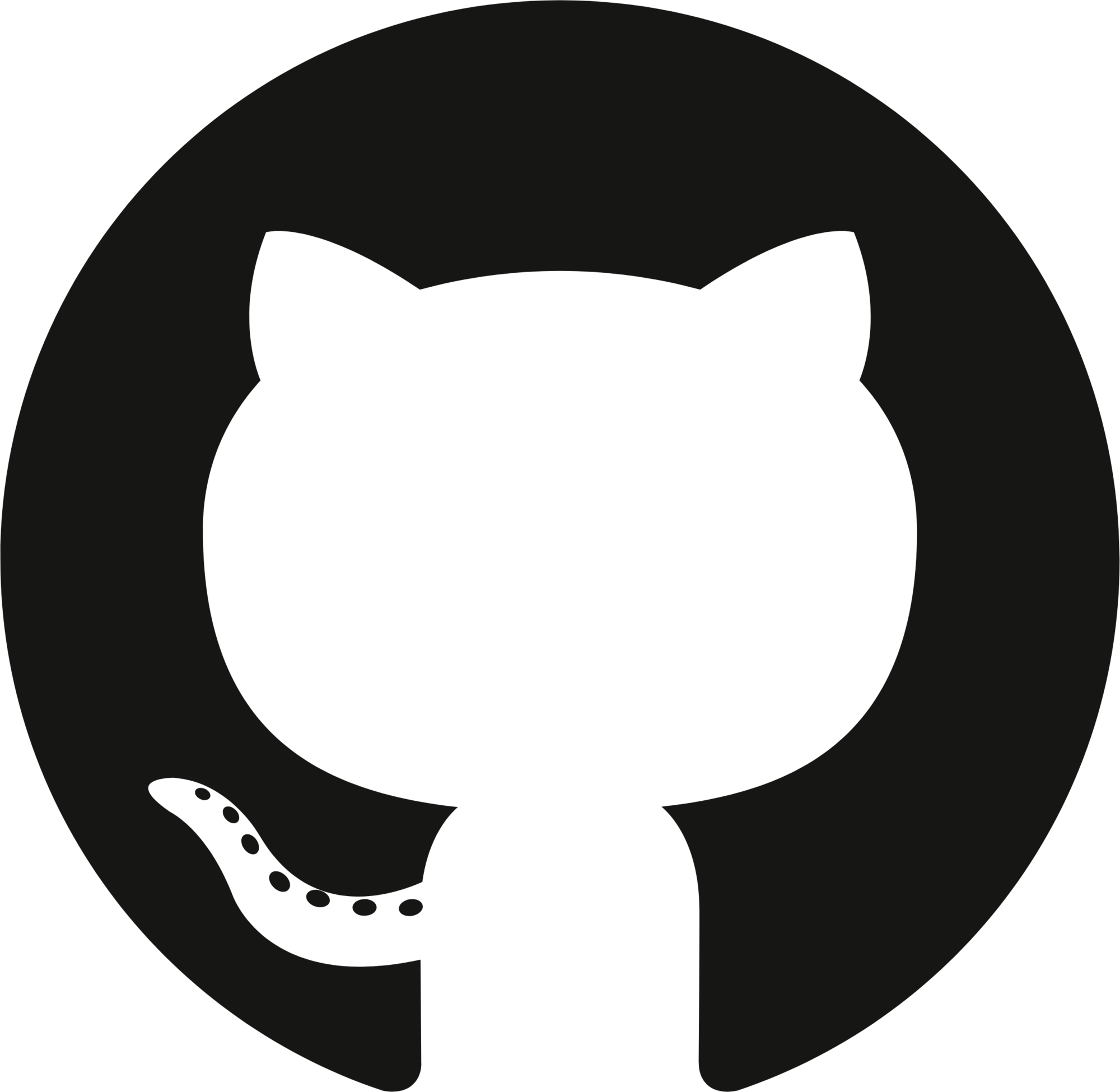}}}

\usepackage{url}

\makeatletter \g@addto@macro\UrlSpecials{\do\!{\newline}}
\makeatother

\PassOptionsToPackage{hyphens}{url}
\PassOptionsToPackage{breaklinks}{hyperref}
\mathchardef\UrlBreakPenalty=100
\mathchardef\UrlBigBreakPenalty=100
\usepackage{breakurl}

\usepackage{fontawesome5}

\title{Shared Doubt: Zero-Shot Cross-Lingual Confidence Estimation for Language Models}

\author{Athina Kyriakou\textsuperscript{\faChessRook },\hspace{.5em}  Dennis Ulmer\textsuperscript{\faBicycle},\hspace{.5em}  Ivan Titov\textsuperscript{\faBicycle, \faChessRook }\\
      \textsuperscript{\faBicycle } ILLC, University of Amsterdam \\
       \textsuperscript{\faChessRook} ILCC, University of Edinburgh \\
       \texttt{athina.skyriakou@gmail.com},
       \texttt{dennis.ulmer@mailbox.org},
       \texttt{ititov@inf.ed.ac.uk}}

\begin{document}
\maketitle

\begin{abstract}

Confidence estimation (CE), i.e.\@ quantifying the reliability of a model's prediction, has attracted great interest in the context of large language models (LLMs). 
However, most studies focus on English, ignoring the multilingual reality of LLM usage,
while many CE methods degrade or require retraining across languages. 
To address this gap, we investigate whether multilingual LLMs encode shared, language-transferable confidence features in open-ended question answering. 
We use a lightweight linear probe that predicts answer correctness directly from intermediate representations. 
Trained monolingually, the probe generalizes zero-shot to \emph{unseen}, linguistically diverse languages without target-language supervision.
Multiple ablations and learned layer weights reveal that confidence features concentrate in middle layers across languages, suggesting a shared confidence subspace. 
While zero-shot cross-lingual performance depends on similarity to the source language, the probe provides a strong baseline without any retraining and compares favorably to other popular confidence estimation methods.

    \vspace{0.2cm}
    \begin{tabular}{rl}
        {\github} & \href{https://github.com/AthinaKyriakou/shared-doubt}{\texttt{AthinaKyriakou/shared-doubt}} \\
    \end{tabular}
    \vspace{0.2cm}
\end{abstract}

\section{Introduction}\label{sec:introduction}


Large language models (LLMs) have seen unprecedented adoption in recent years \citep{wang2024understanding, humlum2025large}, including in language communities besides English \citep{liu2026earth}. 
Despite both public awareness and academic research on their limitations, \emph{hallucination}, the generation of plausible but factually incorrect statements (see e.g.\@ \citealp{huang2025survey} for an overview), still poses an ongoing problem that risks downstream harms and the loss of user trust.
One potential solution to this problem is confidence estimation (CE), i.e.\@ obtaining numerical values which reflect the probability that the LLM's output is correct. CE has long been studied in machine learning \citep{degroot1983comparison, naeini2015obtaining, guo2017calibration, ovadia2019can, minderer2021revisiting} and has received increasing attention for LLMs \citep{jiang2021can, kadavath2022language, tian2023just, zhu2023calibration, ulmer2024calibrating, kapoor2024calibration}. 
However, the vast majority of works focus exclusively on English, leaving other languages unexplored. 
This is particularly concerning as LLMs may behave less reliably in languages with less training data, making accurate confidence estimation paramount \citep{ahuja2022calibration, ulmer2022exploring, krause2023confidently}.


\begin{figure}[tb!]
    \centering
    \includegraphics[width=\linewidth]{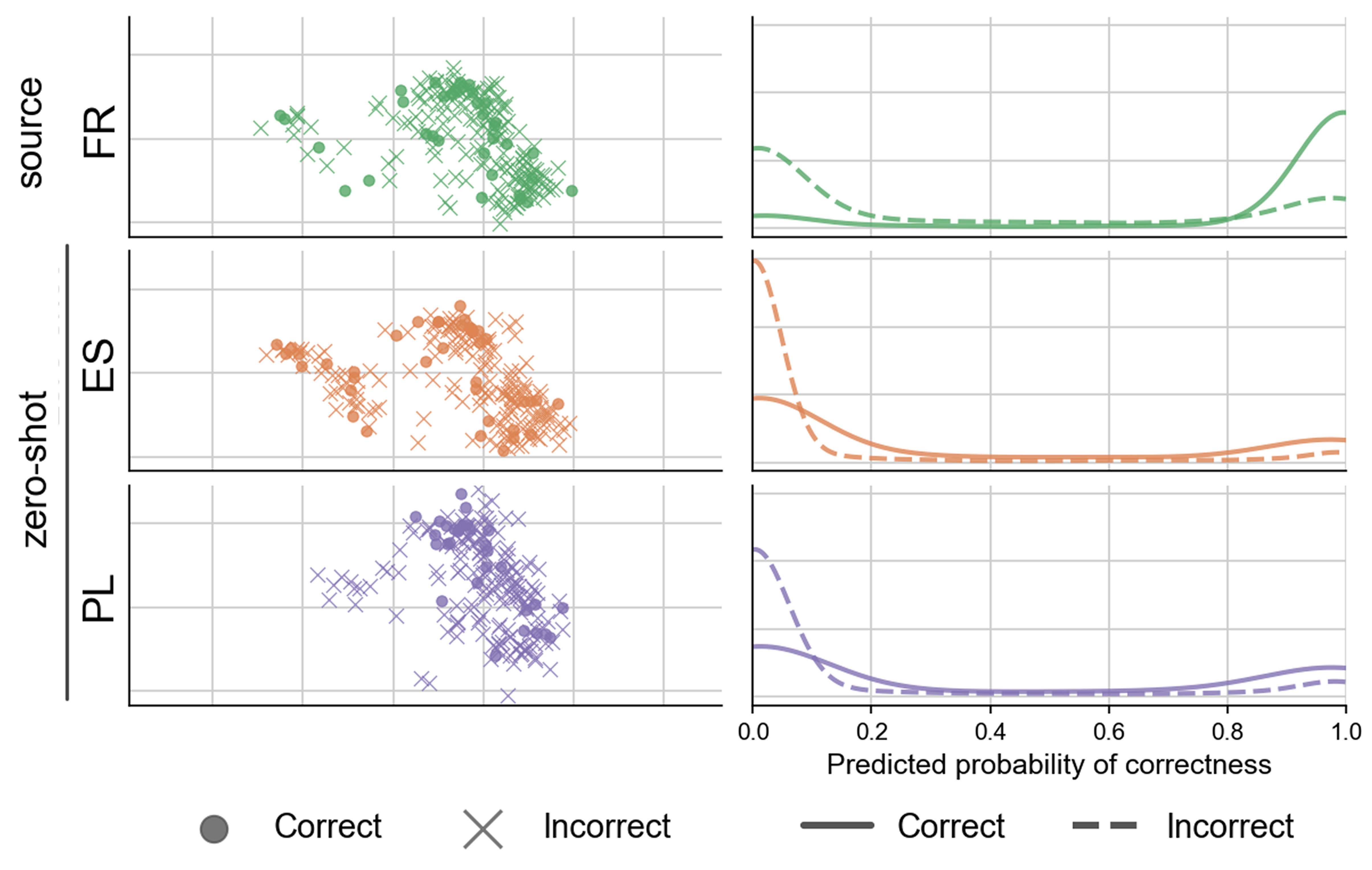} 
    \caption{(Left) PCA scatter plots of (in-)correct samples in the probe's learned aggregation of the LLM latent space. (Right) Distribution of confidence scores.}\label{fig:figure1}
\end{figure}

In this work, we take a first step toward truly multilingual confidence estimation in LLMs, focusing on short-form open-ended question answering (QA).
Our approach is motivated by a key observation: prior work on multilinguality has shown that LLMs' hidden states capture cross-lingual features \citep{wang-2024-probing, wendler-2024, dumas-2024-separating, schut-2025}, while e.g.\@ \citet{marks-2023, ji2025calibrating} have identified uncertainty-related features. 
We hypothesize that these properties jointly enable cross-lingual transfer of confidence estimates. 
To test this, we apply a lightweight linear probe that learns a weighted combination of hidden layers to predict answer correctness. We show that the probe, trained on an LLM's latent representations in a single language, can predict confidence scores for unseen languages from different genera and scripts, without any target-language supervision (\emph{zero-shot}).
\cref{fig:figure1} illustrates how correct and incorrect responses are distributed similarly in the learned aggregation of the target LLM's latent space. 
The probe can exploit this fact, as shown by its predictive distribution on the right, with some degradation as the zero-shot language becomes more dissimilar.

Our contributions are as follows. To the best of our knowledge, we are the first to demonstrate zero-shot cross-lingual transfer of confidence estimation in open-ended QA through probing of intermediate LLM representations. 
We analyze how performance varies across unseen languages of the same and different language families and scripts, and show that the probe outperforms baselines from several CE method families \emph{without retraining} on the target language. 
Through ablations, we find that the most salient information for zero-shot confidence prediction is localized in the middle layers of the model, consistent with prior work suggesting that these layers encode a shared cross-lingual representational space. 
More broadly, our findings suggest a possible path toward multilingual calibration with little or no target-language supervision, where confidence signals learned in higher-resource languages transfer to lower-resource settings.
We release all code as open source.

\section{Related Work}\label{sec:related-work}

We position our work at the intersection of studies examining the latent space of multilingual LLMs, the probing of model representations to identify whether specific properties are encoded, and the wider landscape of confidence estimation in LLMs. 

\paragraph{Multilingual Latent Spaces.}


Several studies have examined the organization of LLMs’ latent space, finding that layers closest to the input and output are more language-specific. 
However, there is no consensus on whether multilingual processing involves mapping to a particular language or through shared, language-agnostic regions in the representation space.  
Some works support the language dominance hypothesis, according to which inputs are mapped into a representation space shaped by the most prevalent training language (usually English), perform computations in that space, and translate the result back into the original language \citep{zhang-2023-dont, wendler-2024, schut-2025, zhao-2024, tezuka-2025}. 
\citet{zhao-2024} show that the middle layers reason in English while leveraging multilingual knowledge, while \citet{tezuka-2025} identify two types of neurons,
namely ones responsible for mapping from a language-specific space to a shared semantic space in the early and middle layers, and those operating in reverse in the upper layers.
\citet{schut-2025} find that semantically meaningful words pass through an English-centric representation space, and \citet{saji-2025} extend these findings to writing systems, suggesting that middle layers represent tokens in Latin script. 
By contrast, \citet{zhang2024unveiling, wang-2024-sharing} indicate that multilinguality is facilitated by shared language-agnostic representations, identifying parameters essential to cross-lingual performance. 
\citet{bhattacharya2023unveiling, tang2024language, kojima2024multilingual, zeng-2025} discover language-specific neurons primarily in the lower and upper layers, with a gap in the middle layers. \citet{dumas-2024-separating} further show that the output language is encoded in earlier layers, whereas the concept to be translated emerges in later layers. 
\citet{shani-2025} find no consistently dominant language in the middle layers.


\paragraph{Probing Latent Representations.}

Probes are lightweight models that map a model’s internal representations to a property of interest \cite{alain-2017, belinkov-2022}, and are widely used in mechanistic interpretability to test whether a property is encoded in LLM representations \cite{bereska-2024, ferrando-2024}.
In multilingual LLM research, probes have been used to study morphosyntax \cite{brinkmann-2025, wang-2024-probing}, factual knowledge \cite{li-2025}, semantics \cite{li-2025, wang-2024-probing}, and context \cite{ju-2024}, with shared features typically emerging in the middle layers. 
Relevant to our work, probes have been trained on LLM representations in English to detect truthfulness \citep{li-2023, marks-2023, he-2024, azaria-2023}, with \citet{li-2025} finding stronger signals in high-resource than in low-resource languages. 
While \citet{li-2023, marks-2023} find that truthfulness signals peak in the middle layers, \citet{jin-2025} show that more complex properties require deeper layers. 

\paragraph{Confidence Estimation in LLMs.}
As mentioned in the introduction, confidence estimation (CE), i.e.\@ obtaining a score that reflects the probability that a model's answer is correct, has recently attracted a flurry of new work (see e.g.\@ \citealp{baan2023uncertainty, huang2024survey, ulmer2024uncertainty, liu2025uncertainty, geng2024survey, shorinwa2025survey}).
Since some methods rely on the generation process itself, for example token logits or additional model responses, they are immediately applicable to any language, albeit without any guarantee of performance.
Other methods, by contrast, are trained on a specific language. Applying them to unseen languages therefore constitutes a zero-shot prediction problem. Probing approaches, which have also been applied to CE \citep{kadavath2022language, marks-2023, burns2023discovering, liu2024uncertainty, xiong2024efficient, liu2024litcablightweightlanguagemodel, fathullah2024needs, shelmanov2025head}, fall into this setting.
Importantly, however, few works test performance beyond English, especially on lower-resourced languages. \citet{krause2023confidently} show that the calibration of verbalized confidence scores varies greatly across languages, with \citet{yang2023calibration} showing the same for sequence-likelihood-based confidence. \citet{xue2025mlingconf} introduce a multilingual CE benchmark and also report wide performance differences across existing methods. In all three, however, confidence is read from model outputs rather than internal representations, and any improvement relies on target-language data or language-specific prompting. None demonstrates zero-shot prediction. Closer to our approach, \citet{zhou2025beyond} locate the most salient multilingual CE features in the middle layers, but their methods require target-language labels and are defined only for multiple-choice QA. We therefore treat their findings as complementary. Our approach extends to the zero-shot, open-ended QA setting, and our ablations corroborate the middle-layer concentration of multilingual confidence signal.


\section{Methodology}\label{sec:probe-training}

Based on the insights from the literature, we now discuss which representations to consider, verify our assumptions, and define the probe.

\subsection{Selected Intermediate Representations}

Our work focuses on decoder-only transformer-based LLMs  \cite{vaswani-2017, radford2019language}. 
Given an input token sequence, the model embeds the token at every time step $t$ into a high-dimensional space $\mathbf{x}_t \in \mathbb{R}^D$ and produces a hidden state $\mathbf{h}_t^{(\ell)}$ by running sequentially through each transformer block at layer $\ell$, before finally generating sequence $y_1, y_2, ..., y_T$. 
Each block updates the previous hidden states in a residual way:
\begin{equation}\label{eq:residual-stream}
    \mathbf{h}_t^{(\ell)} = \mathbf{h}_t^{(\ell-1)} + g_\ell\Big(\mathbf{h}_1^{(\ell-1)}, \dots, \mathbf{h}_{t}^{(\ell-1)}\Big),
\end{equation}
\noindent where $g_\ell$ is the model-specific transformation of the $\ell$-th block applied to the prefix and current token.
This is also referred to as the residual stream, serving as a central pathway that carries and incrementally refines each token’s representation. 

Following \citet{zhou2025beyond}, we exploit signals from all layers but focus on the hidden states at two time steps: the prompt's last token (\emph{last query token}) and the generated answer's last token (\emph{last answer token}).
The former serves as a proxy for the entire input, integrating information from all preceding tokens and capturing the model's confidence at the start of generation. This token is widely used in multilingual interpretability and uncertainty studies
\citep{wendler-2024, dumas-2024-separating, wang-2024-sharing, brinkmann-2025, li-2025, he-2024, duan-2024, ji-2024}.
The latter represents the model's state after generation and may encode its confidence in the completed response. In other contexts it has been shown to encode knowledge \citep{burns2023discovering} or truthfulness \citep{azaria-2023}.

\subsection{A Cross-lingual Confidence Probe}\label{sec:confidence-probe}

\begin{figure*}[htb]
    \centering
    \includegraphics[width=\textwidth]{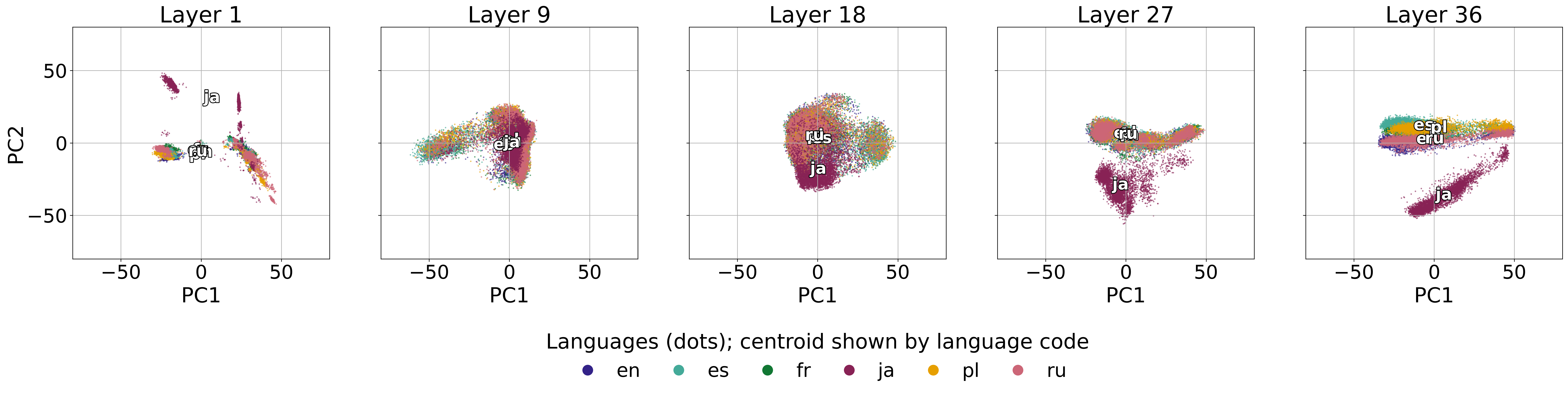}
    \caption{Organization of latent representations in Qwen3 8B on Global-MMLU, using the last query token.}\label{fig:multilingual-pca}
\end{figure*}

Prior work has identified considerable overlap of latent representations for different language representations in the middle layers \citep{wendler-2024, wang-2024-probing, zhou2025beyond, zhang2024unveiling}.
We confirm this in our case, showing a selection of PCA plots in \cref{fig:multilingual-pca}.
Since that implies that some layers will carry more generalizable information, we define a probe that learns a weighted average of layer hidden representations $\mathbf{h}_t^{(\ell)}$ by introducing a learnable parameter $\mathbf{w} \in \mathbb{R}^L$, using it to compute layer weights $\alpha_l$ and pooled representation $\bar{\mathbf{h}}_t \in \mathbb{R}^D$:

\begin{equation} \label{eq:layer_imp}
    \alpha_\ell = \mathrm{softmax}(\mathbf{w})_\ell = \frac{\exp(w_\ell)}{\sum_{k=1}^L \exp(w_k)},
\end{equation}
\begin{equation} \label{eq:repr_pooling}
    \bar{\mathbf{h}}_t = \sum_{\ell=1}^L \alpha_\ell \, \mathbf{h}_{t}^{(\ell)}.
\end{equation}
This representation allows the probe to select the most informative layers for confidence estimation during training. It further assumes that each dimension is approximately
consistent across layers, a side effect of \cref{eq:residual-stream} as information propagates through layers.
Finally, a linear layer with learnable parameters $\mathbf{v} \in \mathbb{R}^D, b \in \mathbb{R}$ and the sigmoid function $\sigma(\cdot)$ are applied to obtain a confidence score $\hat{p}_t$:
\begin{equation} \label{eq:logit}
    \hat{p}_t = \sigma\big(\mathbf{v}^\top \mathbf{\bar{h}}_t + b\big).
\end{equation}

While non-linear designs of the probe are possible, evidence from the linear representation hypothesis \cite{bereska-2024} and other studies \citep{park2023linear, jiang2024origins, ji2025calibrating, merullo2025linear}, suggests that features are often represented linearly in latent space.

\section{Experiments} 



\subsection{Setup}


\paragraph{Datasets.} 
We base our experiments on two popular parallel question-answering datasets that enable comparable evaluation across languages. The first is MKQA \cite{mkqa}, which provides open-ended questions across 26 languages, with the original English questions extracted from Natural Questions \cite{natural-questions}. We consider short-phrase, entity, numeric, numeric-with-unit, and date question types and exclude time-sensitive questions, retaining at most 5237 per language. Further details of the selection process are in \cref{sec:data_pre_appendix_mkqa}. The second is Global-MMLU \cite{globalmmlu}, a multilingual multiple-choice dataset covering 42 languages, derived from the English MMLU dataset \cite{hendrycks-2021}. Since multiple-choice evaluation is discriminative and not representative of real-world generative scenarios \cite{chandak-2025}, we keep only questions that can be posed in an open-ended manner, following the process detailed in \cref{sec:data_pre_appendix_gmmlu}. This yields a final set of 4215 questions per language.


\begin{table}[tb]
\centering
\renewcommand{\arraystretch}{1.2}
\resizebox{0.48\textwidth}{!}{ 
\begin{tabular}{lrrrr}
\hline
\textbf{Language} & \textbf{Genus} & \textbf{Script} & \makecell[tr]{\textbf{MKQA}\\\textbf{translation}} & \makecell[tr]{\textbf{Global-MMLU}\\\textbf{translation}} \\\hline
English & Germanic & Latin & N/A & N/A \\
French & Romance & Latin & Human & Human \\
Spanish & Romance & Latin & Human & Human \\
Polish & Slavic & Latin & Human & Google Translate \\
Russian & Slavic & Cyrillic & Human & \makecell[tr]{Human \& \\ Google Translate} \\
Japanese & Japanese & Japanese & Human & Human \\
\hline
\end{tabular}
}
\caption{Languages and translation types by dataset.}
\label{tab:language_types}
\end{table}

\paragraph{Selected Languages.}
To capture cross-lingual variation, we select the six languages shown in \cref{tab:language_types}: English (en) is included as it dominates most LLMs' training data. We further include French (fr) and Spanish (es), two closely related languages that belong to the Romance genus \cite{wals} and use the Latin script.
Polish (pl) and Russian (ru) are chosen to represent the more dissimilar Slavic languages \cite{wals}, with Russian additionally introducing the Cyrillic script.
Finally, Japanese (ja) differs in both linguistic family and script. We choose French rather than English as the source language for training the probe: this lets us test on Spanish as a near-out-of-distribution case and on English as a different-genus, training-dominant one, alongside the other languages we consider. \footnote{Our main results for all source-target language pairs are reported in \cref{app:conf-est-all-sources}.}

\paragraph{Models.} 
We test our approach on two widely used multilingual auto-regressive LLMs, Llama 3.1 8B Instruct \citep{grattafiori2024llama} (32 layers) and Qwen3 8B \citep{qwen3} (36 layers), the latter run in non-thinking mode for direct short-form answers. We generate with the recommended temperatures (0.6 for Llama, 0.7 for Qwen) and a maximum of 40 new tokens. The per-dataset and per-language system prompts are given in \cref{fig:prompts} in \cref{sec:llm_parameters_appendix}. All experiments were run on a single NVIDIA A100-80G-PCIe GPU.


\begin{table}[tb] 
\centering
\renewcommand{\arraystretch}{1.4}
\resizebox{0.485\textwidth}{!}{
    \begin{tabular}{@{}llcccccc@{}}
    \toprule
    Dataset & LLM & fr & en & es & pl & ru & ja \\
    \midrule
    \multirow{2}{*}{MKQA}
    & \makecell[tl]{Llama 3.1 8B} & .27 & .38 & .24 & .26 & .17 & .14 \\
    & Qwen3 8B      & .18 & .26 & .18 & .16 & .13 & .12 \\
    \hline
    \multirow{2}{*}{Global-MMLU}
    & \makecell[tl]{Llama 3.1 8B} & .24 & .35 & .24 & .19 & .19 & .17  \\
    & Qwen3 8B      & .27 & .40 & .26 & .23 & .22 & .22 \\
    \bottomrule
    \end{tabular}
}
\caption{LLM accuracy per dataset and language.}
\label{tab:llm_performance}
\end{table}

\paragraph{Answer Correctness.}

We assess answer correctness with GPT-4.1 models as judges \citep{openai2025gpt41}, prompted as shown in \cref{fig:gpt4.1-prompt-corr-sys}. To validate these labels, two native speakers per language independently annotated 50 generations sampled from both datasets. Inter-annotator agreement is high across languages (Cohen's $\kappa = .80$-$.92$) and the judge matches the resulting gold labels on 90.3\% of items, with per-language $\kappa = .75$-$.83$. An exception is Japanese ($.65$), where errors are almost exclusively false negatives (see \cref{tab:judge_validation}). 
We obtain the accuracies in \cref{tab:llm_performance}, with two patterns being notable: accuracies differ across languages, reflecting differences in factual consistency \citep{fierro2022factual, qi2023cross}.
Performance is the  strongest in English, declining across the remaining languages with their relatedness to English and their presumed share of pretraining data. Further detail is given in \cref{sec:llm_parameters_appendix}.



\paragraph{Evaluation Metrics.}

We evaluate confidence estimates with standard metrics. For discrimination, i.e.\@ how well they separate correct from incorrect responses, we report the area under the receiver operating characteristic (AUROC) and the precision-recall curve (AUPR). For calibration, i.e.\@ how well confidence scores match the probability of correctness, we report expected calibration error with $10$ bins (ECE; \citealp{naeini2015obtaining, guo2017calibration}) and the Brier score \citep{glenn1950verification}.

\subsection{Zero-Shot Confidence Estimation Results}\label{sec:zero-shot-ce}

We first evaluate the probe's zero-shot CE performance in
isolation.\footnote{Details on probe training are given in
\cref{app:training}.} We include a majority-class baseline
(\emph{majority}), which in our case is the negative class, and a
\emph{prior-probability} baseline, which draws binary correctness labels from a Bernoulli distribution with parameter $p$ set to the training accuracy. We also train a probe only on examples from the target language (\emph{oracle} probe).


\begin{table*}[tb]
\centering
\ra{1.3}
\renewcommand{\arraystretch}{1.5}
\setlength{\tabcolsep}{4pt}
\footnotesize
\resizebox{\textwidth}{!}{%
\begin{tabular}{llcccccccccc}
\toprule
 & & & \multicolumn{4}{c}{\textbf{MKQA}} & & \multicolumn{4}{c}{\textbf{Global-MMLU}} \\
\cmidrule(lr){4-7} \cmidrule(lr){9-12}
 & & & AUROC$\uparrow$ & AUPR$\uparrow$ & Brier$\downarrow$ & ECE$\downarrow$ & & AUROC$\uparrow$ & AUPR$\uparrow$ & Brier$\downarrow$ & ECE$\downarrow$ \\
\midrule
\multirow{2}{*}{\rotatebox{90}{all*}} & Majority & & .50$\pm$.00 & .58$\pm$.02 & .17$\pm$.05 & .17$\pm$.05 &  & .50$\pm$.00 & .63$\pm$.03 & .26$\pm$.06 & .26$\pm$.06 \\
 & Prior prob. & & .50$\pm$.00 & .58$\pm$.02 & .14$\pm$.03 & .04$\pm$.03 &  & .50$\pm$.00 & .63$\pm$.03 & .19$\pm$.03 & .06$\pm$.03 \\
\multirow{2}{*}{\rotatebox{90}{fr}} & Last Query & & .73$\pm$.01 & .36$\pm$.01 & .17$\pm$.01 & .14$\pm$.00 &  & .71$\pm$.01 & .49$\pm$.02 & .21$\pm$.01 & .16$\pm$.01 \\
 & Last Answer & & .82$\pm$.01 & .49$\pm$.02 & .14$\pm$.01 & .12$\pm$.01 &  & .78$\pm$.00 & .58$\pm$.01 & .19$\pm$.00 & .16$\pm$.00 \\
\midrule
\multirow{2}{*}{\rotatebox{90}{en}} & Last Query & & .70$\pm$.01 \textcolor{gray}{+{.00}} & .47$\pm$.01 \textcolor{gray}{+{.03}} & .22$\pm$.01 \textcolor{gray}{-{.01}} & .18$\pm$.01 \textcolor{gray}{+{.00}} &  & .64$\pm$.02 \textcolor{gray}{-{.07}} & .51$\pm$.02 \textcolor{gray}{-{.09}} & .31$\pm$.01 \textcolor{gray}{+{.06}} & .26$\pm$.01 \textcolor{gray}{+{.07}} \\
 & Last Answer & & .70$\pm$.01 \textcolor{gray}{-{.08}} & .44$\pm$.01 \textcolor{gray}{-{.13}} & .24$\pm$.01 \textcolor{gray}{+{.06}} & .22$\pm$.01 \textcolor{gray}{+{.09}} &  & .72$\pm$.01 \textcolor{gray}{-{.04}} & .60$\pm$.01 \textcolor{gray}{-{.09}} & .27$\pm$.01 \textcolor{gray}{+{.04}} & .24$\pm$.01 \textcolor{gray}{+{.05}} \\
\cmidrule(lr){2-12}
\multirow{2}{*}{\rotatebox{90}{es}} & Last Query & & .71$\pm$.01 \textcolor{gray}{-{.01}} & .34$\pm$.01 \textcolor{gray}{-{.02}} & .17$\pm$.00 \textcolor{gray}{-{.01}} & .14$\pm$.01 \textcolor{gray}{-{.02}} &  & .68$\pm$.01 \textcolor{gray}{+{.00}} & .41$\pm$.02 \textcolor{gray}{-{.03}} & .27$\pm$.02 \textcolor{gray}{+{.05}} & .25$\pm$.03 \textcolor{gray}{+{.07}} \\
 & Last Answer & & .82$\pm$.01 \textcolor{gray}{-{.01}} & .53$\pm$.03 \textcolor{gray}{+{.01}} & .19$\pm$.02 \textcolor{gray}{+{.05}} & .20$\pm$.03 \textcolor{gray}{+{.09}} &  & .74$\pm$.01 \textcolor{gray}{-{.03}} & .52$\pm$.01 \textcolor{gray}{-{.03}} & .23$\pm$.01 \textcolor{gray}{+{.03}} & .21$\pm$.02 \textcolor{gray}{+{.04}} \\
\cmidrule(lr){2-12}
\multirow{2}{*}{\rotatebox{90}{pl}} & Last Query & & .67$\pm$.01 \textcolor{gray}{-{.06}} & .28$\pm$.01 \textcolor{gray}{-{.06}} & .15$\pm$.01 \textcolor{gray}{-{.01}} & .13$\pm$.01 \textcolor{gray}{-{.01}} &  & .73$\pm$.01 \textcolor{gray}{+{.01}} & .40$\pm$.01 \textcolor{gray}{-{.02}} & .20$\pm$.01 \textcolor{gray}{-{.01}} & .17$\pm$.02 \textcolor{gray}{+{.00}} \\
 & Last Answer & & .73$\pm$.01 \textcolor{gray}{-{.09}} & .35$\pm$.03 \textcolor{gray}{-{.13}} & .23$\pm$.03 \textcolor{gray}{+{.10}} & .24$\pm$.03 \textcolor{gray}{+{.13}} &  & .75$\pm$.01 \textcolor{gray}{-{.04}} & .48$\pm$.01 \textcolor{gray}{-{.04}} & .27$\pm$.03 \textcolor{gray}{+{.09}} & .27$\pm$.03 \textcolor{gray}{+{.11}} \\
\cmidrule(lr){2-12}
\multirow{2}{*}{\rotatebox{90}{ru}} & Last Query & & .72$\pm$.01 \textcolor{gray}{+{.03}} & .30$\pm$.01 \textcolor{gray}{+{.02}} & .12$\pm$.00 \textcolor{gray}{-{.04}} & .11$\pm$.01 \textcolor{gray}{-{.05}} &  & .68$\pm$.01 \textcolor{gray}{-{.05}} & .35$\pm$.02 \textcolor{gray}{-{.11}} & .23$\pm$.02 \textcolor{gray}{+{.04}} & .21$\pm$.03 \textcolor{gray}{+{.05}} \\
 & Last Answer & & .80$\pm$.01 \textcolor{gray}{+{.01}} & .43$\pm$.03 \textcolor{gray}{+{.00}} & .15$\pm$.03 \textcolor{gray}{+{.03}} & .16$\pm$.04 \textcolor{gray}{+{.05}} &  & .72$\pm$.01 \textcolor{gray}{-{.08}} & .43$\pm$.01 \textcolor{gray}{-{.13}} & .39$\pm$.03 \textcolor{gray}{+{.22}} & .42$\pm$.03 \textcolor{gray}{+{.28}} \\
\cmidrule(lr){2-12}
\multirow{2}{*}{\rotatebox{90}{ja}} & Last Query & & .57$\pm$.02 \textcolor{gray}{-{.21}} & .16$\pm$.01 \textcolor{gray}{-{.17}} & .13$\pm$.01 \textcolor{gray}{+{.00}} & .11$\pm$.01 \textcolor{gray}{+{.00}} &  & .64$\pm$.01 \textcolor{gray}{-{.10}} & .32$\pm$.02 \textcolor{gray}{-{.13}} & .20$\pm$.02 \textcolor{gray}{+{.01}} & .17$\pm$.03 \textcolor{gray}{+{.00}} \\
 & Last Answer & & .64$\pm$.04 \textcolor{gray}{-{.23}} & .23$\pm$.04 \textcolor{gray}{-{.30}} & .16$\pm$.02 \textcolor{gray}{+{.07}} & .16$\pm$.02 \textcolor{gray}{+{.09}} &  & .72$\pm$.01 \textcolor{gray}{-{.07}} & .44$\pm$.02 \textcolor{gray}{-{.08}} & .31$\pm$.02 \textcolor{gray}{+{.13}} & .31$\pm$.02 \textcolor{gray}{+{.15}} \\
\bottomrule
\end{tabular}%
}
\caption{Results for Qwen3 8B on MKQA and Global-MMLU. Top block: baselines (\emph{majority}, \emph{prior prob.}), reported as the average across all languages (\emph{all*}), and the probe trained and tested on French (\emph{fr}). Lower block: zero-shot performance of the French-trained probe on each test language. \emph{Last Query} / \emph{Last Answer} = probe trained on last query / last answer hidden states. Values are mean\,$\pm$\,std.\ across seeds. Gray numbers give the difference to the target-language oracle probe, i.e.\@ $\Delta < 0$ indicates that the oracle scored higher.}
\label{tab:confidence-estimates-qwen}
\end{table*}

\begin{figure*}[htb]
    \centering
    \includegraphics[width=0.985\textwidth]{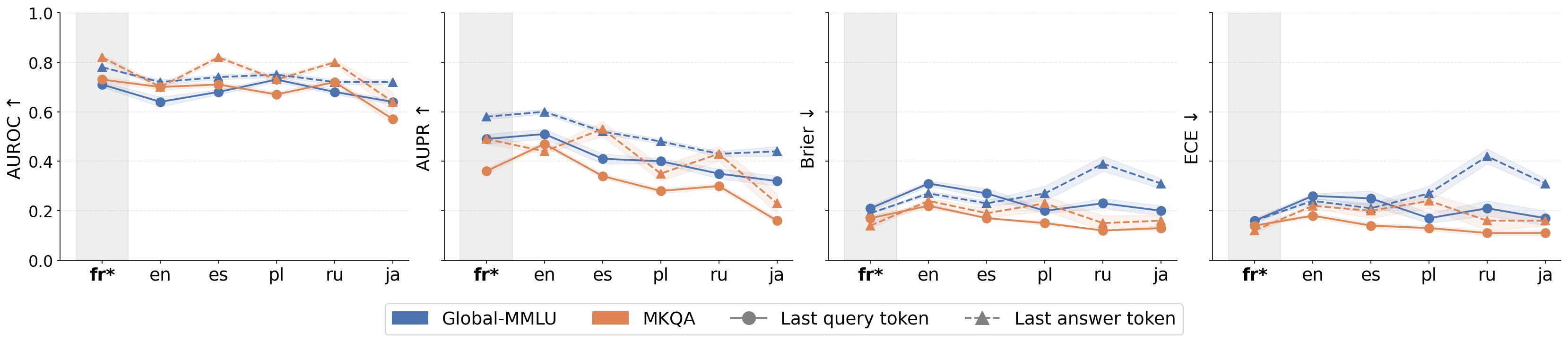}
    \caption{Cross-lingual generalization of the probe trained on French (fr*) for Qwen3 8B. Each panel compares the last query token (circle, solid) against the last answer token (triangle, dashed) across test languages, for Global-MMLU (blue) and MKQA (orange). Shaded bands show ±1 standard deviation.}\label{fig:zero-shot-qwen}
\end{figure*}

\paragraph{Results.}
We show detailed results for Qwen3 8B in \cref{tab:confidence-estimates-qwen} and visualize the overall metric trends in \cref{fig:zero-shot-qwen}, with complementary results for Llama~3.1 8B in \cref{app:conf-est-french-source}. We observe the following.
First, the \emph{majority} and \emph{prior probability} baselines can appear well-calibrated on ECE, which is not a proper scoring rule \citep{liu2023simple}, yet they fail to discriminate correct from incorrect predictions\footnote{We compute AUPR by trapezoidal integration (\texttt{sklearn.metrics.auc}), which inflates it for constant-prediction baselines. Average precision (equal to the positive-class prevalence) is a more appropriate reference, and closely matches trapezoidal AUPR for non-degenerate methods \citep{boyd-2013}.}.
Second, the probe performs well when tested on its training language, French. 
Third, the degradation from French to more distant languages is
clearer for discrimination (AUROC, AUPR) than for calibration (Brier, ECE). Discrimination falls in a fairly consistent, roughly distance-related manner, as exemplified by the difference to the oracle, whereas calibration changes are less systematic, in some settings remaining stable and in others degrading sharply.
Nonetheless, in many settings the zero-shot probe stays close to the target-language oracle, and occasionally matches or exceeds it, indicating that cross-lingual transfer of correctness features is often viable despite these drops. 
The more systematic, distance-related decline in discrimination without a corresponding calibration trend is especially
visible in \cref{fig:zero-shot-qwen}. This asymmetry suggests that the two metrics respond to different kinds of dataset shift  \citep{moreno2012unifying}: discrimination degrades with covariate shift in the multilingual latent space, which grows with linguistic distance, whereas calibration is more sensitive to prior shift in the label distribution, reflecting the lower target LLM accuracy on languages such as Russian and Japanese. 
Lastly, the last answer hidden states yield better discrimination and the last query token lower calibration error. The former is expected, as the query latents do not encode information about the generated answer.
It is worth noting that repeating the analysis for all source–target pairs (see \cref{app:conf-est-all-sources}) shows that discrimination and calibration transfer decouple. 
Discrimination broadly follows linguistic distance, with languages of the same genus and script transferring best. 
Zero-shot transfer to the most distant language is consistently most challenging, but does not necessarily indicate the same for the opposite direction. 
Calibration transfer instead tracks how close the LLM's accuracy is in source and target (\cref{tab:llm_performance}), likely because similar accuracy induces a smaller label-prior shift.

\subsection{Comparison To Other Estimators}\label{sec:comparison-estimator}
We further compare our probe (Cross-ling.\@ Probe) to other common CE methods.\footnote{We only show results for the last answer token for comparability. See \cref{app:uncertainty-baselines} for the last query token results.} We consider the length-normalized sequence likelihood (Seq.\@ Likelihood), P(True) \citep{kadavath2022language}, which asks whether a response was true or false and considers the probability of the ``true'' token, verbalized confidence (Verbalized Conf.; \citealp{lin2022teaching, xiong2024can,
ulmer2025anthropomimetic}), which elicits a confidence score from $0$-$100$, prompted in the target language, and the mass-mean probe by \citet{marks-2023}, which is not trained but directly derived from the centroids of correct and incorrect predictions.\footnote{Compared to \citeauthor{marks-2023}, we do not use patching, but apply their probe directly to the middle-layer latents of correct and incorrect generations.} Prompts for P(True) and verbalized confidence are shown in \cref{sec:llm_parameters_appendix}. 
Note that, apart from the mass-mean probe, the baselines are not zero-shot in the sense of having an explicit train-test distribution shift. They are yet intuitive choices for CE in unseen languages.


\begin{figure*}
  \begin{minipage}[b]{.7\linewidth}
    \centering
\ra{1.3}
\renewcommand{\arraystretch}{1.5}
\setlength{\tabcolsep}{4pt}
\footnotesize
\resizebox{1\textwidth}{!}{%
\begin{tabular}{llcccccccccc}
\toprule
 & & & \multicolumn{4}{c}{\textbf{MKQA}} & & \multicolumn{4}{c}{\textbf{Global-MMLU}} \\
\cmidrule(lr){4-7} \cmidrule(lr){9-12}
 & & & AUROC$\uparrow$ & AUPR$\uparrow$ & Brier$\downarrow$ & ECE$\downarrow$ & & AUROC$\uparrow$ & AUPR$\uparrow$ & Brier$\downarrow$ & ECE$\downarrow$ \\
\midrule
\multirow{6}{*}{\rotatebox{90}{es}} & Seq. Likelihood & & .81$\pm$.00 & .52$\pm$.00 & .51$\pm$.00 & .61$\pm$.00 &  & .70$\pm$.00 & .51$\pm$.00 & .53$\pm$.00 & .59$\pm$.00 \\
 & P(True) & & .30$\pm$.00 & .13$\pm$.00 & .19$\pm$.00 & .19$\pm$.00 &  & .41$\pm$.00 & .23$\pm$.00 & .26$\pm$.00 & .26$\pm$.00 \\
 & Mass-Mean Probe & & .65$\pm$.00 & .52$\pm$.00 & .45$\pm$.00 & .45$\pm$.00 &  & .59$\pm$.00 & .50$\pm$.00 & .41$\pm$.00 & .41$\pm$.00 \\
 & Verbalized Conf. & & .75$\pm$.00 & .51$\pm$.00 & .54$\pm$.00 & .59$\pm$.00 &  & .73$\pm$.00 & .55$\pm$.00 & .52$\pm$.00 & .58$\pm$.00 \\
\noalign{\vskip 1pt}
\cdashline{2-12}
\noalign{\vskip 1pt}
 & Cross-ling. Probe & & .82$\pm$.01 & .53$\pm$.03 & .19$\pm$.02 & .20$\pm$.03 &  & .74$\pm$.01 & .52$\pm$.01 & .23$\pm$.01 & .21$\pm$.02 \\
\cmidrule(lr){2-12}
\multirow{6}{*}{\rotatebox{90}{ru}} & Seq. Likelihood & & .79$\pm$.00 & .35$\pm$.00 & .49$\pm$.00 & .62$\pm$.00 &  & .75$\pm$.00 & .47$\pm$.00 & .54$\pm$.00 & .62$\pm$.00 \\
 & P(True) & & .43$\pm$.00 & .12$\pm$.00 & .12$\pm$.00 & .12$\pm$.00 &  & .45$\pm$.00 & .21$\pm$.00 & .21$\pm$.00 & .21$\pm$.00 \\
 & Mass-Mean Probe & & .73$\pm$.00 & .51$\pm$.00 & .40$\pm$.00 & .41$\pm$.00 &  & .62$\pm$.00 & .57$\pm$.00 & .51$\pm$.00 & .51$\pm$.00 \\
 & Verbalized Conf. & & .72$\pm$.00 & .36$\pm$.00 & .55$\pm$.00 & .61$\pm$.00 &  & .73$\pm$.00 & .49$\pm$.00 & .51$\pm$.00 & .57$\pm$.00 \\
\noalign{\vskip 1pt}
\cdashline{2-12}
\noalign{\vskip 1pt}
 & Cross-ling. Probe & & .80$\pm$.01 & .43$\pm$.03 & .15$\pm$.03 & .16$\pm$.04 &  & .72$\pm$.01 & .43$\pm$.01 & .39$\pm$.03 & .42$\pm$.03 \\
\cmidrule(lr){2-12}
\multirow{6}{*}{\rotatebox{90}{ja}} & Seq. Likelihood & & .81$\pm$.00 & .41$\pm$.00 & .43$\pm$.00 & .56$\pm$.00 &  & .74$\pm$.00 & .49$\pm$.00 & .48$\pm$.00 & .57$\pm$.00 \\
 & P(True) & & .60$\pm$.00 & .16$\pm$.00 & .13$\pm$.00 & .13$\pm$.00 &  & .54$\pm$.00 & .24$\pm$.00 & .21$\pm$.00 & .21$\pm$.00 \\
 & Mass-Mean Probe & & .83$\pm$.00 & .57$\pm$.00 & .22$\pm$.00 & .23$\pm$.00 &  & .52$\pm$.00 & .59$\pm$.00 & .73$\pm$.00 & .73$\pm$.00 \\
 & Verbalized Conf. & & .67$\pm$.00 & .37$\pm$.00 & .66$\pm$.00 & .72$\pm$.00 &  & .72$\pm$.00 & .44$\pm$.00 & .57$\pm$.00 & .63$\pm$.00 \\
\noalign{\vskip 1pt}
\cdashline{2-12}
\noalign{\vskip 1pt}
 & Cross-ling. Probe & & .64$\pm$.04 & .23$\pm$.04 & .16$\pm$.02 & .16$\pm$.02 &  & .72$\pm$.01 & .44$\pm$.02 & .31$\pm$.02 & .31$\pm$.02 \\
\bottomrule
\end{tabular}%
}
\captionof{table}{Comparison of zero-shot confidence estimation methods for Qwen3 8B on MKQA and Global-MMLU on selected languages.}
\label[table]{tab:baseline-comparison-qwen3_8b}
  \end{minipage}\hfill
  \begin{minipage}[b]{.28\linewidth}
    \centering
    \includegraphics[width=\linewidth]{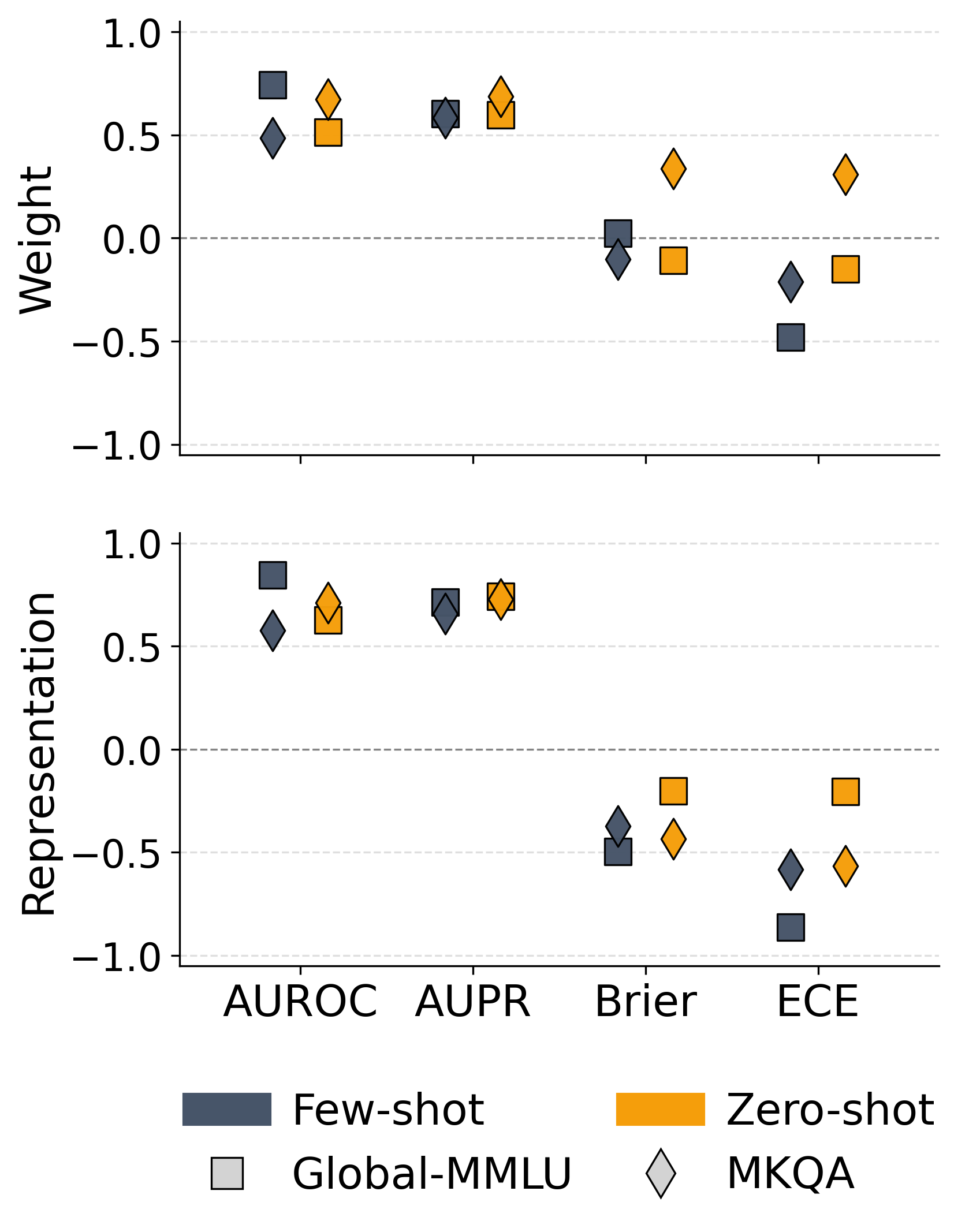} 
    \caption{Spearman's $\rho$ between layer importance and per-layer ablation impact for last answer token for Qwen3 8B. (Top): weight-based; (Bottom): representation-based.}\label{fig:spear-qwen-answer}
  \end{minipage}
\end{figure*}

\paragraph{Results.}
\cref{tab:baseline-comparison-qwen3_8b} reports results of the last token answer hidden states for selected languages, with the full evaluation in \cref{app:uncertainty-baselines}. 
At the closest target, Spanish, the probe is the strongest estimator overall, leading on AUROC and remaining competitive on AUPR and calibration. 
On the more distant Slavic languages, it remains competitive on both discrimination and calibration in most setups.
Discrimination is weakest for Japanese, the most distant target, where baselines outperform the probe on MKQA but none dominates across setups. Relative to the baselines, calibration is the probe's most consistent dimension.
\\\indent
Overall, the probe pairs competitive discrimination and calibration. Sequence likelihood discriminates comparably but is miscalibrated, verbalized confidence discriminates only moderately and is the most poorly calibrated of all methods, the mass-mean probe discriminates competitively in places but is inconsistently calibrated, and P(True) is well-calibrated but ranks poorly. 
That both the cross-lingual and mass-mean probes recover transferable signal supports the hypothesis of language-agnostic confidence features, and suggests that calibration may benefit from improved probe training data and architecture or from post-hoc recalibration methods such as temperature scaling \citep{guo2017calibration} or isotonic regression \citep{oron2017centered}.

\subsection{Understanding Layer Importance}\label{sec:ablations}

\begin{figure*}[htb]
    \begin{subfigure}[t]{0.485\linewidth}
    \centering
        \includegraphics[width=\linewidth]{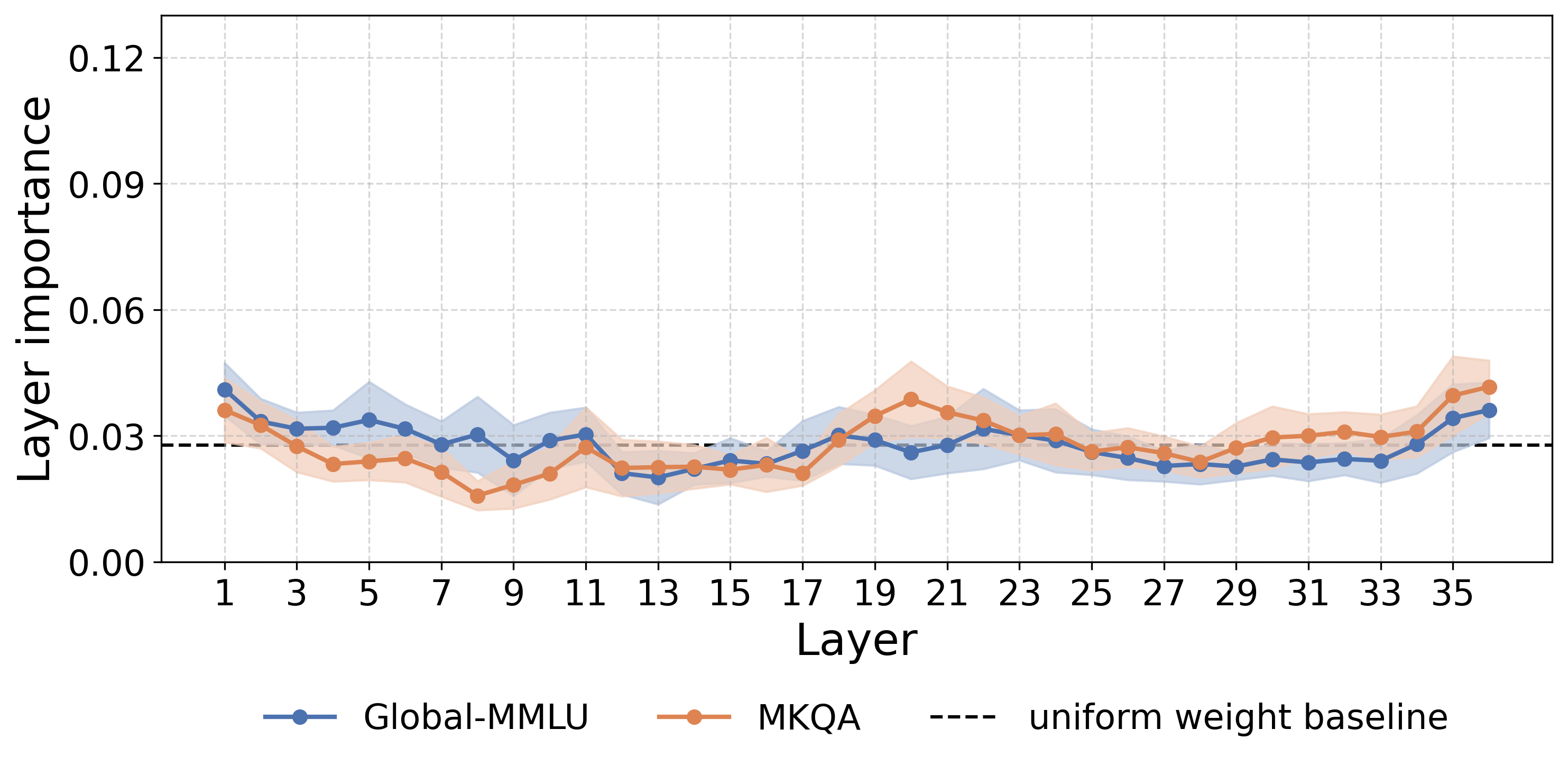}
        \caption{Layer weights for last query token.}
    \end{subfigure}
    \begin{subfigure}[t]{0.485\linewidth}
    \centering
        \includegraphics[width=\linewidth]{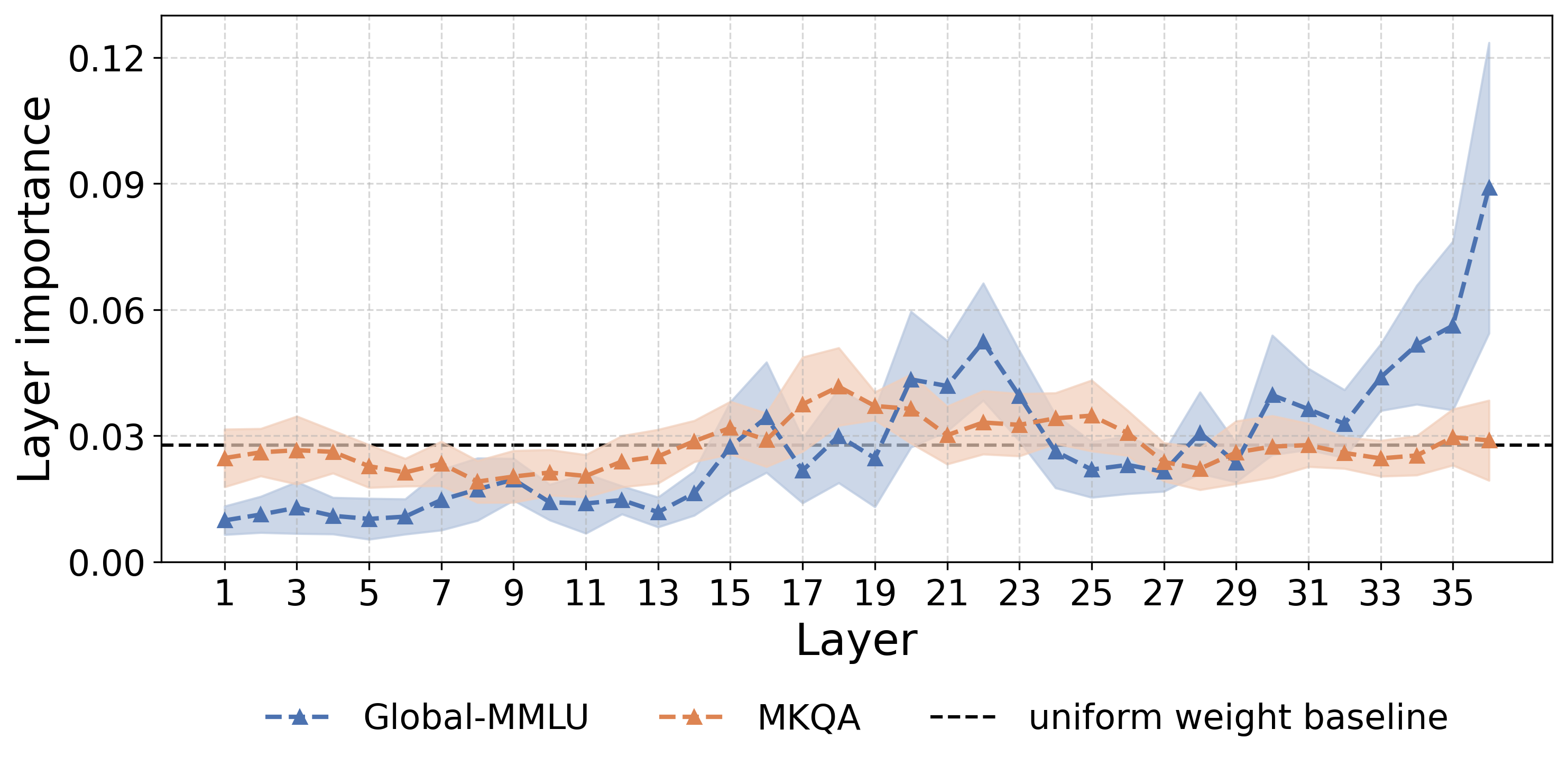}
        \caption{Layer weights for last answer token.}
    \end{subfigure}
    \caption{Distribution of learned layer weights from probes trained with different random seeds on the hidden states of the last (a) query token and (b) answer token for Qwen3 8B on Global-MMLU and MKQA. Shaded areas indicate ±1 standard deviation, dashed line shows uniform weight.}\label{fig:layer_importance_qwen}
\end{figure*}


To understand what makes the probe work, we first examine the learned layer weights in \cref{fig:layer_importance_qwen} (Llama 3.1 8B in \cref{app:layer-weights-french}). The weight distribution is more uniform for the last query token, whereas for the last answer token both datasets down-weight early layers in favor of middle ones. This structure is largely independent of the source language. Layer-weights vary less across source languages than across random seeds ($\sigma_{\text{lang}}/\sigma_{\text{seed}} = .49$-$.84$), and sources agree on relative layer importance (mean pairwise Spearman $\rho = .69$ over all 15 language pairs per condition). Full results are in \cref{app:layer-weights-all-sources}.

To test whether these weights \emph{causally} affect performance, we apply the interventional framework of \citet{geiger-2025}, assessing the importance of layer $\ell$ via the performance gap on a metric $\mathcal{P}$ between a model $\mathcal{M}$ and a variant $\mathcal{M}^{\setminus \ell}$ in which only the component $\ell$ is \emph{changed}:
\begin{equation} \label{eq:delta_ablation}
    \Delta(\mathcal{M}, \ell) :=  \mathcal{P}(\mathcal{M}) - \mathcal{P}(\mathcal{M} ^{\setminus \ell}),
\end{equation} 
where positive values of $\Delta$ indicate that the component $\ell$ contributes positively to performance.
We apply this in two ways: 
by setting specific weights $\alpha_\ell = 0$ and re-normalizing the distribution (\emph{weight ablation}), and by shuffling each layer's representations across examples (\emph{representation ablation}).\footnote{Given the redundancy of adjacent transformer layers \citep{men2025shortgpt}, we use a sliding window of size $3$.} Further details are provided in \cref{app:ablation-results}.

\begin{figure*}[htb]
    \centering
    \begin{subfigure}[t]{0.49\linewidth}
        \centering
        \includegraphics[width=0.95\linewidth]{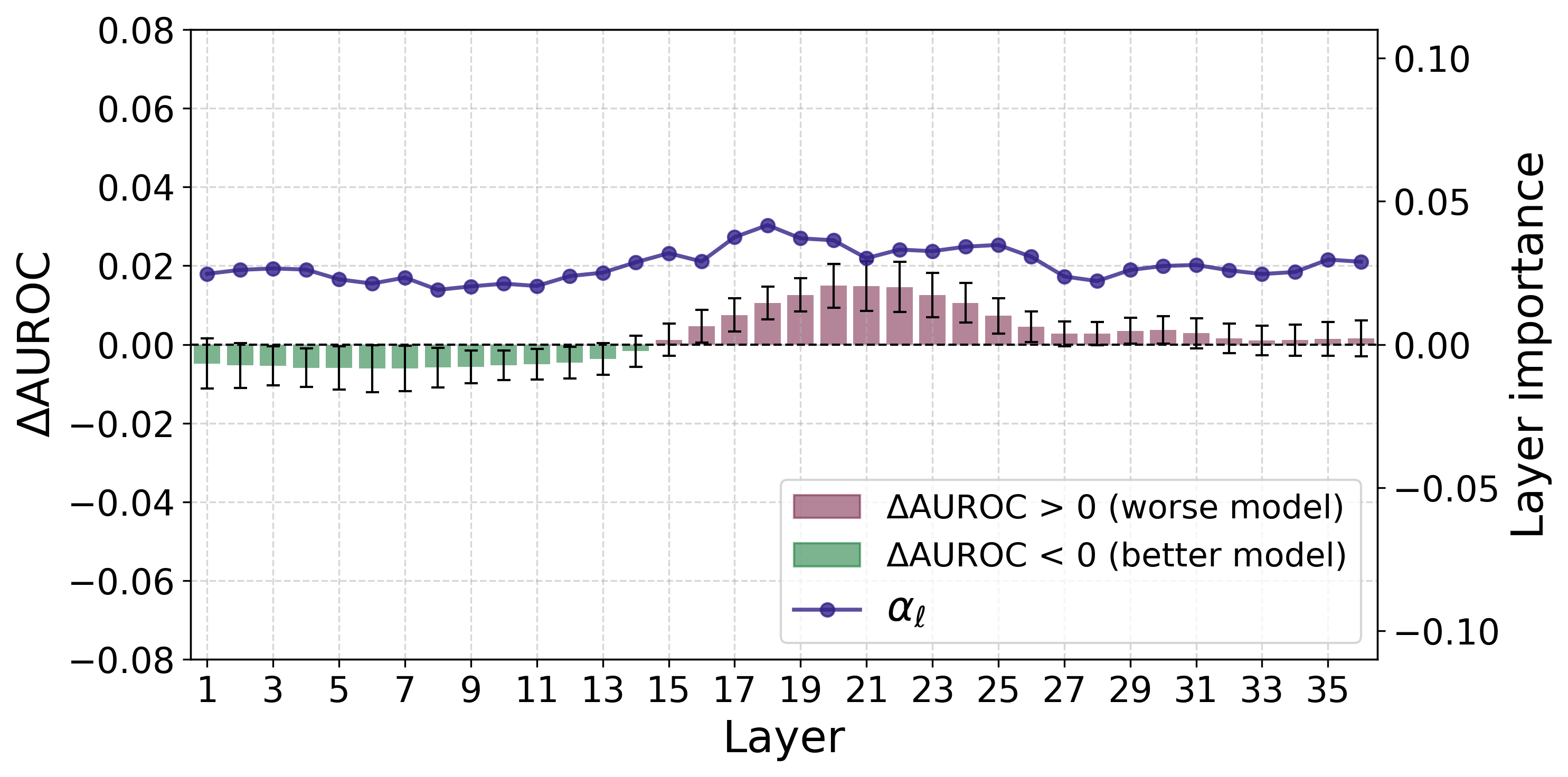}
        \caption{$\Delta$ in AUROC on MKQA.}
    \end{subfigure}
    \begin{subfigure}[t]{0.49\linewidth}
    \centering
        \includegraphics[width=0.95\linewidth]
        {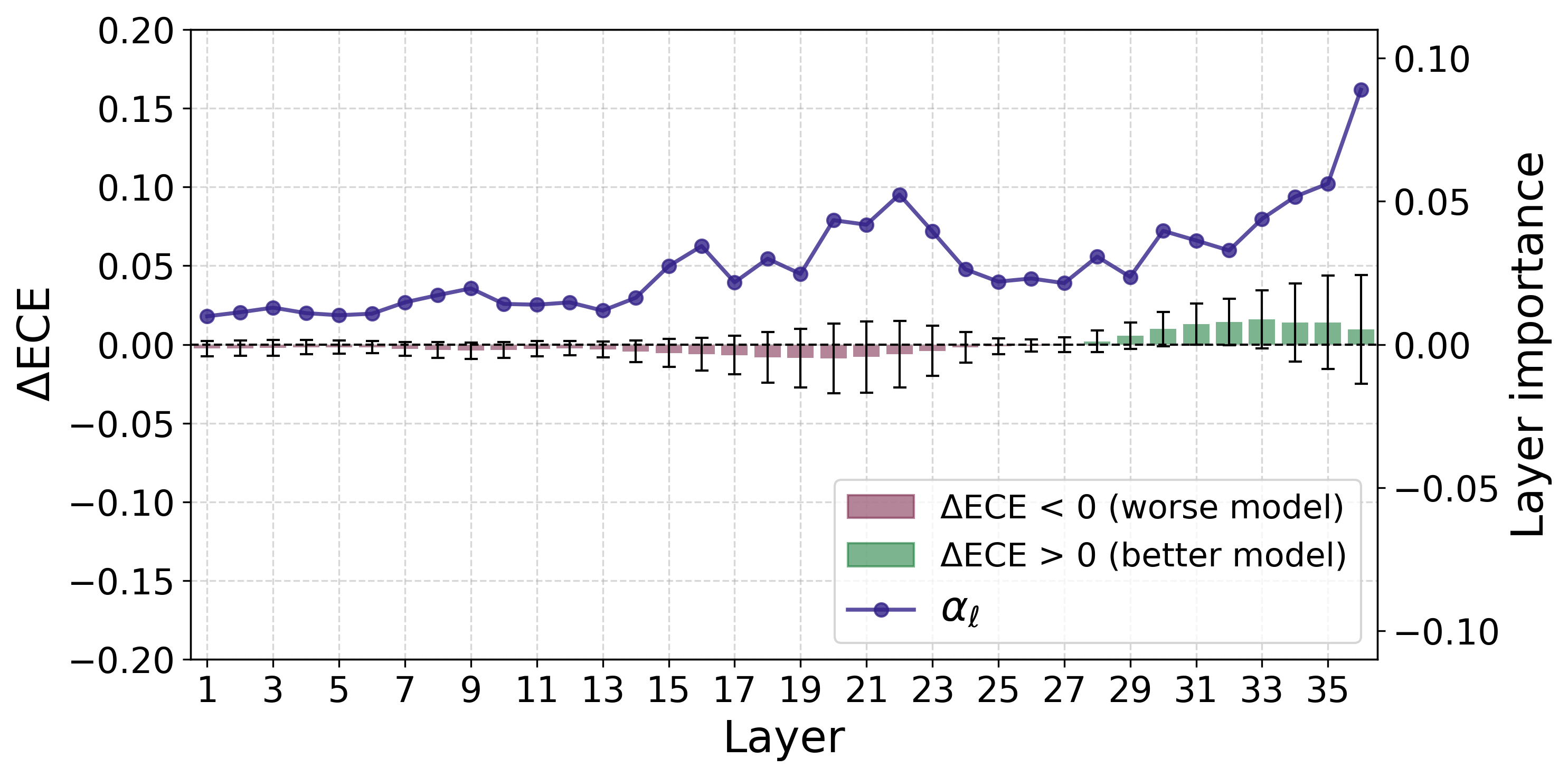}
        \caption{$\Delta$ in ECE score on Global-MMLU.}
    \end{subfigure}
    \caption{Impact of weight ablation on zero-shot AUROC and ECE (avg.\@ over all languages) for Qwen3 8B on the last answer token hidden states. Bars show mean change in metric value over 10 random seeds. Whiskers denote $\pm$1 standard deviation. Red indicates that ablation degrades performance, i.e.\ the layer is beneficial, green that ablation improves performance, i.e.\ the layer is not beneficial for zero-shot transfer.}\label{fig:ablation_examples}
\end{figure*}

\begin{figure*}[htb]
    \begin{subfigure}[t]{0.485\linewidth}
    \centering
        \includegraphics[width=\linewidth]{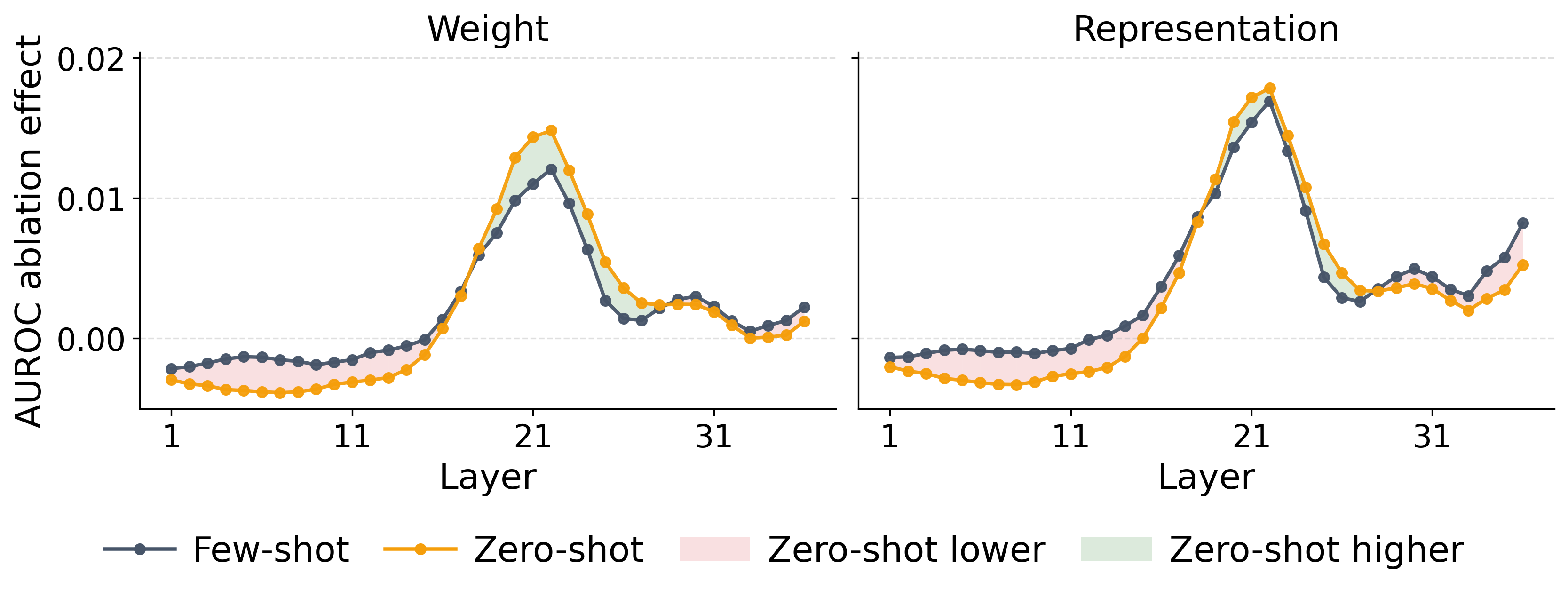}
        \caption{Ablation effect on AUROC.}
    \end{subfigure}
    \begin{subfigure}[t]{0.485\linewidth}
    \centering
        \includegraphics[width=\linewidth]{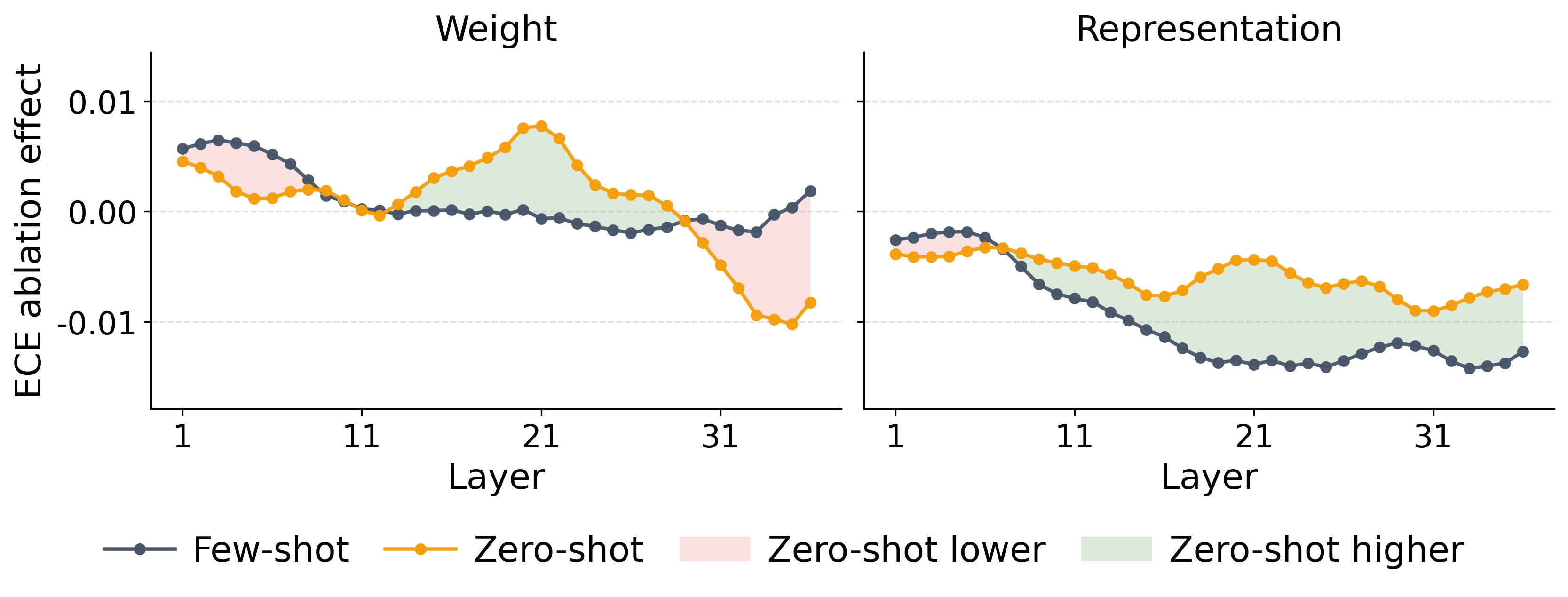}
        \caption{Ablation effect on ECE.}
    \end{subfigure}
    \caption{Ablation effect of layers on (a) AUROC and (b) ECE by few-shot (fr) and zero-shot results for Qwen3 8B and probe trained on last answer token hidden states.}\label{fig:few_zero_ablation_curves}
\end{figure*}

\begin{figure*}[htb]
    \begin{subfigure}[t]{0.485\linewidth}
    \centering
        \includegraphics[width=\linewidth]{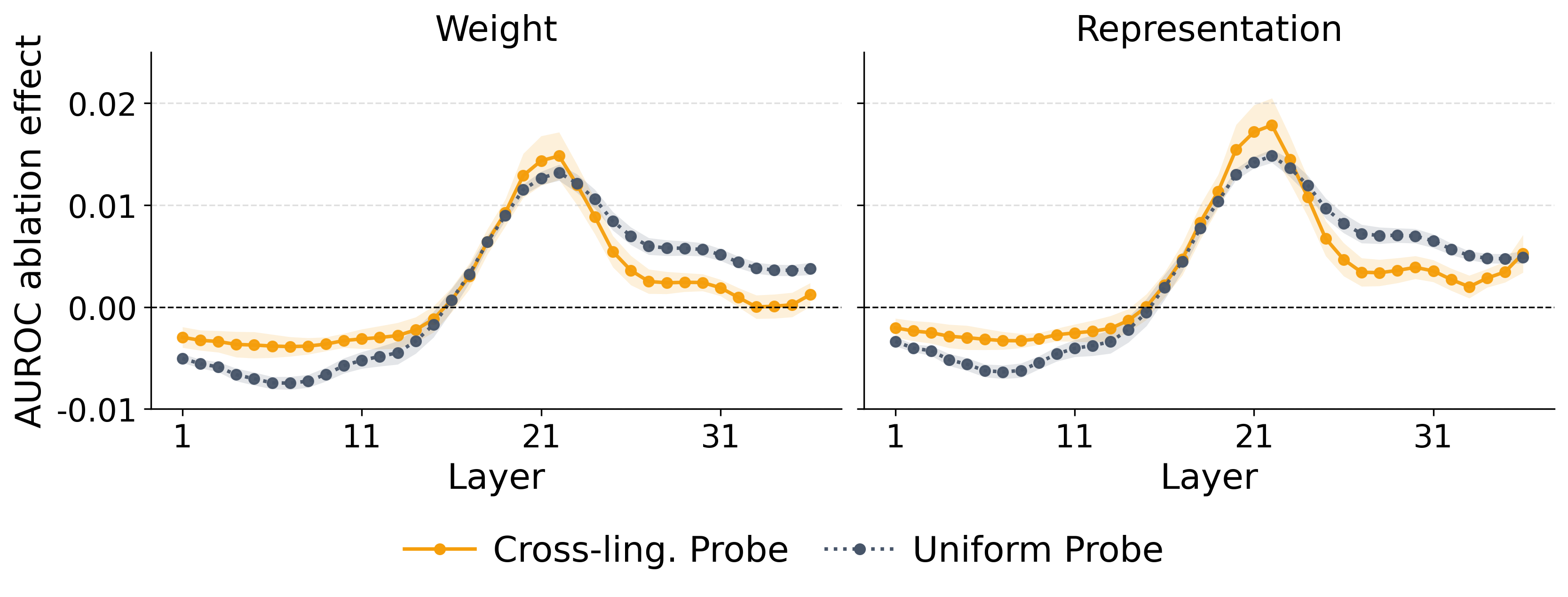}
        \caption{Ablation effect on AUROC.}
    \end{subfigure}
    \begin{subfigure}[t]{0.485\linewidth}
    \centering
        \includegraphics[width=\linewidth]{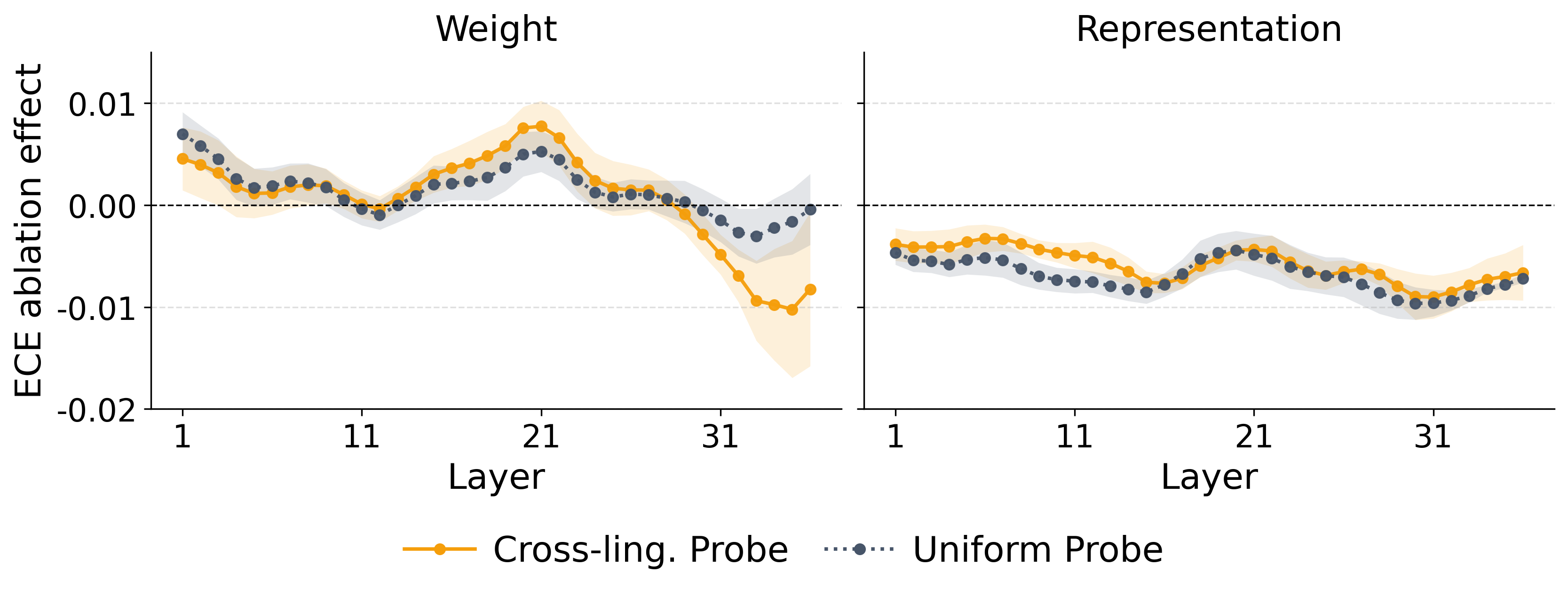}
        \caption{Ablation effect on ECE.}
    \end{subfigure}
    \caption{Ablation effect of layers on (a) AUROC and (b) ECE for the learned-weight (Cross-ling.) and uniform-weight probes, for Qwen3 8B trained on French on the last answer token hidden states, averaged over both datasets and all zero-shot languages. Positive values indicate that ablation degrades performance, for both metrics. Shaded areas indicate $\pm$1 standard deviation over 10 random seeds.}\label{fig:uniform_vs_learned}
\end{figure*}

\paragraph{Results.} 
We show a selection of results in \cref{fig:ablation_examples}, with the rest given in \cref{app:ablation-results}.
First, we observe that removing middle layers leads to a less performant probe. In some cases, the ablation of later layers yields improvements, likely due to the probe learning language-specific information about the source that is not useful in a zero-shot setting.
Second, by measuring Spearman's $\rho$ between layer weights and ablation impact per metric in \cref{fig:spear-qwen-answer}, we find a correlation between layer weights and ablation impact that is preserved in zero-shot settings, though the effect is less consistent on calibration. 
Third, by plotting the layer ablation impact separately for the few-shot vs.\@ zero-shot setting in \cref{fig:few_zero_ablation_curves}, we find that the middle-layer ablation has a stronger degrading effect on calibration in the zero-shot setting, while its effect on discrimination remains similar across both settings.
\\\indent
Finally, \cref{app:uniform-probe-weights} shows that a probe with fixed uniform layer weights performs comparably to the learned-weight probe.
We attribute this to the residual stream (\cref{eq:residual-stream}), which makes adjacent layers similar \citep{men2025shortgpt}, as \cref{fig:multilingual-pca} illustrates. 
This enables the linear layer $\mathbf{v}$ to recover middle-layer information preserved in the pooled representation.
Thus, the uniform probe's comparable performance implies that the confidence signal can be recovered in multiple ways, not that it is spread evenly across layers. 
Repeating both ablations on the uniform-weight probe supports this. 
Averaged over datasets, token positions and zero-shot languages, both probes peak at the same layers and their per-layer AUROC/AUPR profiles agree closely (Spearman $\rho \geq$ .92 across all eight configurations, for both ablation types).
Yet, calibration effects are smaller and less localized.
E.g., \cref{fig:uniform_vs_learned} shows the last-answer-token results for Qwen3 8B. Therefore, the layer-wise structure our ablations reveal is a property of the representations, not an artifact of the learned weights.

\section{Discussion}\label{sec:discussion}

Our results indicate that the language-general spaces in multilingual LLMs found by prior work (e.g.\@ \citealp{wendler-2024, zhao-2024, schut-2025}) carry useful signal for confidence estimation. While \citet{zhou2025beyond} show that the best representations for multilingual confidence estimation indeed lie in middle layers, something that our ablations confirm, we are the first to demonstrate that this information can be used \emph{zero-shot} on unseen languages. Notably, this middle-layer structure is a property of the representations rather than of the learned weights, as a probe with fixed uniform layer weights performs comparably and shows near-identical per-layer ablation profiles (\cref{sec:ablations}).
However, discrimination and calibration transfer differently. As similarity to the source decreases (\cref{sec:zero-shot-ce}), discrimination degrades broadly with linguistic distance. Calibration changes less systematically, tracking the difference in LLM accuracy between source and target more closely than linguistic distance.
Compared to common CE methods in \cref{sec:comparison-estimator}, the probe nonetheless combines competitive discrimination and calibration in many setups. Since the probe reads hidden states already computed during generation, it adds negligible inference cost compared to methods requiring additional queries or generations.
We also found that the choice of time step matters for zero-shot transfer. The last query token, often used in related probing work, 
was better calibrated on unseen languages,
while the lesser-used last answer token achieved better discrimination.
More broadly, these findings position internal representations as a promising foundation for multilingual confidence estimation, complementing methods that operate purely on model outputs.

\section{Conclusion}\label{sec:conclusion}

In this work, we presented the first evidence that language-agnostic latent spaces in multilingual LLMs contain correctness signals that can be exploited for confidence estimation on unseen languages, without any target-language supervision. Consistent with prior work, we find this signal concentrated in the middle, language-general layers. While transfer weakens as the target language grows more distant from the source, the probe remains a strong and practical baseline across linguistically diverse languages and, among the estimators we compare, most consistently pairs competitive discrimination with calibration.
It remains especially interesting to explore how to make this approach more robust to unseen language families and scripts by training on multiple source languages, identifying the most critical latent representations, and applying common post-hoc calibration techniques.
More broadly, we believe that these findings open research directions towards multilingual confidence estimation with little or no supervision, especially for under-served languages.


\section*{Limitations}
We acknowledge the following limitations.
First, our work focuses on open-source models of 8 billion parameters. We nonetheless expect our approach to extend to larger ones. The localization of language-agnostic \citep{zeng-2025, schut-2025, wendler-2024, tang2024language, liu-2024} and correctness-related \citep{khanmohammadi-etal-2025-calibrating, bao-etal-2025-probing, marks-2023} signals follows consistent patterns across model scales. We do not, however, test this directly, and transfer behavior at scale remains an open question.
Second, while our language selection spans different families and scripts, it remains weighted toward European, relatively high-resource languages. 
This partly reflects the limited availability of parallel open-ended QA data. The few additional languages covered by the intersection of MKQA and Global-MMLU are not generally considered among the lowest-resource tiers \citep[Table 7]{globalmmlu}.
We expect the zero-shot performance to degrade further for languages that are less related to the source, use a different script, or are less represented in training, consistent with the greater representational dissimilarity observed for more distant languages \citep{li-2025, kudugunta2019investigating, singh2019bert}.
Third, correctness labels are generated using an LLM-as-a-judge, whose reliability may vary across languages, potentially introducing language-correlated label noise.
Fourth, we focus on short-form QA, whereas recent works have also started to consider reasoning \citep{zhang2025token, yoon2025reasoning, zhang2025all, razghandi2025cer, mao2026confidence} and long-form uncertainty quantification \citep{liu2024litcablightweightlanguagemodel, zhang2025atomic, fan2026iuq, bouchard2026fine}. Our approach may extend to these settings, but longer generations likely require more careful selection of the hidden-state time step.
Finally, affine probes may be insufficient in some cases. Self-attention and feed-forward blocks can substantially rotate and mix representation dimensions, so readouts may miss non-linearly encoded structure, while our weighted average of layer representations may not capture relevant cross-layer interactions.

\section*{Ethical Considerations}

We foresee no immediate ethical issues.

\section*{Acknowledgments}
This work is supported by the Dutch National Science Foundation (NWO Vici VI.C.212.053) and the HELLENiQ ENERGY's "Proud of Youth" Scholarship. Computational resources were provided by the Cirrus UK National Tier-2 HPC Service at EPCC (\url{http://www.cirrus.ac.uk}), funded by the University of Edinburgh and EPSRC (EP/P020267/1). We thank Rochelle Choenni for her help in conceptualizing this project. We further thank Adam Ko\l{}ek, Alexey Dronov, Alina Erkulova, Diego Villa Almeyda, Diego Z\'u\~niga Olave, Jakub Baczy\'nski-Keller, Jos\'e Carlos Mart\'inez Santamar\'ia, Marta Jankowska, Masato Yamamoto, Rintaro Nagashima, and Taiyo Yamamoto for reviewing our prompts and annotating generations for correctness, without whom the multilingual nature of this study would not have been possible.





\bibliography{custom}

@article{mkqa,
	title        = {{MKQA}: A Linguistically Diverse Benchmark for Multilingual Open Domain Question Answering},
	author       = {Longpre, Shayne  and Lu, Yi  and Daiber, Joachim},
	year         = 2021,
	journal      = {Transactions of the Association for Computational Linguistics},
	publisher    = {MIT Press},
	address      = {Cambridge, MA},
	volume       = 9,
	pages        = {1389--1406},
	doi          = {10.1162/tacl_a_00433},
	url          = {https://aclanthology.org/2021.tacl-1.82/},
	editor       = {Roark, Brian  and Nenkova, Ani}
}

@article{natural-questions,
	title        = {Natural Questions: A Benchmark for Question Answering Research},
	author       = {Kwiatkowski, Tom  and Palomaki, Jennimaria  and Redfield, Olivia  and Collins, Michael  and Parikh, Ankur  and Alberti, Chris  and Epstein, Danielle  and Polosukhin, Illia  and Devlin, Jacob  and Lee, Kenton  and Toutanova, Kristina  and Jones, Llion  and Kelcey, Matthew  and Chang, Ming-Wei  and Dai, Andrew M.  and Uszkoreit, Jakob  and Le, Quoc  and Petrov, Slav},
	year         = 2019,
	journal      = {Transactions of the Association for Computational Linguistics},
	publisher    = {MIT Press},
	address      = {Cambridge, MA},
	volume       = 7,
	pages        = {452--466},
	doi          = {10.1162/tacl_a_00276},
	url          = {https://aclanthology.org/Q19-1026/},
	editor       = {Lee, Lillian  and Johnson, Mark  and Roark, Brian  and Nenkova, Ani}
}

@book{wals,
	title        = {WALS Online (v2020.4)},
	year         = 2013,
	publisher    = {Zenodo},
	doi          = {10.5281/zenodo.13950591},
	url          = {https://doi.org/10.5281/zenodo.13950591},
	editor       = {Matthew S. Dryer and Martin Haspelmath},
	type         = {Data set}
}

@inproceedings{globalmmlu,
    title = "Global {MMLU}: Understanding and Addressing Cultural and Linguistic Biases in Multilingual Evaluation",
    author = "Singh, Shivalika  and
      Romanou, Angelika  and
      Fourrier, Cl{\'e}mentine  and
      Adelani, David Ifeoluwa  and
      Ngui, Jian Gang  and
      Vila-Suero, Daniel  and
      Limkonchotiwat, Peerat  and
      Marchisio, Kelly  and
      Leong, Wei Qi  and
      Susanto, Yosephine  and
      Ng, Raymond  and
      Longpre, Shayne  and
      Ruder, Sebastian  and
      Ko, Wei-Yin  and
      Bosselut, Antoine  and
      Oh, Alice  and
      Martins, Andre  and
      Choshen, Leshem  and
      Ippolito, Daphne  and
      Ferrante, Enzo  and
      Fadaee, Marzieh  and
      Ermis, Beyza  and
      Hooker, Sara",
    editor = "Che, Wanxiang  and
      Nabende, Joyce  and
      Shutova, Ekaterina  and
      Pilehvar, Mohammad Taher",
    booktitle = "Proceedings of the 63rd Annual Meeting of the Association for Computational Linguistics (Volume 1: Long Papers)",
    month = jul,
    year = "2025",
    address = "Vienna, Austria",
    publisher = "Association for Computational Linguistics",
    url = "https://aclanthology.org/2025.acl-long.919/",
    doi = "10.18653/v1/2025.acl-long.919",
    pages = "18761--18799",
    ISBN = "979-8-89176-251-0"
}

@article{chandak-2025,
	title        = {Answer Matching Outperforms Multiple Choice for Language Model Evaluation},
	author       = {Nikhil Chandak and Shashwat Goel and Ameya Prabhu and Moritz Hardt and Jonas Geiping},
	year         = 2025,
	journal      = {CoRR},
	volume       = {abs/2507.02856},
	doi          = {10.48550/ARXIV.2507.02856},
	url          = {https://doi.org/10.48550/arXiv.2507.02856},
	eprinttype   = {arXiv},
	eprint       = {2507.02856},
	bibsource    = {dblp computer science bibliography, https://dblp.org}
}

@inproceedings{vaswani-2017,
	title        = {Attention is All you Need},
	author       = {Ashish Vaswani and Noam Shazeer and Niki Parmar and Jakob Uszkoreit and Llion Jones and Aidan N. Gomez and Lukasz Kaiser and Illia Polosukhin},
	year         = 2017,
	booktitle    = {Advances in Neural Information Processing Systems 30: Annual Conference on Neural Information Processing Systems 2017, December 4-9, 2017, Long Beach, CA, {USA}},
	pages        = {5998--6008},
	url          = {https://proceedings.neurips.cc/paper/2017/hash/3f5ee243547dee91fbd053c1c4a845aa-Abstract.html},
	editor       = {Isabelle Guyon and Ulrike von Luxburg and Samy Bengio and Hanna M. Wallach and Rob Fergus and S. V. N. Vishwanathan and Roman Garnett},
	bibsource    = {dblp computer science bibliography, https://dblp.org}
}

@inproceedings{wang-2024-probing,
	title        = {Probing the Emergence of Cross-lingual Alignment during {LLM} Training},
	author       = {Wang, Hetong  and Minervini, Pasquale  and Ponti, Edoardo},
	year         = 2024,
	month        = aug,
	booktitle    = {Findings of the Association for Computational Linguistics: ACL 2024},
	publisher    = {Association for Computational Linguistics},
	address      = {Bangkok, Thailand},
	pages        = {12159--12173},
	doi          = {10.18653/v1/2024.findings-acl.724},
	url          = {https://aclanthology.org/2024.findings-acl.724/},
	editor       = {Ku, Lun-Wei  and Martins, Andre  and Srikumar, Vivek}
}

@inproceedings{wendler-2024,
	title        = {Do Llamas Work in {E}nglish? On the Latent Language of Multilingual Transformers},
	author       = {Wendler, Chris  and Veselovsky, Veniamin  and Monea, Giovanni  and West, Robert},
	year         = 2024,
	month        = aug,
	booktitle    = {Proceedings of the 62nd Annual Meeting of the Association for Computational Linguistics (Volume 1: Long Papers)},
	publisher    = {Association for Computational Linguistics},
	address      = {Bangkok, Thailand},
	pages        = {15366--15394},
	doi          = {10.18653/v1/2024.acl-long.820},
	url          = {https://aclanthology.org/2024.acl-long.820/},
	editor       = {Ku, Lun-Wei  and Martins, Andre  and Srikumar, Vivek}
}

@inproceedings{dumas-2024-separating,
	title        = {Separating Tongue from Thought: Activation Patching Reveals Language-Agnostic Concept Representations in Transformers},
	author       = {Cl{\'{e}}ment Dumas and Chris Wendler and Veniamin Veselovsky and Giovanni Monea and Robert West},
	year         = 2025,
	booktitle    = {Proceedings of the 63rd Annual Meeting of the Association for Computational Linguistics (Volume 1: Long Papers), {ACL} 2025, Vienna, Austria, July 27 - August 1, 2025},
	publisher    = {Association for Computational Linguistics},
	pages        = {31822--31841},
	url          = {https://aclanthology.org/2025.acl-long.1536/},
	editor       = {Wanxiang Che and Joyce Nabende and Ekaterina Shutova and Mohammad Taher Pilehvar},
	bibsource    = {dblp computer science bibliography, https://dblp.org}
}

@article{schut-2025,
	title        = {Do Multilingual LLMs Think In English?},
	author       = {Lisa Schut and Yarin Gal and Sebastian Farquhar},
	year         = 2025,
	journal      = {CoRR},
	volume       = {abs/2502.15603},
	doi          = {10.48550/ARXIV.2502.15603},
	url          = {https://doi.org/10.48550/arXiv.2502.15603},
	eprinttype   = {arXiv},
	eprint       = {2502.15603},
	bibsource    = {dblp computer science bibliography, https://dblp.org}
}

@article{marks-2023,
	title        = {The Geometry of Truth: Emergent Linear Structure in Large Language Model Representations of True/False Datasets},
	author       = {Samuel Marks and Max Tegmark},
	year         = 2023,
	journal      = {CoRR},
	volume       = {abs/2310.06824},
	doi          = {10.48550/ARXIV.2310.06824},
	url          = {https://doi.org/10.48550/arXiv.2310.06824},
	eprinttype   = {arXiv},
	eprint       = {2310.06824},
	bibsource    = {dblp computer science bibliography, https://dblp.org}
}

@article{wang-2024-sharing,
	title        = {Sharing Matters: Analysing Neurons Across Languages and Tasks in LLMs},
	author       = {Weixuan Wang and Barry Haddow and Wei Peng and Alexandra Birch},
	year         = 2024,
	journal      = {CoRR},
	volume       = {abs/2406.09265},
	doi          = {10.48550/ARXIV.2406.09265},
	url          = {https://doi.org/10.48550/arXiv.2406.09265},
	eprinttype   = {arXiv},
	eprint       = {2406.09265},
	bibsource    = {dblp computer science bibliography, https://dblp.org}
}

@inproceedings{brinkmann-2025,
	title        = {Large Language Models Share Representations of Latent Grammatical Concepts Across Typologically Diverse Languages},
	author       = {Jannik Brinkmann and Chris Wendler and Christian Bartelt and Aaron Mueller},
	year         = 2025,
	booktitle    = {Proceedings of the 2025 Conference of the Nations of the Americas Chapter of the Association for Computational Linguistics: Human Language Technologies, {NAACL} 2025 - Volume 1: Long Papers, Albuquerque, New Mexico, USA, April 29 - May 4, 2025},
	publisher    = {Association for Computational Linguistics},
	pages        = {6131--6150},
	doi          = {10.18653/V1/2025.NAACL-LONG.312},
	url          = {https://doi.org/10.18653/v1/2025.naacl-long.312},
	editor       = {Luis Chiruzzo and Alan Ritter and Lu Wang},
	bibsource    = {dblp computer science bibliography, https://dblp.org}
}

@article{li-2025,
	title        = {Exploring Multilingual Probing in Large Language Models: {A} Cross-Language Analysis},
	author       = {Daoyang Li and Mingyu Jin and Qingcheng Zeng and Haiyan Zhao and Mengnan Du},
	year         = 2024,
	journal      = {CoRR},
	volume       = {abs/2409.14459},
	doi          = {10.48550/ARXIV.2409.14459},
	url          = {https://doi.org/10.48550/arXiv.2409.14459},
	eprinttype   = {arXiv},
	eprint       = {2409.14459},
	bibsource    = {dblp computer science bibliography, https://dblp.org}
}

@inproceedings{he-2024,
	title        = {{LLM} Factoscope: Uncovering {LLM}s' Factual Discernment through Measuring Inner States},
	author       = {He, Jinwen  and Gong, Yujia  and Lin, Zijin  and Wei, Cheng{'}an  and Zhao, Yue  and Chen, Kai},
	year         = 2024,
	month        = aug,
	booktitle    = {Findings of the Association for Computational Linguistics: ACL 2024},
	publisher    = {Association for Computational Linguistics},
	address      = {Bangkok, Thailand},
	pages        = {10218--10230},
	doi          = {10.18653/v1/2024.findings-acl.608},
	url          = {https://aclanthology.org/2024.findings-acl.608/},
	editor       = {Ku, Lun-Wei  and Martins, Andre  and Srikumar, Vivek}
}

@article{duan-2024,
	title        = {Do LLMs Know about Hallucination? An Empirical Investigation of LLM's Hidden States},
	author       = {Hanyu Duan and Yi Yang and Kar Yan Tam},
	year         = 2024,
	journal      = {CoRR},
	volume       = {abs/2402.09733},
	doi          = {10.48550/ARXIV.2402.09733},
	url          = {https://doi.org/10.48550/arXiv.2402.09733},
	eprinttype   = {arXiv},
	eprint       = {2402.09733},
	bibsource    = {dblp computer science bibliography, https://dblp.org}
}

@inproceedings{ji-2024,
	title        = {{LLM} Internal States Reveal Hallucination Risk Faced With a Query},
	author       = {Ji, Ziwei  and Chen, Delong  and Ishii, Etsuko  and Cahyawijaya, Samuel  and Bang, Yejin  and Wilie, Bryan  and Fung, Pascale},
	year         = 2024,
	month        = nov,
	booktitle    = {Proceedings of the 7th BlackboxNLP Workshop: Analyzing and Interpreting Neural Networks for NLP},
	publisher    = {Association for Computational Linguistics},
	address      = {Miami, Florida, US},
	pages        = {88--104},
	doi          = {10.18653/v1/2024.blackboxnlp-1.6},
	url          = {https://aclanthology.org/2024.blackboxnlp-1.6/},
	editor       = {Belinkov, Yonatan  and Kim, Najoung  and Jumelet, Jaap  and Mohebbi, Hosein  and Mueller, Aaron  and Chen, Hanjie}
}

@inproceedings{hendrycks-2021,
	title        = {Measuring Massive Multitask Language Understanding},
	author       = {Dan Hendrycks and Collin Burns and Steven Basart and Andy Zou and Mantas Mazeika and Dawn Song and Jacob Steinhardt},
	year         = 2021,
	booktitle    = {9th International Conference on Learning Representations, {ICLR} 2021, Virtual Event, Austria, May 3-7, 2021},
	publisher    = {OpenReview.net},
	url          = {https://openreview.net/forum?id=d7KBjmI3GmQ},
	bibsource    = {dblp computer science bibliography, https://dblp.org}
}

@misc{qwen3,
	title        = {Qwen3 Technical Report},
	author       = {{Qwen Team}},
	year         = 2025,
	url          = {https://arxiv.org/abs/2505.09388},
	eprint       = {2505.09388},
	archiveprefix = {arXiv},
	primaryclass = {cs.CL}
}

@inproceedings{shani-2025,
	title        = {Language Dominance in Multilingual Large Language Models},
	author       = {Shani, Nadav  and Basirat, Ali},
	year         = 2025,
	month        = nov,
	booktitle    = {Proceedings of the 8th BlackboxNLP Workshop: Analyzing and Interpreting Neural Networks for NLP},
	publisher    = {Association for Computational Linguistics},
	address      = {Suzhou, China},
	pages        = {137--148},
	doi          = {10.18653/v1/2025.blackboxnlp-1.7},
	isbn         = {979-8-89176-346-3},
	url          = {https://aclanthology.org/2025.blackboxnlp-1.7/},
	editor       = {Belinkov, Yonatan  and Mueller, Aaron  and Kim, Najoung  and Mohebbi, Hosein  and Chen, Hanjie  and Arad, Dana  and Sarti, Gabriele}
}

@inproceedings{zhao-2024,
	title        = {How do large language models handle multilingualism?},
	author       = {Zhao, Yiran and Zhang, Wenxuan and Chen, Guizhen and Kawaguchi, Kenji and Bing, Lidong},
	year         = 2024,
	booktitle    = {Proceedings of the 38th International Conference on Neural Information Processing Systems},
	location     = {Vancouver, BC, Canada},
	publisher    = {Curran Associates Inc.},
	address      = {Red Hook, NY, USA},
	series       = {NIPS '24},
	isbn         = 9798331314385,
	articleno    = 489,
	numpages     = 24
}

@inproceedings{kapoor2024calibration,
	title        = {Calibration-tuning: Teaching large language models to know what they don’t know},
	author       = {Kapoor, Sanyam and Gruver, Nate and Roberts, Manley and Pal, Arka and Dooley, Samuel and Goldblum, Micah and Wilson, Andrew},
	year         = 2024,
	booktitle    = {Proceedings of the 1st Workshop on Uncertainty-Aware NLP (UncertaiNLP 2024)},
	pages        = {1--14}
}

@inproceedings{tang2024language,
	title        = {Language-specific neurons: The key to multilingual capabilities in large language models},
	author       = {Tang, Tianyi and Luo, Wenyang and Huang, Haoyang and Zhang, Dongdong and Wang, Xiaolei and Zhao, Wayne Xin and Wei, Furu and Wen, Ji-Rong},
	year         = 2024,
	booktitle    = {Proceedings of the 62nd Annual Meeting of the Association for Computational Linguistics (Volume 1: Long Papers)},
	pages        = {5701--5715}
}

@inproceedings{zhang2024unveiling,
	title        = {Unveiling linguistic regions in large language models},
	author       = {Zhang, Zhihao and Zhao, Jun and Zhang, Qi and Gui, Tao and Huang, Xuan-Jing},
	year         = 2024,
	booktitle    = {Proceedings of the 62nd Annual Meeting of the Association for Computational Linguistics (Volume 1: Long Papers)},
	pages        = {6228--6247}
}

@inproceedings{kojima2024multilingual,
	title        = {On the multilingual ability of decoder-based pre-trained language models: Finding and controlling language-specific neurons},
	author       = {Kojima, Takeshi and Okimura, Itsuki and Iwasawa, Yusuke and Yanaka, Hitomi and Matsuo, Yutaka},
	year         = 2024,
	booktitle    = {Proceedings of the 2024 Conference of the North American Chapter of the Association for Computational Linguistics: Human Language Technologies (Volume 1: Long Papers)},
	pages        = {6919--6971}
}

@inproceedings{bhattacharya2023unveiling,
	title        = {Unveiling multilinguality in transformer models: Exploring language specificity in feed-forward networks},
	author       = {Bhattacharya, Sunit and Bojar, Ond{\v{r}}ej},
	year         = 2023,
	booktitle    = {Proceedings of the 6th BlackboxNLP Workshop: Analyzing and Interpreting Neural Networks for NLP},
	pages        = {120--126}
}

@inproceedings{zhu2023calibration,
	title        = {On the calibration of large language models and alignment},
	author       = {Zhu, Chiwei and Xu, Benfeng and Wang, Quan and Zhang, Yongdong and Mao, Zhendong},
	year         = 2023,
	booktitle    = {Findings of the Association for Computational Linguistics: EMNLP 2023},
	pages        = {9778--9795}
}

@inproceedings{ulmer2024calibrating,
	title        = {Calibrating large language models using their generations only},
	author       = {Ulmer, Dennis and Gubri, Martin and Lee, Hwaran and Yun, Sangdoo and Oh, Seong},
	year         = 2024,
	booktitle    = {Proceedings of the 62nd Annual Meeting of the Association for Computational Linguistics (Volume 1: Long Papers)},
	pages        = {15440--15459}
}

@article{jiang2021can,
	title        = {How can we know when language models know? on the calibration of language models for question answering},
	author       = {Jiang, Zhengbao and Araki, Jun and Ding, Haibo and Neubig, Graham},
	year         = 2021,
	journal      = {Transactions of the Association for Computational Linguistics},
	publisher    = {MIT Press One Rogers Street, Cambridge, MA 02142-1209, USA journals-info~…},
	volume       = 9,
	pages        = {962--977}
}

@inproceedings{tian2023just,
	title        = {Just ask for calibration: Strategies for eliciting calibrated confidence scores from language models fine-tuned with human feedback},
	author       = {Tian, Katherine and Mitchell, Eric and Zhou, Allan and Sharma, Archit and Rafailov, Rafael and Yao, Huaxiu and Finn, Chelsea and Manning, Christopher D},
	year         = 2023,
	booktitle    = {Proceedings of the 2023 Conference on Empirical Methods in Natural Language Processing},
	pages        = {5433--5442}
}

@article{kadavath2022language,
	title        = {Language models (mostly) know what they know},
	author       = {Kadavath, Saurav and Conerly, Tom and Askell, Amanda and Henighan, Tom and Drain, Dawn and Perez, Ethan and Schiefer, Nicholas and Hatfield-Dodds, Zac and DasSarma, Nova and Tran-Johnson, Eli and others},
	year         = 2022,
	journal      = {arXiv preprint arXiv:2207.05221}
}

@article{huang2025survey,
	title        = {A survey on hallucination in large language models: Principles, taxonomy, challenges, and open questions},
	author       = {Huang, Lei and Yu, Weijiang and Ma, Weitao and Zhong, Weihong and Feng, Zhangyin and Wang, Haotian and Chen, Qianglong and Peng, Weihua and Feng, Xiaocheng and Qin, Bing and others},
	year         = 2025,
	journal      = {ACM Transactions on Information Systems},
	publisher    = {ACM New York, NY},
	volume       = 43,
	number       = 2,
	pages        = {1--55}
}

@article{wang2024understanding,
	title        = {Understanding user experience in large language model interactions},
	author       = {Wang, Jiayin and Ma, Weizhi and Sun, Peijie and Zhang, Min and Nie, Jian-Yun},
	year         = 2024,
	journal      = {arXiv preprint arXiv:2401.08329}
}

@techreport{humlum2025large,
	title        = {Large language models, small labor market effects},
	author       = {Humlum, Anders and Vestergaard, Emilie},
	year         = 2025,
	institution  = {National Bureau of Economic Research}
}

@article{minderer2021revisiting,
	title        = {Revisiting the calibration of modern neural networks},
	author       = {Minderer, Matthias and Djolonga, Josip and Romijnders, Rob and Hubis, Frances and Zhai, Xiaohua and Houlsby, Neil and Tran, Dustin and Lucic, Mario},
	year         = 2021,
	journal      = {Advances in neural information processing systems},
	volume       = 34,
	pages        = {15682--15694}
}

@article{degroot1983comparison,
	title        = {The comparison and evaluation of forecasters},
	author       = {DeGroot, Morris H and Fienberg, Stephen E},
	year         = 1983,
	journal      = {Journal of the Royal Statistical Society: Series D (The Statistician)},
	publisher    = {Wiley Online Library},
	volume       = 32,
	number       = {1-2},
	pages        = {12--22}
}

@inproceedings{naeini2015obtaining,
	title        = {Obtaining well calibrated probabilities using bayesian binning},
	author       = {Naeini, Mahdi Pakdaman and Cooper, Gregory and Hauskrecht, Milos},
	year         = 2015,
	booktitle    = {Proceedings of the AAAI conference on artificial intelligence},
	volume       = 29,
	number       = 1
}

@article{ovadia2019can,
	title        = {Can you trust your model's uncertainty? evaluating predictive uncertainty under dataset shift},
	author       = {Ovadia, Yaniv and Fertig, Emily and Ren, Jie and Nado, Zachary and Sculley, David and Nowozin, Sebastian and Dillon, Joshua and Lakshminarayanan, Balaji and Snoek, Jasper},
	year         = 2019,
	journal      = {Advances in neural information processing systems},
	volume       = 32
}

@inproceedings{guo2017calibration,
	title        = {On calibration of modern neural networks},
	author       = {Guo, Chuan and Pleiss, Geoff and Sun, Yu and Weinberger, Kilian Q},
	year         = 2017,
	booktitle    = {International conference on machine learning},
	pages        = {1321--1330},
	organization = {PMLR}
}

@inproceedings{zhang-2023-dont,
	title        = {Don{'}t Trust {C}hat{GPT} when your Question is not in {E}nglish: A Study of Multilingual Abilities and Types of {LLM}s},
	author       = {Zhang, Xiang  and Li, Senyu  and Hauer, Bradley  and Shi, Ning  and Kondrak, Grzegorz},
	year         = 2023,
	month        = dec,
	booktitle    = {Proceedings of the 2023 Conference on Empirical Methods in Natural Language Processing},
	publisher    = {Association for Computational Linguistics},
	address      = {Singapore},
	pages        = {7915--7927},
	doi          = {10.18653/v1/2023.emnlp-main.491},
	url          = {https://aclanthology.org/2023.emnlp-main.491/},
	editor       = {Bouamor, Houda  and Pino, Juan  and Bali, Kalika}
}

@inproceedings{saji-2025,
	title        = {{R}oman{L}ens: The Role Of Latent {R}omanization In Multilinguality In {LLM}s},
	author       = {Saji, Alan  and Husain, Jaavid Aktar  and Jayakumar, Thanmay  and Dabre, Raj  and Kunchukuttan, Anoop  and Puduppully, Ratish},
	year         = 2025,
	month        = jul,
	booktitle    = {Findings of the Association for Computational Linguistics: ACL 2025},
	publisher    = {Association for Computational Linguistics},
	address      = {Vienna, Austria},
	pages        = {26410--26429},
	doi          = {10.18653/v1/2025.findings-acl.1354},
	isbn         = {979-8-89176-256-5},
	url          = {https://aclanthology.org/2025.findings-acl.1354/},
	editor       = {Che, Wanxiang  and Nabende, Joyce  and Shutova, Ekaterina  and Pilehvar, Mohammad Taher}
}

@inproceedings{tezuka-2025,
	title        = {The Transfer Neurons Hypothesis: An Underlying Mechanism for Language Latent Space Transitions in Multilingual {LLM}s},
	author       = {Tezuka, Hinata  and Inoue, Naoya},
	year         = 2025,
	month        = nov,
	booktitle    = {Proceedings of the 2025 Conference on Empirical Methods in Natural Language Processing},
	publisher    = {Association for Computational Linguistics},
	address      = {Suzhou, China},
	pages        = {31742--31792},
	doi          = {10.18653/v1/2025.emnlp-main.1618},
	isbn         = {979-8-89176-332-6},
	url          = {https://aclanthology.org/2025.emnlp-main.1618/},
	editor       = {Christodoulopoulos, Christos  and Chakraborty, Tanmoy  and Rose, Carolyn  and Peng, Violet}
}

@article{grattafiori2024llama,
	title        = {The llama 3 herd of models},
	author       = {Grattafiori, Aaron and Dubey, Abhimanyu and Jauhri, Abhinav and Pandey, Abhinav and Kadian, Abhishek and Al-Dahle, Ahmad and Letman, Aiesha and Mathur, Akhil and Schelten, Alan and Vaughan, Alex and others},
	year         = 2024,
	journal      = {arXiv preprint arXiv:2407.21783}
}

@article{park2023linear,
	title        = {The linear representation hypothesis and the geometry of large language models},
	author       = {Park, Kiho and Choe, Yo Joong and Veitch, Victor},
	year         = 2023,
	journal      = {arXiv preprint arXiv:2311.03658}
}

@article{jiang2024origins,
	title        = {On the origins of linear representations in large language models},
	author       = {Jiang, Yibo and Rajendran, Goutham and Ravikumar, Pradeep and Aragam, Bryon and Veitch, Victor},
	year         = 2024,
	journal      = {arXiv preprint arXiv:2403.03867}
}

@inproceedings{ahuja2022calibration,
	title        = {On the calibration of massively multilingual language models},
	author       = {Ahuja, Kabir and Sitaram, Sunayana and Dandapat, Sandipan and Choudhury, Monojit},
	year         = 2022,
	booktitle    = {Proceedings of the 2022 Conference on Empirical Methods in Natural Language Processing},
	pages        = {4310--4323}
}

@article{liu2026earth,
	title        = {Who on Earth is using generative AI?},
	author       = {Liu, Yan and Wang, He},
	year         = 2026,
	journal      = {World Development},
	publisher    = {Elsevier},
	volume       = 199,
	pages        = 107260
}

@inproceedings{men2025shortgpt,
	title        = {Shortgpt: Layers in large language models are more redundant than you expect},
	author       = {Men, Xin and Xu, Mingyu and Zhang, Qingyu and Yuan, Qianhao and Wang, Bingning and Lin, Hongyu and Lu, Yaojie and Han, Xianpei and Chen, Weipeng},
	year         = 2025,
	booktitle    = {Findings of the Association for Computational Linguistics: ACL 2025},
	pages        = {20192--20204}
}

@article{geiger-2025,
	title        = {Causal abstraction: A theoretical foundation for mechanistic interpretability},
	author       = {Geiger, Atticus and Ibeling, Duligur and Zur, Amir and Chaudhary, Maheep and Chauhan, Sonakshi and Huang, Jing and Arora, Aryaman and Wu, Zhengxuan and Goodman, Noah and Potts, Christopher and others},
	year         = 2025,
	journal      = {Journal of Machine Learning Research},
	volume       = 26,
	number       = 83,
	pages        = {1--64}
}

@article{moreno2012unifying,
	title        = {A unifying view on dataset shift in classification},
	author       = {Moreno-Torres, Jose G and Raeder, Troy and Alaiz-Rodr{\'\i}guez, Roc{\'\i}o and Chawla, Nitesh V and Herrera, Francisco},
	year         = 2012,
	journal      = {Pattern recognition},
	publisher    = {Elsevier},
	volume       = 45,
	number       = 1,
	pages        = {521--530}
}

@article{liu2023simple,
	title        = {A simple approach to improve single-model deep uncertainty via distance-awareness},
	author       = {Liu, Jeremiah Zhe and Padhy, Shreyas and Ren, Jie and Lin, Zi and Wen, Yeming and Jerfel, Ghassen and Nado, Zachary and Snoek, Jasper and Tran, Dustin and Lakshminarayanan, Balaji},
	year         = 2023,
	journal      = {Journal of Machine Learning Research},
	volume       = 24,
	number       = 42,
	pages        = {1--63}
}

@article{ulmer2025anthropomimetic,
	title        = {Anthropomimetic uncertainty: What verbalized uncertainty in language models is missing},
	author       = {Ulmer, Dennis and Lorson, Alexandra and Titov, Ivan and Hardmeier, Christian},
	year         = 2025,
	journal      = {arXiv preprint arXiv:2507.10587}
}

@inproceedings{xiong2024can,
	title        = {Can llms express their uncertainty? an empirical evaluation of confidence elicitation in llms},
	author       = {Xiong, Miao and Hu, Zhiyuan and Lu, Xinyang and Li, Yifei and Fu, Jie and He, Junxian and Hooi, Bryan},
	year         = 2024,
	booktitle    = {International Conference on Learning Representations},
	volume       = 2024,
	pages        = {23650--23678}
}

@article{lin2022teaching,
	title        = {Teaching models to express their uncertainty in words},
	author       = {Lin, Stephanie and Hilton, Jacob and Evans, Owain},
	year         = 2022,
	journal      = {arXiv preprint arXiv:2205.14334}
}

@inproceedings{krause2023confidently,
	title        = {Confidently wrong: Exploring the calibration and expression of (un) certainty of large language models in a multilingual setting},
	author       = {Krause, Lea and Tufa, Wondimagegnhue and Santamar{\'\i}a, Selene B{\'a}ez and Daza, Angel and Khurana, Urja and Vossen, Piek},
	year         = 2023,
	booktitle    = {Proceedings of the workshop on multimodal, multilingual natural language generation and multilingual WebNLG Challenge (MM-NLG 2023)},
	pages        = {1--9}
}

@inproceedings{merullo2025linear,
	title        = {On linear representations and pretraining data frequency in language models},
	author       = {Merullo, Jack and Smith, Noah and Wiegreffe, Sarah and Elazar, Yanai},
	year         = 2025,
	booktitle    = {International Conference on Learning Representations},
	volume       = 2025,
	pages        = {71963--71987}
}

@misc{openai2025gpt41,
	title        = {Introducing {GPT-4.1} in the {API}},
	author       = {{OpenAI}},
	year         = 2025,
	month        = apr,
	day          = 14,
	note         = {Accessed: 2026-05-25},
	howpublished = {\url{https://openai.com/index/gpt-4-1/}}
}

@article{oron2017centered,
	title        = {Centered isotonic regression: point and interval estimation for dose--response studies},
	author       = {Oron, Assaf P and Flournoy, Nancy},
	year         = 2017,
	journal      = {Statistics in Biopharmaceutical Research},
	publisher    = {Taylor \& Francis},
	volume       = 9,
	number       = 3,
	pages        = {258--267}
}

@inproceedings{ji2025calibrating,
	title        = {Calibrating Verbal Uncertainty as a Linear Feature to Reduce Hallucinations},
	author       = {Ji, Ziwei and Yu, Lei and Koishekenov, Yeskendir and Bang, Yejin and Hartshorn, Anthony and Schelten, Alan and Zhang, Cheng and Fung, Pascale and Cancedda, Nicola},
	year         = 2025,
	booktitle    = {Proceedings of the 2025 Conference on Empirical Methods in Natural Language Processing},
	pages        = {3769--3793}
}

@article{glenn1950verification,
	title        = {Verification of forecasts expressed in terms of probability},
	author       = {Brier, Glenn W},
	year         = 1950,
	journal      = {Monthly weather review},
	publisher    = {War Department, Office of the Chief Signal Officer},
	volume       = 78,
	number       = 1,
	pages        = {1--3}
}

@inproceedings{qi2023cross,
	title        = {Cross-lingual consistency of factual knowledge in multilingual language models},
	author       = {Qi, Jirui and Fern{\'a}ndez, Raquel and Bisazza, Arianna},
	year         = 2023,
	booktitle    = {Proceedings of the 2023 Conference on Empirical Methods in Natural Language Processing},
	pages        = {10650--10666}
}

@inproceedings{fierro2022factual,
	title        = {Factual consistency of multilingual pretrained language models},
	author       = {Fierro, Constanza and S{\o}gaard, Anders},
	year         = 2022,
	booktitle    = {Findings of the Association for Computational Linguistics: ACL 2022},
	pages        = {3046--3052}
}

@inproceedings{zeng-2025,
	title        = {Converging to a Lingua Franca: Evolution of Linguistic Regions and Semantics Alignment in Multilingual Large Language Models},
	author       = {Zeng, Hongchuan  and Han, Senyu  and Chen, Lu  and Yu, Kai},
	year         = 2025,
	month        = jan,
	booktitle    = {Proceedings of the 31st International Conference on Computational Linguistics},
	publisher    = {Association for Computational Linguistics},
	address      = {Abu Dhabi, UAE},
	pages        = {10602--10617},
	url          = {https://aclanthology.org/2025.coling-main.707/},
	editor       = {Rambow, Owen  and Wanner, Leo  and Apidianaki, Marianna  and Al-Khalifa, Hend  and Eugenio, Barbara Di  and Schockaert, Steven}
}

@article{bereska-2024,
	title        = {Mechanistic Interpretability for {AI} Safety - {A} Review},
	author       = {Leonard Bereska and Stratis Gavves},
	year         = 2024,
	journal      = {Trans. Mach. Learn. Res.},
	volume       = 2024,
	url          = {https://openreview.net/forum?id=ePUVetPKu6},
	bibsource    = {dblp computer science bibliography, https://dblp.org}
}

@article{belinkov-2022,
	title        = {Probing Classifiers: Promises, Shortcomings, and Advances},
	author       = {Yonatan Belinkov},
	year         = 2022,
	journal      = {Comput. Linguistics},
	volume       = 48,
	number       = 1,
	pages        = {207--219},
	doi          = {10.1162/COLI\_A\_00422},
	url          = {https://doi.org/10.1162/coli\_a\_00422},
	bibsource    = {dblp computer science bibliography, https://dblp.org}
}

@inproceedings{alain-2017,
	title        = {Understanding intermediate layers using linear classifier probes},
	author       = {Guillaume Alain and Yoshua Bengio},
	year         = 2017,
	booktitle    = {5th International Conference on Learning Representations, {ICLR} 2017, Toulon, France, April 24-26, 2017, Workshop Track Proceedings},
	publisher    = {OpenReview.net},
	url          = {https://openreview.net/forum?id=HJ4-rAVtl},
	bibsource    = {dblp computer science bibliography, https://dblp.org}
}

@inproceedings{jin-2025,
	title        = {Exploring Concept Depth: How Large Language Models Acquire Knowledge and Concept at Different Layers?},
	author       = {Jin, Mingyu  and Yu, Qinkai  and Huang, Jingyuan  and Zeng, Qingcheng  and Wang, Zhenting  and Hua, Wenyue  and Zhao, Haiyan  and Mei, Kai  and Meng, Yanda  and Ding, Kaize  and Yang, Fan  and Du, Mengnan  and Zhang, Yongfeng},
	year         = 2025,
	month        = jan,
	booktitle    = {Proceedings of the 31st International Conference on Computational Linguistics},
	publisher    = {Association for Computational Linguistics},
	address      = {Abu Dhabi, UAE},
	pages        = {558--573},
	url          = {https://aclanthology.org/2025.coling-main.37/},
	editor       = {Rambow, Owen  and Wanner, Leo  and Apidianaki, Marianna  and Al-Khalifa, Hend  and Eugenio, Barbara Di  and Schockaert, Steven}
}

@inproceedings{ju-2024,
	title        = {How Large Language Models Encode Context Knowledge? A Layer-Wise Probing Study},
	author       = {Ju, Tianjie  and Sun, Weiwei  and Du, Wei  and Yuan, Xinwei  and Ren, Zhaochun  and Liu, Gongshen},
	year         = 2024,
	month        = may,
	booktitle    = {Proceedings of the 2024 Joint International Conference on Computational Linguistics, Language Resources and Evaluation (LREC-COLING 2024)},
	publisher    = {ELRA and ICCL},
	address      = {Torino, Italia},
	pages        = {8235--8246},
	url          = {https://aclanthology.org/2024.lrec-main.722/},
	editor       = {Calzolari, Nicoletta  and Kan, Min-Yen  and Hoste, Veronique  and Lenci, Alessandro  and Sakti, Sakriani  and Xue, Nianwen}
}

@article{ferrando-2024,
	title        = {A Primer on the Inner Workings of Transformer-based Language Models},
	author       = {Javier Ferrando and Gabriele Sarti and Arianna Bisazza and Marta R. Costa{-}juss{\`{a}}},
	year         = 2024,
	journal      = {CoRR},
	volume       = {abs/2405.00208},
	doi          = {10.48550/ARXIV.2405.00208},
	url          = {https://doi.org/10.48550/arXiv.2405.00208},
	eprinttype   = {arXiv},
	eprint       = {2405.00208},
	bibsource    = {dblp computer science bibliography, https://dblp.org}
}

@inproceedings{azaria-2023,
	title        = {The Internal State of an {LLM} Knows When It{'}s Lying},
	author       = {Azaria, Amos  and Mitchell, Tom},
	year         = 2023,
	month        = dec,
	booktitle    = {Findings of the Association for Computational Linguistics: EMNLP 2023},
	publisher    = {Association for Computational Linguistics},
	address      = {Singapore},
	pages        = {967--976},
	doi          = {10.18653/v1/2023.findings-emnlp.68},
	url          = {https://aclanthology.org/2023.findings-emnlp.68/},
	editor       = {Bouamor, Houda  and Pino, Juan  and Bali, Kalika}
}

@inproceedings{li-2023,
	title        = {Inference-time intervention: eliciting truthful answers from a language model},
	author       = {Li, Kenneth and Patel, Oam and Vi\'{e}gas, Fernanda and Pfister, Hanspeter and Wattenberg, Martin},
	year         = 2023,
	booktitle    = {Proceedings of the 37th International Conference on Neural Information Processing Systems},
	location     = {New Orleans, LA, USA},
	publisher    = {Curran Associates Inc.},
	address      = {Red Hook, NY, USA},
	series       = {NIPS '23},
	articleno    = 1797,
	numpages     = 80
}

@inproceedings{zhang2025atomic,
	title        = {Atomic calibration of llms in long-form generations},
	author       = {Zhang, Caiqi and Yang, Ruihan and Zhang, Zhisong and Huang, Xinting and Yang, Sen and Yu, Dong and Collier, Nigel},
	year         = 2025,
	booktitle    = {Proceedings of the 14th International Joint Conference on Natural Language Processing and the 4th Conference of the Asia-Pacific Chapter of the Association for Computational Linguistics},
	pages        = {148--169}
}

@article{fan2026iuq,
	title        = {IUQ: Interrogative Uncertainty Quantification for Long-Form Large Language Model Generation},
	author       = {Fan, Haozhi and Duan, Jinhao and Xu, Kaidi},
	year         = 2026,
	journal      = {arXiv preprint arXiv:2604.15109}
}

@article{bouchard2026fine,
	title        = {Fine-Grained Uncertainty Quantification for Long-Form Language Model Outputs: A Comparative Study},
	author       = {Bouchard, Dylan and Chauhan, Mohit Singh and Bajaj, Viren and Skarbrevik, David},
	year         = 2026,
	journal      = {arXiv preprint arXiv:2602.17431}
}

@article{zhang2025token,
	title        = {Token-level uncertainty estimation for large language model reasoning},
	author       = {Zhang, Tunyu and Shi, Haizhou and Wang, Yibin and Wang, Hengyi and He, Xiaoxiao and Li, Zhuowei and Chen, Haoxian and Han, Ligong and Xu, Kai and Zhang, Huan and others},
	year         = 2025,
	journal      = {arXiv e-prints},
	pages        = {arXiv--2505}
}

@article{yoon2025reasoning,
	title        = {Reasoning models better express their confidence},
	author       = {Yoon, Dongkeun and Kim, Seungone and Yang, Sohee and Kim, Sunkyoung and Kim, Soyeon and Kim, Yongil and Choi, Eunbi and Kim, Yireun and Seo, Minjoon},
	year         = 2025,
	journal      = {arXiv preprint arXiv:2505.14489}
}

@inproceedings{razghandi2025cer,
	title        = {Cer: Confidence enhanced reasoning in llms},
	author       = {Razghandi, Ali and Hosseini, Seyed Mohammad Hadi and Baghshah, Mahdieh Soleymani},
	year         = 2025,
	booktitle    = {Proceedings of the 63rd Annual Meeting of the Association for Computational Linguistics (Volume 1: Long Papers)},
	pages        = {7918--7938}
}

@inproceedings{zhang2025all,
	title        = {All roads lead to Rome: Graph-based confidence estimation for large language model reasoning},
	author       = {Zhang, Caiqi and Shu, Chang and Shareghi, Ehsan and Collier, Nigel},
	year         = 2025,
	booktitle    = {Proceedings of the 2025 Conference on Empirical Methods in Natural Language Processing},
	pages        = {31802--31812}
}

@article{mao2026confidence,
	title        = {Confidence over Time: Confidence Calibration with Temporal Logic for Large Language Model Reasoning},
	author       = {Mao, Zhenjiang and Venkat, Anirudhh and Bisliouk, Artem and Kothiyal, Akshat and Subramanian, Sindhura Kumbakonam and Singhu, Saithej and Ruchkin, Ivan},
	year         = 2026,
	journal      = {arXiv preprint arXiv:2601.13387}
}

@inproceedings{kudugunta2019investigating,
	title        = {Investigating multilingual NMT representations at scale},
	author       = {Kudugunta, Sneha and Bapna, Ankur and Caswell, Isaac and Firat, Orhan},
	year         = 2019,
	booktitle    = {Proceedings of the 2019 Conference on Empirical Methods in Natural Language Processing and the 9th International Joint Conference on Natural Language Processing (EMNLP-IJCNLP)},
	pages        = {1565--1575}
}

@inproceedings{singh2019bert,
	title        = {BERT is not an interlingua and the bias of tokenization},
	author       = {Singh, Jasdeep and McCann, Bryan and Socher, Richard and Xiong, Caiming},
	year         = 2019,
	booktitle    = {Proceedings of the 2nd Workshop on Deep Learning Approaches for Low-Resource NLP (DeepLo 2019)},
	pages        = {47--55}
}

@inproceedings{liu2025uncertainty,
	title        = {Uncertainty quantification and confidence calibration in large language models: A survey},
	author       = {Liu, Xiaoou and Chen, Tiejin and Da, Longchao and Chen, Chacha and Lin, Zhen and Wei, Hua},
	year         = 2025,
	booktitle    = {Proceedings of the 31st ACM SIGKDD Conference on Knowledge Discovery and Data Mining V. 2},
	pages        = {6107--6117}
}

@article{ulmer2024uncertainty,
	title        = {On uncertainty in natural language processing},
	author       = {Ulmer, Dennis},
	year         = 2024,
	journal      = {arXiv preprint arXiv:2410.03446}
}

@article{baan2023uncertainty,
	title        = {Uncertainty in natural language generation: From theory to applications},
	author       = {Baan, Joris and Daheim, Nico and Ilia, Evgenia and Ulmer, Dennis and Li, Haau-Sing and Fern{\'a}ndez, Raquel and Plank, Barbara and Sennrich, Rico and Zerva, Chrysoula and Aziz, Wilker},
	year         = 2023,
	journal      = {arXiv preprint arXiv:2307.15703}
}

@article{huang2024survey,
	title        = {A survey of uncertainty estimation in llms: Theory meets practice},
	author       = {Huang, Hsiu-Yuan and Yang, Yutong and Zhang, Zhaoxi and Lee, Sanwoo and Wu, Yunfang},
	year         = 2024,
	journal      = {arXiv preprint arXiv:2410.15326}
}

@article{shorinwa2025survey,
	title        = {A survey on uncertainty quantification of large language models: Taxonomy, open research challenges, and future directions},
	author       = {Shorinwa, Ola and Mei, Zhiting and Lidard, Justin and Ren, Allen Z and Majumdar, Anirudha},
	year         = 2025,
	journal      = {ACM Computing Surveys},
	publisher    = {ACM New York, NY},
	volume       = 58,
	number       = 3,
	pages        = {1--38}
}

@inproceedings{geng2024survey,
	title        = {A survey of confidence estimation and calibration in large language models},
	author       = {Geng, Jiahui and Cai, Fengyu and Wang, Yuxia and Koeppl, Heinz and Nakov, Preslav and Gurevych, Iryna},
	year         = 2024,
	booktitle    = {Proceedings of the 2024 Conference of the North American Chapter of the Association for Computational Linguistics: Human Language Technologies (Volume 1: Long Papers)},
	pages        = {6577--6595}
}

@inproceedings{xiong2024efficient,
	title        = {Efficient and effective uncertainty quantification for LLMs},
	author       = {Xiong, Miao and Santilli, Andrea and Kirchhof, Michael and Golinski, Adam and Williamson, Sinead},
	year         = 2024,
	booktitle    = {Neurips Safe Generative AI Workshop 2024}
}

@article{radford2019language,
	title        = {Language models are unsupervised multitask learners},
	author       = {Radford, Alec and Wu, Jeffrey and Child, Rewon and Luan, David and Amodei, Dario and Sutskever, Ilya and others},
	year         = 2019,
	journal      = {OpenAI blog},
	volume       = 1,
	number       = 8,
	pages        = 9
}

@inproceedings{shelmanov2025head,
	title        = {A head to predict and a head to question: Pre-trained uncertainty quantification heads for hallucination detection in llm outputs},
	author       = {Shelmanov, Artem and Fadeeva, Ekaterina and Tsvigun, Akim and Tsvigun, Ivan and Xie, Zhuohan and Kiselev, Igor and Daheim, Nico and Zhang, Caiqi and Vazhentsev, Artem and Sachan, Mrinmaya and others},
	year         = 2025,
	booktitle    = {Proceedings of the 2025 Conference on Empirical Methods in Natural Language Processing},
	pages        = {35700--35719}
}

@inproceedings{fathullah2024needs,
	title        = {Who needs decoders? efficient estimation of sequence-level attributes with proxies},
	author       = {Fathullah, Yassir and Radmard, Puria and Liusie, Adian and Gales, Mark},
	year         = 2024,
	booktitle    = {Proceedings of the 18th Conference of the European Chapter of the Association for Computational Linguistics (Volume 1: Long Papers)},
	pages        = {1478--1496}
}

@article{liu2024uncertainty,
	title        = {Uncertainty estimation and quantification for llms: A simple supervised approach},
	author       = {Liu, Linyu and Pan, Yu and Li, Xiaocheng and Chen, Guanting},
	year         = 2024,
	journal      = {arXiv preprint arXiv:2404.15993}
}

@inproceedings{liu2024litcablightweightlanguagemodel,
	title        = {LitCab: Lightweight Language Model Calibration over Short- and Long-form Responses},
	author       = {Xin Liu and Muhammad Khalifa and Lu Wang},
	year         = 2024,
	booktitle    = {Proceedings of the Twelfth International Conference on Learning Representations},
	publisher    = {ICLR},
	url          = {https://arxiv.org/abs/2310.19208}
}

@inproceedings{burns2023discovering,
	title        = {Discovering Latent Knowledge in Language Models Without Supervision},
	author       = {Collin Burns and Haotian Ye and Dan Klein and Jacob Steinhardt},
	year         = 2023,
	booktitle    = {The Eleventh International Conference on Learning Representations},
	url          = {https://openreview.net/forum?id=ETKGuby0hcs}
}

@inproceedings{ulmer2022exploring,
	title        = {Exploring predictive uncertainty and calibration in NLP: A study on the impact of method \& data scarcity},
	author       = {Ulmer, Dennis and Frellsen, Jes and Hardmeier, Christian},
	year         = 2022,
	booktitle    = {Findings of the Association for Computational Linguistics: EMNLP 2022},
	pages        = {2707--2735}
}

@article{yang2023calibration,
	title        = {On the calibration of multilingual question answering llms},
	author       = {Yang, Yahan and Dan, Soham and Roth, Dan and Lee, Insup},
	year         = 2023,
	journal      = {arXiv preprint arXiv:2311.08669}
}

@inproceedings{xue2025mlingconf,
	title        = {Mlingconf: A comprehensive study of multilingual confidence estimation on large language models},
	author       = {Xue, Boyang and Wang, Hongru and Wang, Rui and Wang, Sheng and Wang, Zezhong and Du, Yiming and Liang, Bin and Zhang, Wenxuan and Wong, Kam-Fai},
	year         = 2025,
	booktitle    = {Findings of the Association for Computational Linguistics: ACL 2025},
	pages        = {2535--2556}
}

@article{zhou2025beyond,
	title        = {Beyond the final layer: Intermediate representations for better multilingual calibration in large language models},
	author       = {Zhou, Ej and Zhang, Caiqi and Hu, Tiancheng and Li, Chengzu and Collier, Nigel and Vuli{\'c}, Ivan and Korhonen, Anna},
	year         = 2025,
	journal      = {arXiv preprint arXiv:2510.03136}
}

@article{sun-2025,
	title        = {The Curse of Depth in Large Language Models},
	author       = {Wenfang Sun and Xinyuan Song and Pengxiang Li and Lu Yin and Yefeng Zheng and Shiwei Liu},
	year         = 2025,
	journal      = {CoRR},
	volume       = {abs/2502.05795},
	doi          = {10.48550/ARXIV.2502.05795},
	url          = {https://doi.org/10.48550/arXiv.2502.05795},
	eprinttype   = {arXiv},
	eprint       = {2502.05795},
	bibsource    = {dblp computer science bibliography, https://dblp.org}
}

@inproceedings{liu-2024,
    title = "Unraveling {B}abel: Exploring Multilingual Activation Patterns of {LLM}s and Their Applications",
    author = "Liu, Weize  and
      Xu, Yinlong  and
      Xu, Hongxia  and
      Chen, Jintai  and
      Hu, Xuming  and
      Wu, Jian",
    editor = "Al-Onaizan, Yaser  and
      Bansal, Mohit  and
      Chen, Yun-Nung",
    booktitle = "Proceedings of the 2024 Conference on Empirical Methods in Natural Language Processing",
    month = nov,
    year = "2024",
    address = "Miami, Florida, USA",
    publisher = "Association for Computational Linguistics",
    url = "https://aclanthology.org/2024.emnlp-main.662/",
    doi = "10.18653/v1/2024.emnlp-main.662",
    pages = "11855--11881"
}

@inproceedings{khanmohammadi-etal-2025-calibrating,
    title = "Calibrating {LLM} Confidence by Probing Perturbed Representation Stability",
    author = "Khanmohammadi, Reza  and
      Miahi, Erfan  and
      Mardikoraem, Mehrsa  and
      Kaur, Simerjot  and
      Brugere, Ivan  and
      Smiley, Charese  and
      Thind, Kundan S  and
      Ghassemi, Mohammad M.",
    editor = "Christodoulopoulos, Christos  and
      Chakraborty, Tanmoy  and
      Rose, Carolyn  and
      Peng, Violet",
    booktitle = "Proceedings of the 2025 Conference on Empirical Methods in Natural Language Processing",
    month = nov,
    year = "2025",
    address = "Suzhou, China",
    publisher = "Association for Computational Linguistics",
    url = "https://aclanthology.org/2025.emnlp-main.530/",
    doi = "10.18653/v1/2025.emnlp-main.530",
    pages = "10448--10514",
    ISBN = "979-8-89176-332-6"
}

@inproceedings{bao-etal-2025-probing,
    title = "Probing the Geometry of Truth: Consistency and Generalization of Truth Directions in {LLM}s Across Logical Transformations and Question Answering Tasks",
    author = "Bao, Yuntai  and
      Zhang, Xuhong  and
      Du, Tianyu  and
      Zhao, Xinkui  and
      Feng, Zhengwen  and
      Peng, Hao  and
      Yin, Jianwei",
    editor = "Che, Wanxiang  and
      Nabende, Joyce  and
      Shutova, Ekaterina  and
      Pilehvar, Mohammad Taher",
    booktitle = "Findings of the Association for Computational Linguistics: ACL 2025",
    month = jul,
    year = "2025",
    address = "Vienna, Austria",
    publisher = "Association for Computational Linguistics",
    url = "https://aclanthology.org/2025.findings-acl.38/",
    doi = "10.18653/v1/2025.findings-acl.38",
    pages = "682--700",
    ISBN = "979-8-89176-256-5"
}

@inproceedings{boyd-2013,
  author       = {Kendrick Boyd and
                  Kevin H. Eng and
                  C. David Page Jr.},
  editor       = {Hendrik Blockeel and
                  Kristian Kersting and
                  Siegfried Nijssen and
                  Filip Zelezn{\'{y}}},
  title        = {Area under the Precision-Recall Curve: Point Estimates and Confidence
                  Intervals},
  booktitle    = {Machine Learning and Knowledge Discovery in Databases - European Conference,
                  {ECML} {PKDD} 2013, Prague, Czech Republic, September 23-27, 2013,
                  Proceedings, Part {III}},
  series       = {Lecture Notes in Computer Science},
  volume       = {8190},
  pages        = {451--466},
  publisher    = {Springer},
  year         = {2013},
  url          = {https://doi.org/10.1007/978-3-642-40994-3\_29},
  doi          = {10.1007/978-3-642-40994-3\_29},
  bibsource    = {dblp computer science bibliography, https://dblp.org}
}

\appendix

\section{Additional Results}

\subsection{Confidence Estimation Results}\label{app:additional-ce-results}

\begin{figure*}[htb]
    \centering
    \includegraphics[width=0.985\textwidth]{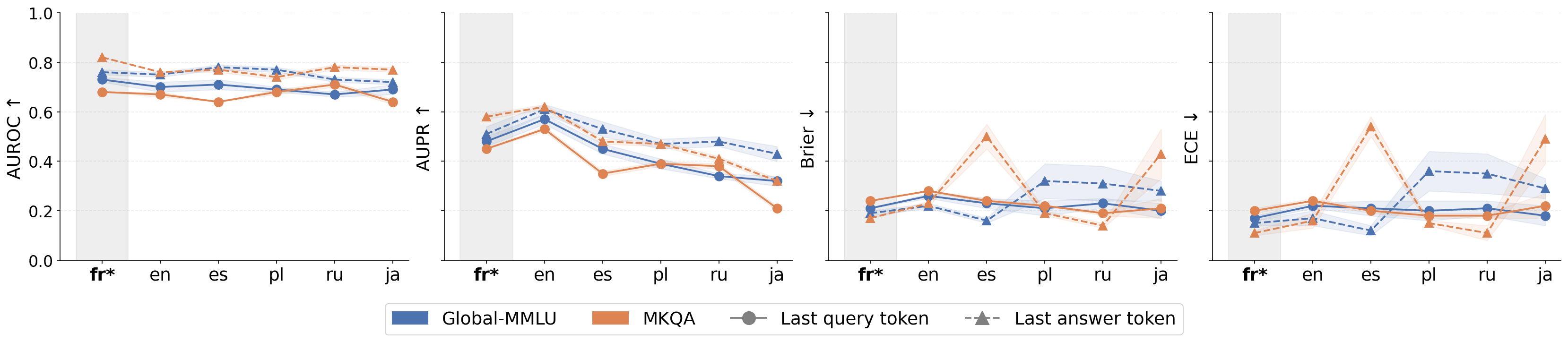}
    \caption{Cross-lingual generalization of the probe trained on French (fr*) for Llama 3.1 8B. Each panel compares the last query token (circle, solid) against the last answer token (triangle, dashed) across test languages, for Global-MMLU (blue) and MKQA (orange). Shaded bands show ±1 standard deviation.}\label{fig:zero-shot-llama}
\end{figure*}

\begin{table*}[tb]
\centering
\ra{1.3}
\renewcommand{\arraystretch}{1.5}
\setlength{\tabcolsep}{4pt}
\footnotesize
\resizebox{\textwidth}{!}{%
\begin{tabular}{llcccccccccc}
\toprule
 & & & \multicolumn{4}{c}{\textbf{MKQA}} & & \multicolumn{4}{c}{\textbf{Global-MMLU}} \\
\cmidrule(lr){4-7} \cmidrule(lr){9-12}
 & & & AUROC$\uparrow$ & AUPR$\uparrow$ & Brier$\downarrow$ & ECE$\downarrow$ & & AUROC$\uparrow$ & AUPR$\uparrow$ & Brier$\downarrow$ & ECE$\downarrow$ \\
\midrule
\multirow{2}{*}{\rotatebox{90}{all*}} & Majority & & .50$\pm$.00 & .62$\pm$.04 & .24$\pm$.08 & .24$\pm$.08 &  & .50$\pm$.00 & .62$\pm$.03 & .23$\pm$.06 & .23$\pm$.06 \\
 & Prior prob. & & .50$\pm$.00 & .62$\pm$.04 & .18$\pm$.04 & .07$\pm$.06 &  & .50$\pm$.00 & .62$\pm$.03 & .18$\pm$.03 & .05$\pm$.04 \\
\multirow{2}{*}{\rotatebox{90}{fr}} & Last Query & & .68$\pm$.00 & .45$\pm$.00 & .24$\pm$.00 & .20$\pm$.01 &  & .73$\pm$.01 & .48$\pm$.02 & .21$\pm$.00 & .17$\pm$.01 \\
 & Last Answer & & .82$\pm$.00 & .58$\pm$.01 & .17$\pm$.00 & .11$\pm$.01 &  & .76$\pm$.01 & .51$\pm$.03 & .19$\pm$.01 & .15$\pm$.01 \\
\midrule
\multirow{2}{*}{\rotatebox{90}{en}} & Last Query & & .67$\pm$.01 \textcolor{gray}{+{.02}} & .53$\pm$.01 \textcolor{gray}{+{.00}} & .28$\pm$.00 \textcolor{gray}{+{.00}} & .24$\pm$.01 \textcolor{gray}{+{.02}} &  & .70$\pm$.02 \textcolor{gray}{+{.01}} & .57$\pm$.02 \textcolor{gray}{+{.05}} & .26$\pm$.01 \textcolor{gray}{+{.00}} & .22$\pm$.01 \textcolor{gray}{+{.00}} \\
 & Last Answer & & .76$\pm$.00 \textcolor{gray}{-{.02}} & .62$\pm$.01 \textcolor{gray}{-{.05}} & .23$\pm$.01 \textcolor{gray}{+{.03}} & .16$\pm$.03 \textcolor{gray}{+{.03}} &  & .75$\pm$.01 \textcolor{gray}{-{.02}} & .61$\pm$.02 \textcolor{gray}{-{.05}} & .22$\pm$.01 \textcolor{gray}{+{.01}} & .17$\pm$.03 \textcolor{gray}{+{.01}} \\
\cmidrule(lr){2-12}
\multirow{2}{*}{\rotatebox{90}{es}} & Last Query & & .64$\pm$.00 \textcolor{gray}{+{.04}} & .35$\pm$.01 \textcolor{gray}{+{.00}} & .24$\pm$.01 \textcolor{gray}{-{.03}} & .20$\pm$.01 \textcolor{gray}{-{.04}} &  & .71$\pm$.02 \textcolor{gray}{+{.05}} & .45$\pm$.02 \textcolor{gray}{+{.07}} & .23$\pm$.02 \textcolor{gray}{+{.00}} & .21$\pm$.03 \textcolor{gray}{+{.01}} \\
 & Last Answer & & .77$\pm$.01 \textcolor{gray}{-{.04}} & .48$\pm$.01 \textcolor{gray}{-{.10}} & .50$\pm$.05 \textcolor{gray}{+{.33}} & .54$\pm$.04 \textcolor{gray}{+{.42}} &  & .78$\pm$.01 \textcolor{gray}{+{.00}} & .53$\pm$.03 \textcolor{gray}{-{.04}} & .16$\pm$.01 \textcolor{gray}{-{.01}} & .12$\pm$.02 \textcolor{gray}{-{.01}} \\
\cmidrule(lr){2-12}
\multirow{2}{*}{\rotatebox{90}{pl}} & Last Query & & .68$\pm$.01 \textcolor{gray}{+{.00}} & .39$\pm$.01 \textcolor{gray}{-{.03}} & .22$\pm$.00 \textcolor{gray}{-{.01}} & .18$\pm$.01 \textcolor{gray}{-{.01}} &  & .69$\pm$.01 \textcolor{gray}{-{.01}} & .39$\pm$.02 \textcolor{gray}{+{.03}} & .21$\pm$.03 \textcolor{gray}{+{.03}} & .20$\pm$.04 \textcolor{gray}{+{.04}} \\
 & Last Answer & & .74$\pm$.01 \textcolor{gray}{-{.05}} & .47$\pm$.01 \textcolor{gray}{-{.09}} & .19$\pm$.01 \textcolor{gray}{+{.01}} & .15$\pm$.01 \textcolor{gray}{+{.02}} &  & .77$\pm$.01 \textcolor{gray}{+{.01}} & .47$\pm$.02 \textcolor{gray}{+{.01}} & .32$\pm$.07 \textcolor{gray}{+{.16}} & .36$\pm$.08 \textcolor{gray}{+{.22}} \\
\cmidrule(lr){2-12}
\multirow{2}{*}{\rotatebox{90}{ru}} & Last Query & & .71$\pm$.01 \textcolor{gray}{+{.05}} & .38$\pm$.01 \textcolor{gray}{+{.05}} & .19$\pm$.01 \textcolor{gray}{+{.00}} & .18$\pm$.01 \textcolor{gray}{+{.00}} &  & .67$\pm$.01 \textcolor{gray}{-{.09}} & .34$\pm$.01 \textcolor{gray}{-{.11}} & .23$\pm$.02 \textcolor{gray}{+{.06}} & .21$\pm$.03 \textcolor{gray}{+{.07}} \\
 & Last Answer & & .78$\pm$.01 \textcolor{gray}{-{.05}} & .41$\pm$.01 \textcolor{gray}{-{.13}} & .14$\pm$.01 \textcolor{gray}{+{.00}} & .11$\pm$.03 \textcolor{gray}{+{.00}} &  & .73$\pm$.01 \textcolor{gray}{-{.06}} & .48$\pm$.02 \textcolor{gray}{-{.05}} & .31$\pm$.07 \textcolor{gray}{+{.15}} & .35$\pm$.08 \textcolor{gray}{+{.22}} \\
\cmidrule(lr){2-12}
\multirow{2}{*}{\rotatebox{90}{ja}} & Last Query & & .64$\pm$.01 \textcolor{gray}{-{.09}} & .21$\pm$.01 \textcolor{gray}{-{.08}} & .21$\pm$.04 \textcolor{gray}{+{.05}} & .22$\pm$.05 \textcolor{gray}{+{.08}} &  & .69$\pm$.02 \textcolor{gray}{+{.05}} & .32$\pm$.02 \textcolor{gray}{+{.06}} & .20$\pm$.03 \textcolor{gray}{+{.01}} & .18$\pm$.04 \textcolor{gray}{+{.01}} \\
 & Last Answer & & .77$\pm$.01 \textcolor{gray}{-{.03}} & .32$\pm$.01 \textcolor{gray}{-{.10}} & .43$\pm$.10 \textcolor{gray}{+{.30}} & .49$\pm$.10 \textcolor{gray}{+{.39}} &  & .72$\pm$.01 \textcolor{gray}{-{.01}} & .43$\pm$.03 \textcolor{gray}{+{.02}} & .28$\pm$.04 \textcolor{gray}{+{.11}} & .29$\pm$.04 \textcolor{gray}{+{.13}} \\
\bottomrule
\end{tabular}%
}
\caption{Results for Llama 3.1 8B on MKQA and Global-MMLU. 
First group shows the results for the probe and baselines trained and tested on French. all* = avg.\@ across all languages. Remaining groups list zero-shot performance on test languages. Last Query / Last Answer = Probe trained on last query / answer hiddens. Gray numbers state performance difference compared to the oracle probe, i.e. $\Delta < 0$ indicating that the oracle scored higher.}
\label{tab:confidence-estimates-llama}
\end{table*}

\begin{table*}[tb]
\centering
\ra{1.3}
\renewcommand{\arraystretch}{1.5}
\setlength{\tabcolsep}{4pt}
\footnotesize
\resizebox{\textwidth}{!}{%
\begin{tabular}{llcccccccccc}
\toprule
 & & & \multicolumn{4}{c}{\textbf{MKQA}} & & \multicolumn{4}{c}{\textbf{Global-MMLU}} \\
\cmidrule(lr){4-7} \cmidrule(lr){9-12}
 & & & AUROC$\uparrow$ & AUPR$\uparrow$ & Brier$\downarrow$ & ECE$\downarrow$ & & AUROC$\uparrow$ & AUPR$\uparrow$ & Brier$\downarrow$ & ECE$\downarrow$ \\
\midrule
\multirow{2}{*}{\rotatebox{90}{en}} & Last Query & & .65$\pm$.01 & .53$\pm$.01 & .28$\pm$.00 & .22$\pm$.01 &  & .69$\pm$.01 & .52$\pm$.01 & .26$\pm$.01 & .22$\pm$.01 \\
 & Last Answer & & .78$\pm$.01 & .67$\pm$.01 & .20$\pm$.00 & .13$\pm$.01 &  & .77$\pm$.01 & .66$\pm$.01 & .21$\pm$.01 & .16$\pm$.02 \\
\midrule
\multirow{2}{*}{\rotatebox{90}{fr}} & Last Query & & .63$\pm$.01 \textcolor{gray}{-{.05}} & .34$\pm$.01 \textcolor{gray}{-{.11}} & .27$\pm$.01 \textcolor{gray}{+{.03}} & .22$\pm$.00 \textcolor{gray}{+{.02}} &  & .66$\pm$.01 \textcolor{gray}{-{.07}} & .35$\pm$.02 \textcolor{gray}{-{.13}} & .23$\pm$.02 \textcolor{gray}{+{.02}} & .21$\pm$.03 \textcolor{gray}{+{.04}} \\
 & Last Answer & & .76$\pm$.01 \textcolor{gray}{-{.06}} & .56$\pm$.01 \textcolor{gray}{-{.02}} & .24$\pm$.04 \textcolor{gray}{+{.07}} & .23$\pm$.06 \textcolor{gray}{+{.12}} &  & .74$\pm$.02 \textcolor{gray}{-{.02}} & .46$\pm$.03 \textcolor{gray}{-{.05}} & .22$\pm$.03 \textcolor{gray}{+{.03}} & .20$\pm$.05 \textcolor{gray}{+{.05}} \\
\cmidrule(lr){2-12}
\multirow{2}{*}{\rotatebox{90}{es}} & Last Query & & .67$\pm$.01 \textcolor{gray}{+{.07}} & .36$\pm$.01 \textcolor{gray}{+{.01}} & .25$\pm$.01 \textcolor{gray}{-{.02}} & .21$\pm$.01 \textcolor{gray}{-{.03}} &  & .67$\pm$.01 \textcolor{gray}{+{.01}} & .36$\pm$.02 \textcolor{gray}{-{.02}} & .23$\pm$.01 \textcolor{gray}{+{.00}} & .19$\pm$.02 \textcolor{gray}{-{.01}} \\
 & Last Answer & & .74$\pm$.01 \textcolor{gray}{-{.07}} & .48$\pm$.02 \textcolor{gray}{-{.10}} & .22$\pm$.03 \textcolor{gray}{+{.05}} & .20$\pm$.05 \textcolor{gray}{+{.08}} &  & .74$\pm$.02 \textcolor{gray}{-{.04}} & .48$\pm$.02 \textcolor{gray}{-{.09}} & .19$\pm$.02 \textcolor{gray}{+{.02}} & .15$\pm$.03 \textcolor{gray}{+{.02}} \\
\cmidrule(lr){2-12}
\multirow{2}{*}{\rotatebox{90}{pl}} & Last Query & & .69$\pm$.01 \textcolor{gray}{+{.01}} & .42$\pm$.01 \textcolor{gray}{+{.00}} & .24$\pm$.01 \textcolor{gray}{+{.01}} & .21$\pm$.02 \textcolor{gray}{+{.02}} &  & .64$\pm$.02 \textcolor{gray}{-{.06}} & .27$\pm$.02 \textcolor{gray}{-{.09}} & .27$\pm$.05 \textcolor{gray}{+{.09}} & .26$\pm$.06 \textcolor{gray}{+{.10}} \\
 & Last Answer & & .72$\pm$.01 \textcolor{gray}{-{.07}} & .50$\pm$.02 \textcolor{gray}{-{.06}} & .40$\pm$.04 \textcolor{gray}{+{.22}} & .41$\pm$.05 \textcolor{gray}{+{.28}} &  & .71$\pm$.02 \textcolor{gray}{-{.05}} & .38$\pm$.03 \textcolor{gray}{-{.08}} & .16$\pm$.01 \textcolor{gray}{+{.00}} & .14$\pm$.02 \textcolor{gray}{+{.00}} \\
\cmidrule(lr){2-12}
\multirow{2}{*}{\rotatebox{90}{ru}} & Last Query & & .69$\pm$.01 \textcolor{gray}{+{.03}} & .33$\pm$.01 \textcolor{gray}{+{.00}} & .28$\pm$.02 \textcolor{gray}{+{.09}} & .28$\pm$.02 \textcolor{gray}{+{.10}} &  & .65$\pm$.01 \textcolor{gray}{-{.11}} & .26$\pm$.01 \textcolor{gray}{-{.19}} & .22$\pm$.02 \textcolor{gray}{+{.05}} & .20$\pm$.03 \textcolor{gray}{+{.06}} \\
 & Last Answer & & .74$\pm$.01 \textcolor{gray}{-{.09}} & .40$\pm$.02 \textcolor{gray}{-{.14}} & .47$\pm$.04 \textcolor{gray}{+{.33}} & .52$\pm$.04 \textcolor{gray}{+{.41}} &  & .73$\pm$.01 \textcolor{gray}{-{.06}} & .42$\pm$.02 \textcolor{gray}{-{.11}} & .15$\pm$.01 \textcolor{gray}{-{.01}} & .11$\pm$.02 \textcolor{gray}{-{.02}} \\
\cmidrule(lr){2-12}
\multirow{2}{*}{\rotatebox{90}{ja}} & Last Query & & .64$\pm$.01 \textcolor{gray}{-{.09}} & .24$\pm$.01 \textcolor{gray}{-{.05}} & .59$\pm$.07 \textcolor{gray}{+{.43}} & .65$\pm$.06 \textcolor{gray}{+{.51}} &  & .62$\pm$.02 \textcolor{gray}{-{.02}} & .23$\pm$.02 \textcolor{gray}{-{.03}} & .52$\pm$.08 \textcolor{gray}{+{.33}} & .55$\pm$.08 \textcolor{gray}{+{.38}} \\
 & Last Answer & & .64$\pm$.01 \textcolor{gray}{-{.16}} & .32$\pm$.03 \textcolor{gray}{-{.10}} & .67$\pm$.03 \textcolor{gray}{+{.54}} & .68$\pm$.03 \textcolor{gray}{+{.58}} &  & .71$\pm$.02 \textcolor{gray}{-{.02}} & .33$\pm$.04 \textcolor{gray}{-{.08}} & .26$\pm$.05 \textcolor{gray}{+{.09}} & .27$\pm$.07 \textcolor{gray}{+{.11}} \\
\bottomrule
\end{tabular}%
}
\caption{Results for Llama 3.1 8B on MKQA and Global-MMLU, with English as the source language.
First group shows the results for the probe and baselines trained and tested on English. Remaining groups list zero-shot performance on test languages. Last Query / Last Answer = Probe trained on last query / answer hiddens. Gray numbers state performance difference compared to the oracle probe, i.e. $\Delta < 0$ indicating that the oracle scored higher.}
\label{tab:confidence-estimates-llama-en}
\end{table*}

\begin{table*}[tb]
\centering
\ra{1.3}
\renewcommand{\arraystretch}{1.5}
\setlength{\tabcolsep}{4pt}
\footnotesize
\resizebox{\textwidth}{!}{%
\begin{tabular}{llcccccccccc}
\toprule
 & & & \multicolumn{4}{c}{\textbf{MKQA}} & & \multicolumn{4}{c}{\textbf{Global-MMLU}} \\
\cmidrule(lr){4-7} \cmidrule(lr){9-12}
 & & & AUROC$\uparrow$ & AUPR$\uparrow$ & Brier$\downarrow$ & ECE$\downarrow$ & & AUROC$\uparrow$ & AUPR$\uparrow$ & Brier$\downarrow$ & ECE$\downarrow$ \\
\midrule
\multirow{2}{*}{\rotatebox{90}{en}} & Last Query & & .70$\pm$.01 & .44$\pm$.01 & .23$\pm$.01 & .18$\pm$.01 &  & .71$\pm$.01 & .60$\pm$.01 & .25$\pm$.01 & .19$\pm$.01 \\
 & Last Answer & & .78$\pm$.01 & .57$\pm$.01 & .18$\pm$.01 & .13$\pm$.01 &  & .76$\pm$.01 & .69$\pm$.01 & .23$\pm$.01 & .19$\pm$.01 \\
\midrule
\multirow{2}{*}{\rotatebox{90}{fr}} & Last Query & & .68$\pm$.01 \textcolor{gray}{-{.05}} & .31$\pm$.01 \textcolor{gray}{-{.05}} & .18$\pm$.01 \textcolor{gray}{+{.01}} & .15$\pm$.01 \textcolor{gray}{+{.01}} &  & .66$\pm$.01 \textcolor{gray}{-{.05}} & .41$\pm$.01 \textcolor{gray}{-{.08}} & .25$\pm$.03 \textcolor{gray}{+{.04}} & .20$\pm$.05 \textcolor{gray}{+{.04}} \\
 & Last Answer & & .77$\pm$.01 \textcolor{gray}{-{.05}} & .44$\pm$.02 \textcolor{gray}{-{.05}} & .16$\pm$.02 \textcolor{gray}{+{.02}} & .14$\pm$.03 \textcolor{gray}{+{.02}} &  & .80$\pm$.01 \textcolor{gray}{+{.02}} & .63$\pm$.02 \textcolor{gray}{+{.05}} & .20$\pm$.01 \textcolor{gray}{+{.01}} & .18$\pm$.02 \textcolor{gray}{+{.02}} \\
\cmidrule(lr){2-12}
\multirow{2}{*}{\rotatebox{90}{es}} & Last Query & & .71$\pm$.01 \textcolor{gray}{-{.01}} & .32$\pm$.02 \textcolor{gray}{-{.04}} & .19$\pm$.01 \textcolor{gray}{+{.01}} & .16$\pm$.01 \textcolor{gray}{+{.00}} &  & .71$\pm$.01 \textcolor{gray}{+{.03}} & .44$\pm$.01 \textcolor{gray}{+{.00}} & .23$\pm$.02 \textcolor{gray}{+{.01}} & .19$\pm$.03 \textcolor{gray}{+{.01}} \\
 & Last Answer & & .79$\pm$.01 \textcolor{gray}{-{.04}} & .47$\pm$.02 \textcolor{gray}{-{.05}} & .23$\pm$.05 \textcolor{gray}{+{.09}} & .25$\pm$.07 \textcolor{gray}{+{.14}} &  & .77$\pm$.01 \textcolor{gray}{+{.00}} & .61$\pm$.01 \textcolor{gray}{+{.06}} & .22$\pm$.02 \textcolor{gray}{+{.02}} & .21$\pm$.03 \textcolor{gray}{+{.04}} \\
\cmidrule(lr){2-12}
\multirow{2}{*}{\rotatebox{90}{pl}} & Last Query & & .66$\pm$.01 \textcolor{gray}{-{.07}} & .33$\pm$.01 \textcolor{gray}{-{.01}} & .18$\pm$.01 \textcolor{gray}{+{.02}} & .16$\pm$.02 \textcolor{gray}{+{.02}} &  & .66$\pm$.01 \textcolor{gray}{-{.06}} & .33$\pm$.01 \textcolor{gray}{-{.09}} & .21$\pm$.01 \textcolor{gray}{+{.00}} & .20$\pm$.01 \textcolor{gray}{+{.03}} \\
 & Last Answer & & .74$\pm$.01 \textcolor{gray}{-{.08}} & .47$\pm$.01 \textcolor{gray}{-{.01}} & .20$\pm$.02 \textcolor{gray}{+{.07}} & .21$\pm$.03 \textcolor{gray}{+{.10}} &  & .76$\pm$.01 \textcolor{gray}{-{.03}} & .46$\pm$.02 \textcolor{gray}{-{.06}} & .26$\pm$.03 \textcolor{gray}{+{.08}} & .27$\pm$.03 \textcolor{gray}{+{.11}} \\
\cmidrule(lr){2-12}
\multirow{2}{*}{\rotatebox{90}{ru}} & Last Query & & .70$\pm$.01 \textcolor{gray}{+{.01}} & .25$\pm$.02 \textcolor{gray}{-{.03}} & .14$\pm$.02 \textcolor{gray}{-{.02}} & .14$\pm$.02 \textcolor{gray}{-{.02}} &  & .64$\pm$.01 \textcolor{gray}{-{.09}} & .31$\pm$.01 \textcolor{gray}{-{.15}} & .22$\pm$.01 \textcolor{gray}{+{.03}} & .19$\pm$.01 \textcolor{gray}{+{.03}} \\
 & Last Answer & & .70$\pm$.02 \textcolor{gray}{-{.09}} & .36$\pm$.02 \textcolor{gray}{-{.07}} & .21$\pm$.03 \textcolor{gray}{+{.09}} & .25$\pm$.04 \textcolor{gray}{+{.14}} &  & .77$\pm$.01 \textcolor{gray}{-{.03}} & .52$\pm$.01 \textcolor{gray}{-{.04}} & .17$\pm$.01 \textcolor{gray}{+{.00}} & .15$\pm$.01 \textcolor{gray}{+{.01}} \\
\cmidrule(lr){2-12}
\multirow{2}{*}{\rotatebox{90}{ja}} & Last Query & & .61$\pm$.03 \textcolor{gray}{-{.17}} & .16$\pm$.01 \textcolor{gray}{-{.17}} & .18$\pm$.03 \textcolor{gray}{+{.05}} & .17$\pm$.05 \textcolor{gray}{+{.06}} &  & .65$\pm$.02 \textcolor{gray}{-{.09}} & .32$\pm$.02 \textcolor{gray}{-{.13}} & .20$\pm$.03 \textcolor{gray}{+{.01}} & .15$\pm$.06 \textcolor{gray}{-{.02}} \\
 & Last Answer & & .78$\pm$.01 \textcolor{gray}{-{.09}} & .35$\pm$.02 \textcolor{gray}{-{.18}} & .19$\pm$.03 \textcolor{gray}{+{.10}} & .20$\pm$.04 \textcolor{gray}{+{.13}} &  & .76$\pm$.01 \textcolor{gray}{-{.03}} & .45$\pm$.03 \textcolor{gray}{-{.07}} & .23$\pm$.03 \textcolor{gray}{+{.05}} & .22$\pm$.03 \textcolor{gray}{+{.06}} \\
\bottomrule
\end{tabular}%
}
\caption{Results for Qwen3  8B on MKQA and Global-MMLU, with English as the source language.
First group shows the results for the probe and baselines trained and tested on English. Remaining groups list zero-shot performance on test languages. Last Query / Last Answer = Probe trained on last query / answer hiddens. Gray numbers state performance difference compared to the oracle probe, i.e. $\Delta < 0$ indicating that the oracle scored higher.}
\label{tab:confidence-estimates-qwen-en}
\end{table*}

This section contains complete results for the confidence estimation in \cref{sec:zero-shot-ce}.
\subsubsection{French as Source Language} \label{app:conf-est-french-source}
\cref{tab:confidence-estimates-llama} shows the results with French as the source language for Llama 3.1. 

\subsubsection{All Source Languages} \label{app:conf-est-all-sources}

\begin{figure}[htb]
    \centering
    \begin{subfigure}{\linewidth}
        \centering
        \includegraphics[width=\linewidth]{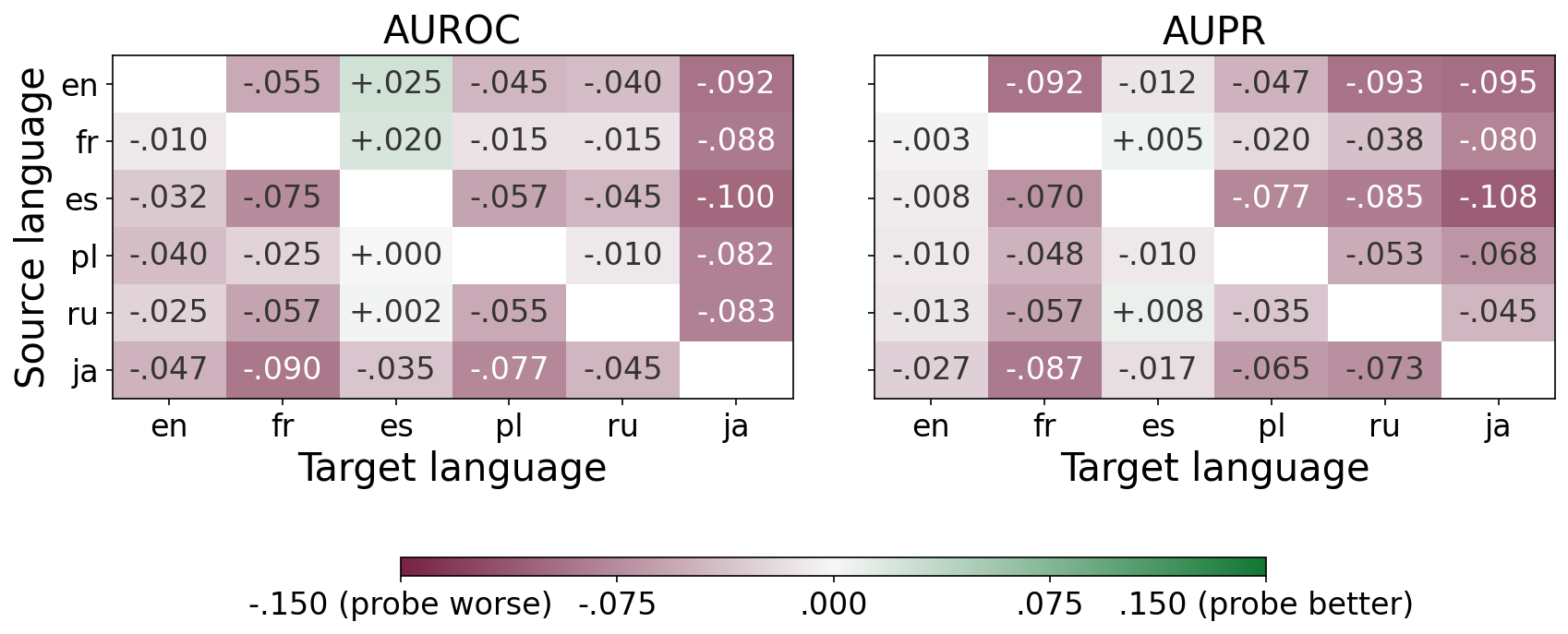}
        \label{fig:cross_query_discr}
    \end{subfigure}
    \\[-8pt]
    \begin{subfigure}{\linewidth}
        \centering
        \includegraphics[width=\linewidth]{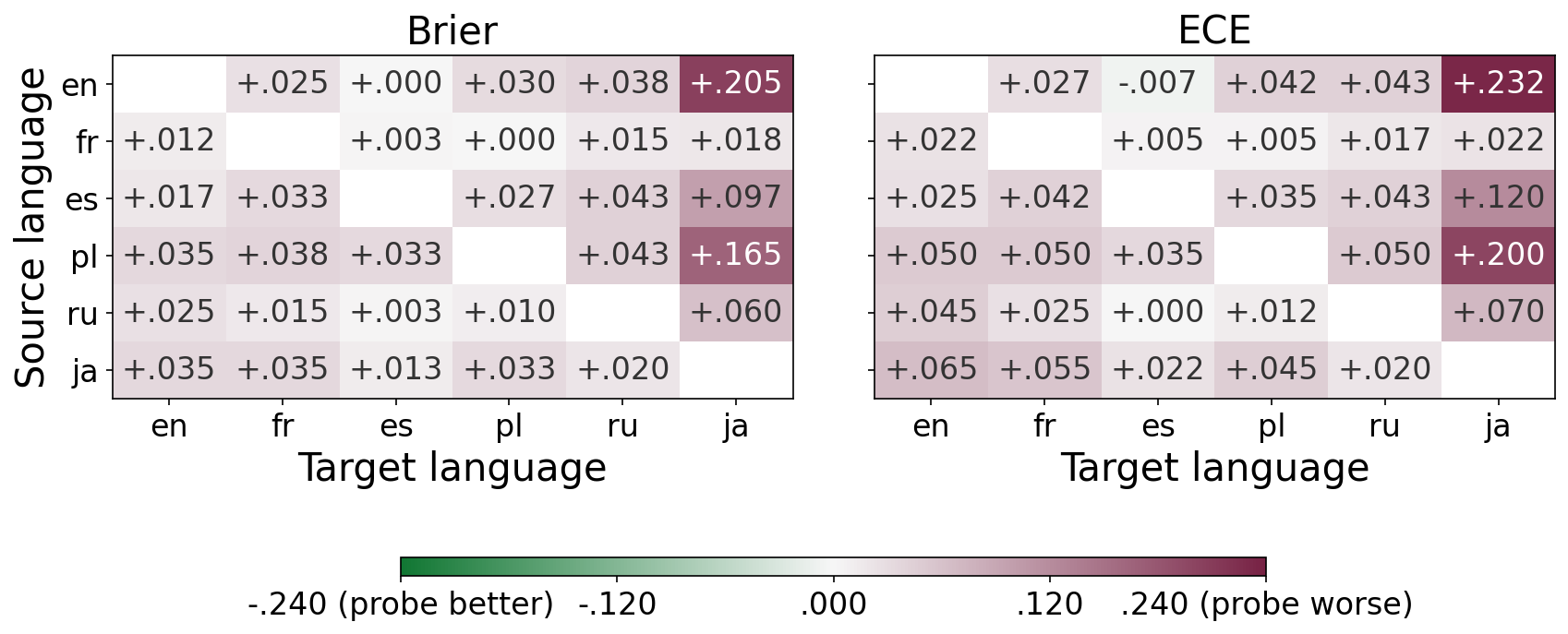}
        \label{fig:cross_query_calib}
    \end{subfigure}
    \vspace{-35pt}
    \caption{Gap between the zero-shot probe and the target-language oracle at the last-query token, averaged over both LLMs and both datasets. Top: discrimination, bottom: calibration.}\label{fig:cross_query}
\end{figure}

\begin{figure}[tb!]
    \centering
    \begin{subfigure}{\linewidth}
        \centering
        \includegraphics[width=\linewidth]{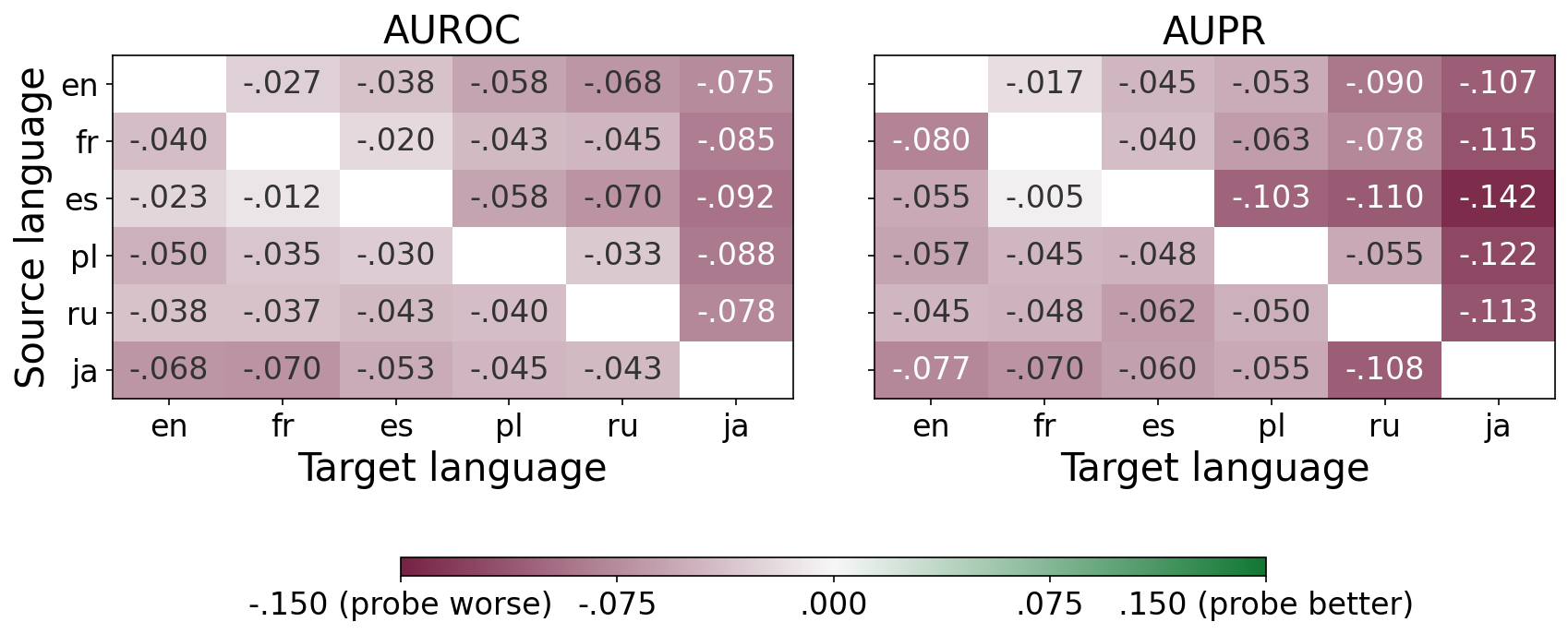}
        \label{fig:cross_answer_discr}
    \end{subfigure}
    \\[-8pt]
    \begin{subfigure}{\linewidth}
        \centering
        \includegraphics[width=\linewidth]{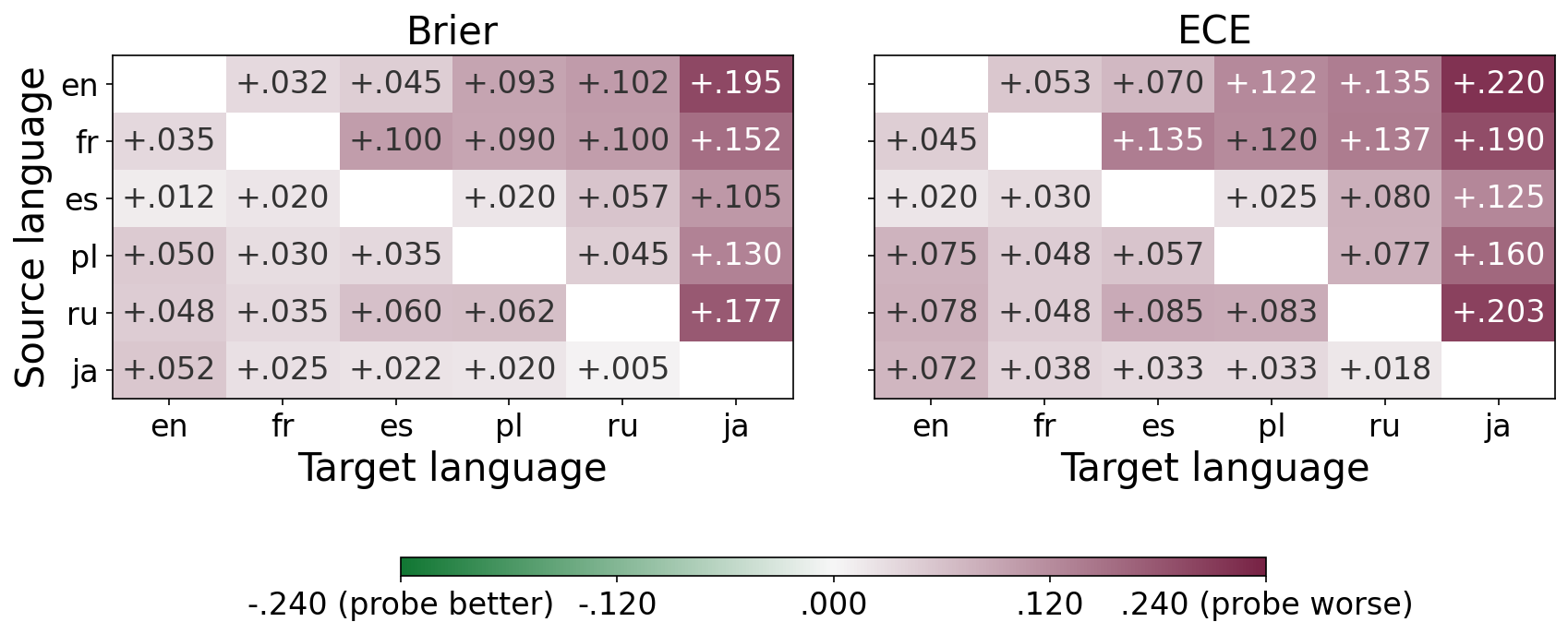}
        \label{fig:cross_answer_calib}
    \end{subfigure}
    \vspace{-35pt}
    \caption{Gap between the zero-shot probe and the target-language oracle at the last-answer token, averaged over both LLMs and both datasets. Top: discrimination, bottom: calibration.}\label{fig:cross_answer}
\end{figure}

We have computed the complete 6 $\times$ 6 source×target matrix for all eight conditions (2 LLMs $\times$ 2 datasets $\times$ 2 token positions) and all four metrics. For easier comparison, we report the average gap between our probe and an oracle probe (directly trained on the target language) for the last-query \cref{fig:cross_query} and the last-answer \cref{fig:cross_answer} hidden states. For the discrimination metrics, the more negative the difference, the worse is our zero-shot probe compared to the oracle. For the calibration metrics, the more positive the difference, the worse is the zero-shot probe. \cref{tab:confidence-estimates-llama-en} and \cref{tab:confidence-estimates-qwen-en} show the complete results with English as the source language.

\subsection{Comparison to Other Estimators}\label{app:uncertainty-baselines}

\begin{table*}[tb]
\centering
\ra{1.3}
\renewcommand{\arraystretch}{1.3}
\setlength{\tabcolsep}{4pt}
\footnotesize
\resizebox{0.8\textwidth}{!}{%
\begin{tabular}{llcccccccccc}
\toprule
 & & & \multicolumn{4}{c}{\textbf{MKQA}} & & \multicolumn{4}{c}{\textbf{Global-MMLU}} \\
\cmidrule(lr){4-7} \cmidrule(lr){9-12}
 & & & AUROC$\uparrow$ & AUPR$\uparrow$ & Brier$\downarrow$ & ECE$\downarrow$ & & AUROC$\uparrow$ & AUPR$\uparrow$ & Brier$\downarrow$ & ECE$\downarrow$ \\
\midrule
\multirow{6}{*}{\rotatebox{90}{fr}} & Seq. Likelihood & & .80$\pm$.00 & .45$\pm$.00 & .52$\pm$.00 & .62$\pm$.00 &  & .72$\pm$.00 & .53$\pm$.00 & .51$\pm$.00 & .58$\pm$.00 \\
 & P(True) & & .63$\pm$.00 & .22$\pm$.00 & .18$\pm$.00 & .18$\pm$.00 &  & .55$\pm$.00 & .32$\pm$.00 & .27$\pm$.00 & .27$\pm$.00 \\
 & Mass-Mean Probe & & .67$\pm$.00 & .52$\pm$.00 & .43$\pm$.00 & .43$\pm$.00 &  & .60$\pm$.00 & .52$\pm$.00 & .40$\pm$.00 & .40$\pm$.00 \\
 & Verbalized Conf. & & .71$\pm$.00 & .44$\pm$.00 & .64$\pm$.00 & .70$\pm$.00 &  & .75$\pm$.00 & .59$\pm$.00 & .54$\pm$.00 & .59$\pm$.00 \\
\noalign{\vskip 1pt}
\cdashline{2-12}
\noalign{\vskip 1pt}
 & Cross-ling. Probe & & .82$\pm$.01 & .49$\pm$.02 & .14$\pm$.01 & .12$\pm$.01 &  & .78$\pm$.00 & .58$\pm$.01 & .19$\pm$.00 & .16$\pm$.00 \\
\cmidrule(lr){2-12}
\multirow{6}{*}{\rotatebox{90}{en}} & Seq. Likelihood & & .80$\pm$.00 & .61$\pm$.00 & .45$\pm$.00 & .54$\pm$.00 &  & .73$\pm$.00 & .66$\pm$.00 & .45$\pm$.00 & .48$\pm$.00 \\
 & P(True) & & .69$\pm$.00 & .42$\pm$.00 & .26$\pm$.00 & .26$\pm$.00 &  & .62$\pm$.00 & .48$\pm$.00 & .39$\pm$.00 & .39$\pm$.00 \\
 & Mass-Mean Probe & & .67$\pm$.00 & .58$\pm$.00 & .36$\pm$.00 & .37$\pm$.00 &  & .56$\pm$.00 & .64$\pm$.00 & .48$\pm$.00 & .48$\pm$.00 \\
 & Verbalized Conf. & & .70$\pm$.00 & .57$\pm$.00 & .61$\pm$.00 & .64$\pm$.00 &  & .72$\pm$.00 & .68$\pm$.00 & .51$\pm$.00 & .54$\pm$.00 \\
\noalign{\vskip 1pt}
\cdashline{2-12}
\noalign{\vskip 1pt}
 & Cross-ling. Probe & & .70$\pm$.01 & .44$\pm$.01 & .24$\pm$.01 & .22$\pm$.01 &  & .72$\pm$.01 & .60$\pm$.01 & .27$\pm$.01 & .24$\pm$.01 \\
\cmidrule(lr){2-12}
\multirow{6}{*}{\rotatebox{90}{es}} & Seq. Likelihood & & .81$\pm$.00 & .52$\pm$.00 & .51$\pm$.00 & .61$\pm$.00 &  & .70$\pm$.00 & .51$\pm$.00 & .53$\pm$.00 & .59$\pm$.00 \\
 & P(True) & & .30$\pm$.00 & .13$\pm$.00 & .19$\pm$.00 & .19$\pm$.00 &  & .41$\pm$.00 & .23$\pm$.00 & .26$\pm$.00 & .26$\pm$.00 \\
 & Mass-Mean Probe & & .65$\pm$.00 & .52$\pm$.00 & .45$\pm$.00 & .45$\pm$.00 &  & .59$\pm$.00 & .50$\pm$.00 & .41$\pm$.00 & .41$\pm$.00 \\
 & Verbalized Conf. & & .75$\pm$.00 & .51$\pm$.00 & .54$\pm$.00 & .59$\pm$.00 &  & .73$\pm$.00 & .55$\pm$.00 & .52$\pm$.00 & .58$\pm$.00 \\
\noalign{\vskip 1pt}
\cdashline{2-12}
\noalign{\vskip 1pt}
 & Cross-ling. Probe & & .82$\pm$.01 & .53$\pm$.03 & .19$\pm$.02 & .20$\pm$.03 &  & .74$\pm$.01 & .52$\pm$.01 & .23$\pm$.01 & .21$\pm$.02 \\
\cmidrule(lr){2-12}
\multirow{6}{*}{\rotatebox{90}{pl}} & Seq. Likelihood & & .82$\pm$.00 & .48$\pm$.00 & .44$\pm$.00 & .57$\pm$.00 &  & .79$\pm$.00 & .53$\pm$.00 & .48$\pm$.00 & .58$\pm$.00 \\
 & P(True) & & .61$\pm$.00 & .19$\pm$.00 & .15$\pm$.00 & .15$\pm$.00 &  & .53$\pm$.00 & .26$\pm$.00 & .21$\pm$.00 & .21$\pm$.00 \\
 & Mass-Mean Probe & & .71$\pm$.00 & .50$\pm$.00 & .34$\pm$.00 & .35$\pm$.00 &  & .59$\pm$.00 & .58$\pm$.00 & .57$\pm$.00 & .57$\pm$.00 \\
 & Verbalized Conf. & & .67$\pm$.00 & .40$\pm$.00 & .60$\pm$.00 & .65$\pm$.00 &  & .76$\pm$.00 & .52$\pm$.00 & .53$\pm$.00 & .59$\pm$.00 \\
\noalign{\vskip 1pt}
\cdashline{2-12}
\noalign{\vskip 1pt}
 & Cross-ling. Probe & & .73$\pm$.01 & .35$\pm$.03 & .23$\pm$.03 & .24$\pm$.03 &  & .75$\pm$.01 & .48$\pm$.01 & .27$\pm$.03 & .27$\pm$.03 \\
\cmidrule(lr){2-12}
\multirow{6}{*}{\rotatebox{90}{ru}} & Seq. Likelihood & & .79$\pm$.00 & .35$\pm$.00 & .49$\pm$.00 & .62$\pm$.00 &  & .75$\pm$.00 & .47$\pm$.00 & .54$\pm$.00 & .62$\pm$.00 \\
 & P(True) & & .43$\pm$.00 & .12$\pm$.00 & .12$\pm$.00 & .12$\pm$.00 &  & .45$\pm$.00 & .21$\pm$.00 & .21$\pm$.00 & .21$\pm$.00 \\
 & Mass-Mean Probe & & .73$\pm$.00 & .51$\pm$.00 & .40$\pm$.00 & .41$\pm$.00 &  & .62$\pm$.00 & .57$\pm$.00 & .51$\pm$.00 & .51$\pm$.00 \\
 & Verbalized Conf. & & .72$\pm$.00 & .36$\pm$.00 & .55$\pm$.00 & .61$\pm$.00 &  & .73$\pm$.00 & .49$\pm$.00 & .51$\pm$.00 & .57$\pm$.00 \\
\noalign{\vskip 1pt}
\cdashline{2-12}
\noalign{\vskip 1pt}
 & Cross-ling. Probe & & .80$\pm$.01 & .43$\pm$.03 & .15$\pm$.03 & .16$\pm$.04 &  & .72$\pm$.01 & .43$\pm$.01 & .39$\pm$.03 & .42$\pm$.03 \\
\cmidrule(lr){2-12}
\multirow{6}{*}{\rotatebox{90}{ja}} & Seq. Likelihood & & .81$\pm$.00 & .41$\pm$.00 & .43$\pm$.00 & .56$\pm$.00 &  & .74$\pm$.00 & .49$\pm$.00 & .48$\pm$.00 & .57$\pm$.00 \\
 & P(True) & & .60$\pm$.00 & .16$\pm$.00 & .13$\pm$.00 & .13$\pm$.00 &  & .54$\pm$.00 & .24$\pm$.00 & .21$\pm$.00 & .21$\pm$.00 \\
 & Mass-Mean Probe & & .83$\pm$.00 & .57$\pm$.00 & .22$\pm$.00 & .23$\pm$.00 &  & .52$\pm$.00 & .59$\pm$.00 & .73$\pm$.00 & .73$\pm$.00 \\
 & Verbalized Conf. & & .67$\pm$.00 & .37$\pm$.00 & .66$\pm$.00 & .72$\pm$.00 &  & .72$\pm$.00 & .44$\pm$.00 & .57$\pm$.00 & .63$\pm$.00 \\
\noalign{\vskip 1pt}
\cdashline{2-12}
\noalign{\vskip 1pt}
 & Cross-ling. Probe & & .64$\pm$.04 & .23$\pm$.04 & .16$\pm$.02 & .16$\pm$.02 &  & .72$\pm$.01 & .44$\pm$.02 & .31$\pm$.02 & .31$\pm$.02 \\
\bottomrule
\end{tabular}%
}
\caption{Complete answer-token confidence estimation comparison for Qwen3 8B, with probes trained in French.}
\label{tab:baseline-comparison-qwen3_8b_full}
\end{table*}

\begin{table*}[tb]
\centering
\ra{1.3}
\renewcommand{\arraystretch}{1.3}
\setlength{\tabcolsep}{4pt}
\footnotesize
\resizebox{0.8\textwidth}{!}{%
\begin{tabular}{llcccccccccc}
\toprule
 & & & \multicolumn{4}{c}{\textbf{MKQA}} & & \multicolumn{4}{c}{\textbf{Global-MMLU}} \\
\cmidrule(lr){4-7} \cmidrule(lr){9-12}
 & & & AUROC$\uparrow$ & AUPR$\uparrow$ & Brier$\downarrow$ & ECE$\downarrow$ & & AUROC$\uparrow$ & AUPR$\uparrow$ & Brier$\downarrow$ & ECE$\downarrow$ \\
\midrule
\multirow{6}{*}{\rotatebox{90}{fr}} & Seq. Likelihood & & .72$\pm$.00 & .51$\pm$.00 & .39$\pm$.00 & .45$\pm$.00 &  & .77$\pm$.00 & .57$\pm$.00 & .28$\pm$.00 & .35$\pm$.00 \\
 & P(True) & & .55$\pm$.00 & .34$\pm$.00 & .27$\pm$.00 & .27$\pm$.00 &  & .53$\pm$.00 & .27$\pm$.00 & .24$\pm$.00 & .24$\pm$.00 \\
 & Mass-Mean Probe & & .61$\pm$.00 & .57$\pm$.00 & .48$\pm$.00 & .48$\pm$.00 &  & .64$\pm$.00 & .56$\pm$.00 & .46$\pm$.00 & .46$\pm$.00 \\
 & Verbalized Conf. & & .62$\pm$.00 & .61$\pm$.00 & .55$\pm$.00 & .57$\pm$.00 &  & .67$\pm$.00 & .60$\pm$.00 & .58$\pm$.00 & .62$\pm$.00 \\
\noalign{\vskip 1pt}
\cdashline{2-12}
\noalign{\vskip 1pt}
 & Cross-ling. Probe & & .82$\pm$.00 & .58$\pm$.01 & .17$\pm$.00 & .11$\pm$.01 &  & .76$\pm$.01 & .51$\pm$.03 & .19$\pm$.01 & .15$\pm$.01 \\
\cmidrule(lr){2-12}
\multirow{6}{*}{\rotatebox{90}{en}} & Seq. Likelihood & & .72$\pm$.00 & .59$\pm$.00 & .32$\pm$.00 & .33$\pm$.00 &  & .80$\pm$.00 & .71$\pm$.00 & .26$\pm$.00 & .28$\pm$.00 \\
 & P(True) & & .35$\pm$.00 & .30$\pm$.00 & .39$\pm$.00 & .39$\pm$.00 &  & .51$\pm$.00 & .35$\pm$.00 & .36$\pm$.00 & .36$\pm$.00 \\
 & Mass-Mean Probe & & .59$\pm$.00 & .65$\pm$.00 & .47$\pm$.00 & .47$\pm$.00 &  & .68$\pm$.00 & .67$\pm$.00 & .38$\pm$.00 & .38$\pm$.00 \\
 & Verbalized Conf. & & .61$\pm$.00 & .67$\pm$.00 & .50$\pm$.00 & .51$\pm$.00 &  & .74$\pm$.00 & .69$\pm$.00 & .43$\pm$.00 & .47$\pm$.00 \\
\noalign{\vskip 1pt}
\cdashline{2-12}
\noalign{\vskip 1pt}
 & Cross-ling. Probe & & .76$\pm$.00 & .62$\pm$.01 & .23$\pm$.01 & .16$\pm$.03 &  & .75$\pm$.01 & .61$\pm$.02 & .22$\pm$.01 & .17$\pm$.03 \\
\cmidrule(lr){2-12}
\multirow{6}{*}{\rotatebox{90}{es}} & Seq. Likelihood & & .73$\pm$.00 & .44$\pm$.00 & .38$\pm$.00 & .46$\pm$.00 &  & .77$\pm$.00 & .58$\pm$.00 & .29$\pm$.00 & .36$\pm$.00 \\
 & P(True) & & .45$\pm$.00 & .21$\pm$.00 & .24$\pm$.00 & .24$\pm$.00 &  & .56$\pm$.00 & .27$\pm$.00 & .23$\pm$.00 & .23$\pm$.00 \\
 & Mass-Mean Probe & & .57$\pm$.00 & .55$\pm$.00 & .56$\pm$.00 & .56$\pm$.00 &  & .66$\pm$.00 & .58$\pm$.00 & .46$\pm$.00 & .46$\pm$.00 \\
 & Verbalized Conf. & & .67$\pm$.00 & .60$\pm$.00 & .51$\pm$.00 & .53$\pm$.00 &  & .73$\pm$.00 & .61$\pm$.00 & .55$\pm$.00 & .61$\pm$.00 \\
\noalign{\vskip 1pt}
\cdashline{2-12}
\noalign{\vskip 1pt}
 & Cross-ling. Probe & & .77$\pm$.01 & .48$\pm$.01 & .50$\pm$.05 & .54$\pm$.04 &  & .78$\pm$.01 & .53$\pm$.03 & .16$\pm$.01 & .12$\pm$.02 \\
\cmidrule(lr){2-12}
\multirow{6}{*}{\rotatebox{90}{pl}} & Seq. Likelihood & & .75$\pm$.00 & .52$\pm$.00 & .31$\pm$.00 & .38$\pm$.00 &  & .80$\pm$.00 & .56$\pm$.00 & .31$\pm$.00 & .42$\pm$.00 \\
 & P(True) & & .63$\pm$.00 & .32$\pm$.00 & .25$\pm$.00 & .25$\pm$.00 &  & .48$\pm$.00 & .18$\pm$.00 & .19$\pm$.00 & .19$\pm$.00 \\
 & Mass-Mean Probe & & .65$\pm$.00 & .49$\pm$.00 & .31$\pm$.00 & .31$\pm$.00 &  & .67$\pm$.00 & .50$\pm$.00 & .36$\pm$.00 & .36$\pm$.00 \\
 & Verbalized Conf. & & .61$\pm$.00 & .56$\pm$.00 & .52$\pm$.00 & .54$\pm$.00 &  & .66$\pm$.00 & .48$\pm$.00 & .51$\pm$.00 & .58$\pm$.00 \\
\noalign{\vskip 1pt}
\cdashline{2-12}
\noalign{\vskip 1pt}
 & Cross-ling. Probe & & .74$\pm$.01 & .47$\pm$.01 & .19$\pm$.01 & .15$\pm$.01 &  & .77$\pm$.01 & .47$\pm$.02 & .32$\pm$.07 & .36$\pm$.08 \\
\cmidrule(lr){2-12}
\multirow{6}{*}{\rotatebox{90}{ru}} & Seq. Likelihood & & .75$\pm$.00 & .40$\pm$.00 & .35$\pm$.00 & .47$\pm$.00 &  & .78$\pm$.00 & .53$\pm$.00 & .29$\pm$.00 & .39$\pm$.00 \\
 & P(True) & & .64$\pm$.00 & .24$\pm$.00 & .14$\pm$.00 & .06$\pm$.00 &  & .55$\pm$.00 & .26$\pm$.00 & .16$\pm$.00 & .10$\pm$.00 \\
 & Mass-Mean Probe & & .66$\pm$.00 & .45$\pm$.00 & .35$\pm$.00 & .35$\pm$.00 &  & .61$\pm$.00 & .48$\pm$.00 & .47$\pm$.00 & .47$\pm$.00 \\
 & Verbalized Conf. & & .60$\pm$.00 & .49$\pm$.00 & .59$\pm$.00 & .60$\pm$.00 &  & .63$\pm$.00 & .53$\pm$.00 & .61$\pm$.00 & .65$\pm$.00 \\
\noalign{\vskip 1pt}
\cdashline{2-12}
\noalign{\vskip 1pt}
 & Cross-ling. Probe & & .78$\pm$.01 & .41$\pm$.01 & .14$\pm$.01 & .11$\pm$.03 &  & .73$\pm$.01 & .48$\pm$.02 & .31$\pm$.07 & .35$\pm$.08 \\
\cmidrule(lr){2-12}
\multirow{6}{*}{\rotatebox{90}{ja}} & Seq. Likelihood & & .76$\pm$.00 & .32$\pm$.00 & .35$\pm$.00 & .48$\pm$.00 &  & .78$\pm$.00 & .53$\pm$.00 & .26$\pm$.00 & .35$\pm$.00 \\
 & P(True) & & .53$\pm$.00 & .13$\pm$.00 & .12$\pm$.00 & .11$\pm$.00 &  & .61$\pm$.00 & .24$\pm$.00 & .18$\pm$.00 & .17$\pm$.00 \\
 & Mass-Mean Probe & & .62$\pm$.00 & .41$\pm$.00 & .42$\pm$.00 & .42$\pm$.00 &  & .69$\pm$.00 & .42$\pm$.00 & .24$\pm$.00 & .24$\pm$.00 \\
 & Verbalized Conf. & & .70$\pm$.00 & .37$\pm$.00 & .43$\pm$.00 & .55$\pm$.00 &  & .61$\pm$.00 & .38$\pm$.00 & .47$\pm$.00 & .55$\pm$.00 \\
\noalign{\vskip 1pt}
\cdashline{2-12}
\noalign{\vskip 1pt}
 & Cross-ling. Probe & & .77$\pm$.01 & .32$\pm$.01 & .43$\pm$.10 & .49$\pm$.10 &  & .72$\pm$.01 & .43$\pm$.03 & .28$\pm$.04 & .29$\pm$.04 \\
\bottomrule
\end{tabular}%

}
\caption{Complete answer-token confidence estimation comparison for Llama 3.1 8B, with probes trained in French.}
\label{tab:baseline-comparison-llama_full}
\end{table*}

\begin{table*}[tb]
\centering
\ra{1.3}
\renewcommand{\arraystretch}{1.3}
\setlength{\tabcolsep}{4pt}
\footnotesize
\resizebox{0.8\textwidth}{!}{%
\begin{tabular}{llcccccccccc}
\toprule
 & & & \multicolumn{4}{c}{\textbf{MKQA}} & & \multicolumn{4}{c}{\textbf{Global-MMLU}} \\
\cmidrule(lr){4-7} \cmidrule(lr){9-12}
 & & & AUROC$\uparrow$ & AUPR$\uparrow$ & Brier$\downarrow$ & ECE$\downarrow$ & & AUROC$\uparrow$ & AUPR$\uparrow$ & Brier$\downarrow$ & ECE$\downarrow$ \\
\midrule
\multirow{4}{*}{\rotatebox{90}{fr}} & Seq. Likelihood & & .63$\pm$.00 & .26$\pm$.00 & .15$\pm$.00 & .12$\pm$.00 &  & .50$\pm$.00 & .26$\pm$.00 & .24$\pm$.00 & .19$\pm$.00 \\
 & Mass-Mean Probe & & .64$\pm$.00 & .47$\pm$.00 & .44$\pm$.00 & .44$\pm$.00 &  & .69$\pm$.00 & .57$\pm$.00 & .38$\pm$.00 & .38$\pm$.00 \\
\noalign{\vskip 1pt} \cdashline{2-12} \noalign{\vskip 1pt}
 & Cross-ling. Probe & & .73$\pm$.01 & .36$\pm$.01 & .17$\pm$.01 & .14$\pm$.00 &  & .71$\pm$.01 & .49$\pm$.02 & .21$\pm$.01 & .16$\pm$.01 \\
\cmidrule(lr){2-12}
\multirow{4}{*}{\rotatebox{90}{en}} & Seq. Likelihood & & .62$\pm$.00 & .38$\pm$.00 & .24$\pm$.00 & .22$\pm$.00 &  & .51$\pm$.00 & .38$\pm$.00 & .36$\pm$.00 & .34$\pm$.00 \\
 & Mass-Mean Probe & & .69$\pm$.00 & .59$\pm$.00 & .41$\pm$.00 & .42$\pm$.00 &  & .65$\pm$.00 & .64$\pm$.00 & .40$\pm$.00 & .40$\pm$.00 \\
\noalign{\vskip 1pt} \cdashline{2-12} \noalign{\vskip 1pt}
 & Cross-ling. Probe & & .70$\pm$.01 & .47$\pm$.01 & .22$\pm$.01 & .18$\pm$.01 &  & .64$\pm$.02 & .51$\pm$.02 & .31$\pm$.01 & .26$\pm$.01 \\
\cmidrule(lr){2-12}
\multirow{4}{*}{\rotatebox{90}{es}} & Seq. Likelihood & & .64$\pm$.00 & .28$\pm$.00 & .16$\pm$.00 & .12$\pm$.00 &  & .52$\pm$.00 & .26$\pm$.00 & .22$\pm$.00 & .18$\pm$.00 \\
 & Mass-Mean Probe & & .64$\pm$.00 & .49$\pm$.00 & .43$\pm$.00 & .44$\pm$.00 &  & .69$\pm$.00 & .56$\pm$.00 & .37$\pm$.00 & .37$\pm$.00 \\
\noalign{\vskip 1pt} \cdashline{2-12} \noalign{\vskip 1pt}
 & Cross-ling. Probe & & .71$\pm$.01 & .34$\pm$.01 & .17$\pm$.00 & .14$\pm$.01 &  & .68$\pm$.01 & .41$\pm$.02 & .27$\pm$.02 & .25$\pm$.03 \\
\cmidrule(lr){2-12}
\multirow{4}{*}{\rotatebox{90}{pl}} & Seq. Likelihood & & .63$\pm$.00 & .26$\pm$.00 & .13$\pm$.00 & .09$\pm$.00 &  & .57$\pm$.00 & .26$\pm$.00 & .18$\pm$.00 & .12$\pm$.00 \\
 & Mass-Mean Probe & & .67$\pm$.00 & .46$\pm$.00 & .41$\pm$.00 & .41$\pm$.00 &  & .72$\pm$.00 & .52$\pm$.00 & .33$\pm$.00 & .33$\pm$.00 \\
\noalign{\vskip 1pt} \cdashline{2-12} \noalign{\vskip 1pt}
 & Cross-ling. Probe & & .67$\pm$.01 & .28$\pm$.01 & .15$\pm$.01 & .13$\pm$.01 &  & .73$\pm$.01 & .40$\pm$.01 & .20$\pm$.01 & .17$\pm$.02 \\
\cmidrule(lr){2-12}
\multirow{4}{*}{\rotatebox{90}{ru}} & Seq. Likelihood & & .65$\pm$.00 & .19$\pm$.00 & .11$\pm$.00 & .06$\pm$.00 &  & .51$\pm$.00 & .24$\pm$.00 & .18$\pm$.00 & .12$\pm$.00 \\
 & Mass-Mean Probe & & .69$\pm$.00 & .46$\pm$.00 & .40$\pm$.00 & .41$\pm$.00 &  & .68$\pm$.00 & .50$\pm$.00 & .39$\pm$.00 & .39$\pm$.00 \\
\noalign{\vskip 1pt} \cdashline{2-12} \noalign{\vskip 1pt}
 & Cross-ling. Probe & & .72$\pm$.01 & .30$\pm$.01 & .12$\pm$.00 & .11$\pm$.01 &  & .68$\pm$.01 & .35$\pm$.02 & .23$\pm$.02 & .21$\pm$.03 \\
\cmidrule(lr){2-12}
\multirow{4}{*}{\rotatebox{90}{ja}} & Seq. Likelihood & & .59$\pm$.00 & .18$\pm$.00 & .12$\pm$.00 & .08$\pm$.00 &  & .52$\pm$.00 & .23$\pm$.00 & .19$\pm$.00 & .16$\pm$.00 \\
 & Mass-Mean Probe & & .56$\pm$.00 & .56$\pm$.00 & .85$\pm$.00 & .85$\pm$.00 &  & .67$\pm$.00 & .52$\pm$.00 & .43$\pm$.00 & .43$\pm$.00 \\
\noalign{\vskip 1pt} \cdashline{2-12} \noalign{\vskip 1pt}
 & Cross-ling. Probe & & .57$\pm$.02 & .16$\pm$.01 & .13$\pm$.01 & .11$\pm$.01 &  & .64$\pm$.01 & .32$\pm$.02 & .20$\pm$.02 & .17$\pm$.03 \\
\bottomrule
\end{tabular}%
}
\caption{Complete query-token confidence estimation comparison for Qwen3 8B, with probes trained in French.}
\label{tab:query-comparison-qwen3_8b}
\end{table*}

\begin{table*}[tb]
\centering
\ra{1.3}
\renewcommand{\arraystretch}{1.3}
\setlength{\tabcolsep}{4pt}
\footnotesize
\resizebox{0.8\textwidth}{!}{%
\begin{tabular}{llcccccccccc}
\toprule
 & & & \multicolumn{4}{c}{\textbf{MKQA}} & & \multicolumn{4}{c}{\textbf{Global-MMLU}} \\
\cmidrule(lr){4-7} \cmidrule(lr){9-12}
 & & & AUROC$\uparrow$ & AUPR$\uparrow$ & Brier$\downarrow$ & ECE$\downarrow$ & & AUROC$\uparrow$ & AUPR$\uparrow$ & Brier$\downarrow$ & ECE$\downarrow$ \\
\midrule
\multirow{4}{*}{\rotatebox{90}{fr}} & Seq. Likelihood & & .66$\pm$.00 & .40$\pm$.00 & .23$\pm$.00 & .20$\pm$.00 &  & .52$\pm$.00 & .26$\pm$.00 & .20$\pm$.00 & .14$\pm$.00 \\
 & Mass-Mean Probe & & .63$\pm$.00 & .50$\pm$.00 & .41$\pm$.00 & .41$\pm$.00 &  & .71$\pm$.00 & .56$\pm$.00 & .33$\pm$.00 & .33$\pm$.00 \\
\noalign{\vskip 1pt} \cdashline{2-12} \noalign{\vskip 1pt}
 & Cross-ling. Probe & & .68$\pm$.00 & .45$\pm$.00 & .24$\pm$.00 & .20$\pm$.01 &  & .73$\pm$.01 & .48$\pm$.02 & .21$\pm$.00 & .17$\pm$.01 \\
\cmidrule(lr){2-12}
\multirow{4}{*}{\rotatebox{90}{en}} & Seq. Likelihood & & .62$\pm$.00 & .47$\pm$.00 & .37$\pm$.00 & .36$\pm$.00 &  & .46$\pm$.00 & .32$\pm$.00 & .33$\pm$.00 & .31$\pm$.00 \\
 & Mass-Mean Probe & & .65$\pm$.00 & .60$\pm$.00 & .39$\pm$.00 & .39$\pm$.00 &  & .73$\pm$.00 & .68$\pm$.00 & .32$\pm$.00 & .32$\pm$.00 \\
\noalign{\vskip 1pt} \cdashline{2-12} \noalign{\vskip 1pt}
 & Cross-ling. Probe & & .67$\pm$.01 & .53$\pm$.01 & .28$\pm$.00 & .24$\pm$.01 &  & .70$\pm$.02 & .57$\pm$.02 & .26$\pm$.01 & .22$\pm$.01 \\
\cmidrule(lr){2-12}
\multirow{4}{*}{\rotatebox{90}{es}} & Seq. Likelihood & & .60$\pm$.00 & .31$\pm$.00 & .21$\pm$.00 & .18$\pm$.00 &  & .57$\pm$.00 & .30$\pm$.00 & .19$\pm$.00 & .13$\pm$.00 \\
 & Mass-Mean Probe & & .66$\pm$.00 & .51$\pm$.00 & .42$\pm$.00 & .43$\pm$.00 &  & .72$\pm$.00 & .57$\pm$.00 & .33$\pm$.00 & .33$\pm$.00 \\
\noalign{\vskip 1pt} \cdashline{2-12} \noalign{\vskip 1pt}
 & Cross-ling. Probe & & .64$\pm$.00 & .35$\pm$.01 & .24$\pm$.01 & .20$\pm$.01 &  & .71$\pm$.02 & .45$\pm$.02 & .23$\pm$.02 & .21$\pm$.03 \\
\cmidrule(lr){2-12}
\multirow{4}{*}{\rotatebox{90}{pl}} & Seq. Likelihood & & .59$\pm$.00 & .35$\pm$.00 & .22$\pm$.00 & .20$\pm$.00 &  & .53$\pm$.00 & .19$\pm$.00 & .16$\pm$.00 & .08$\pm$.00 \\
 & Mass-Mean Probe & & .69$\pm$.00 & .53$\pm$.00 & .36$\pm$.00 & .36$\pm$.00 &  & .73$\pm$.00 & .51$\pm$.00 & .30$\pm$.00 & .30$\pm$.00 \\
\noalign{\vskip 1pt} \cdashline{2-12} \noalign{\vskip 1pt}
 & Cross-ling. Probe & & .68$\pm$.01 & .39$\pm$.01 & .22$\pm$.00 & .18$\pm$.01 &  & .69$\pm$.01 & .39$\pm$.02 & .21$\pm$.03 & .20$\pm$.04 \\
\cmidrule(lr){2-12}
\multirow{4}{*}{\rotatebox{90}{ru}} & Seq. Likelihood & & .63$\pm$.00 & .22$\pm$.00 & .14$\pm$.00 & .11$\pm$.00 &  & .59$\pm$.00 & .25$\pm$.00 & .16$\pm$.00 & .10$\pm$.00 \\
 & Mass-Mean Probe & & .71$\pm$.00 & .44$\pm$.00 & .34$\pm$.00 & .34$\pm$.00 &  & .66$\pm$.00 & .48$\pm$.00 & .37$\pm$.00 & .38$\pm$.00 \\
\noalign{\vskip 1pt} \cdashline{2-12} \noalign{\vskip 1pt}
 & Cross-ling. Probe & & .71$\pm$.01 & .38$\pm$.01 & .19$\pm$.01 & .18$\pm$.01 &  & .67$\pm$.01 & .34$\pm$.01 & .23$\pm$.02 & .21$\pm$.03 \\
\cmidrule(lr){2-12}
\multirow{4}{*}{\rotatebox{90}{ja}} & Seq. Likelihood & & .58$\pm$.00 & .16$\pm$.00 & .11$\pm$.00 & .07$\pm$.00 &  & .60$\pm$.00 & .24$\pm$.00 & .16$\pm$.00 & .11$\pm$.00 \\
 & Mass-Mean Probe & & .69$\pm$.00 & .32$\pm$.00 & .53$\pm$.00 & .55$\pm$.00 &  & .65$\pm$.00 & .50$\pm$.00 & .46$\pm$.00 & .46$\pm$.00 \\
\noalign{\vskip 1pt} \cdashline{2-12} \noalign{\vskip 1pt}
 & Cross-ling. Probe & & .64$\pm$.01 & .21$\pm$.01 & .21$\pm$.04 & .22$\pm$.05 &  & .69$\pm$.02 & .32$\pm$.02 & .20$\pm$.03 & .18$\pm$.04 \\
\bottomrule
\end{tabular}%

}
\caption{Complete query-token confidence estimation comparison for Llama 3.1 8B, with probes trained in French.}
\label{tab:query-comparison-llama_3.1_8b}
\end{table*}

This section contains the full results for the comparison with other confidence estimators in \cref{sec:comparison-estimator}.
This includes first the full results for all languages, including the few-shot setting in French, in \cref{tab:baseline-comparison-qwen3_8b_full,tab:baseline-comparison-llama_full} for Qwen3 and Llama 3.1 respectively. 
In \cref{sec:comparison-estimator} we chose to compare other estimators to our probe trained on last answer token hidden representations, since other methods also, implicitly or explicitly, have access to full generation of the model. 
To also benchmark the probe trained on the last input token hidden representations, we provide \cref{tab:query-comparison-qwen3_8b,tab:query-comparison-llama_3.1_8b}.

\subsection{Layer Weight Results}\label{app:layer-weights}

\subsubsection{French as Source Language}
\label{app:layer-weights-french}
\begin{figure*}[htb]
    \begin{subfigure}[t]{0.485\linewidth}
    \centering
         \includegraphics[width=\linewidth]{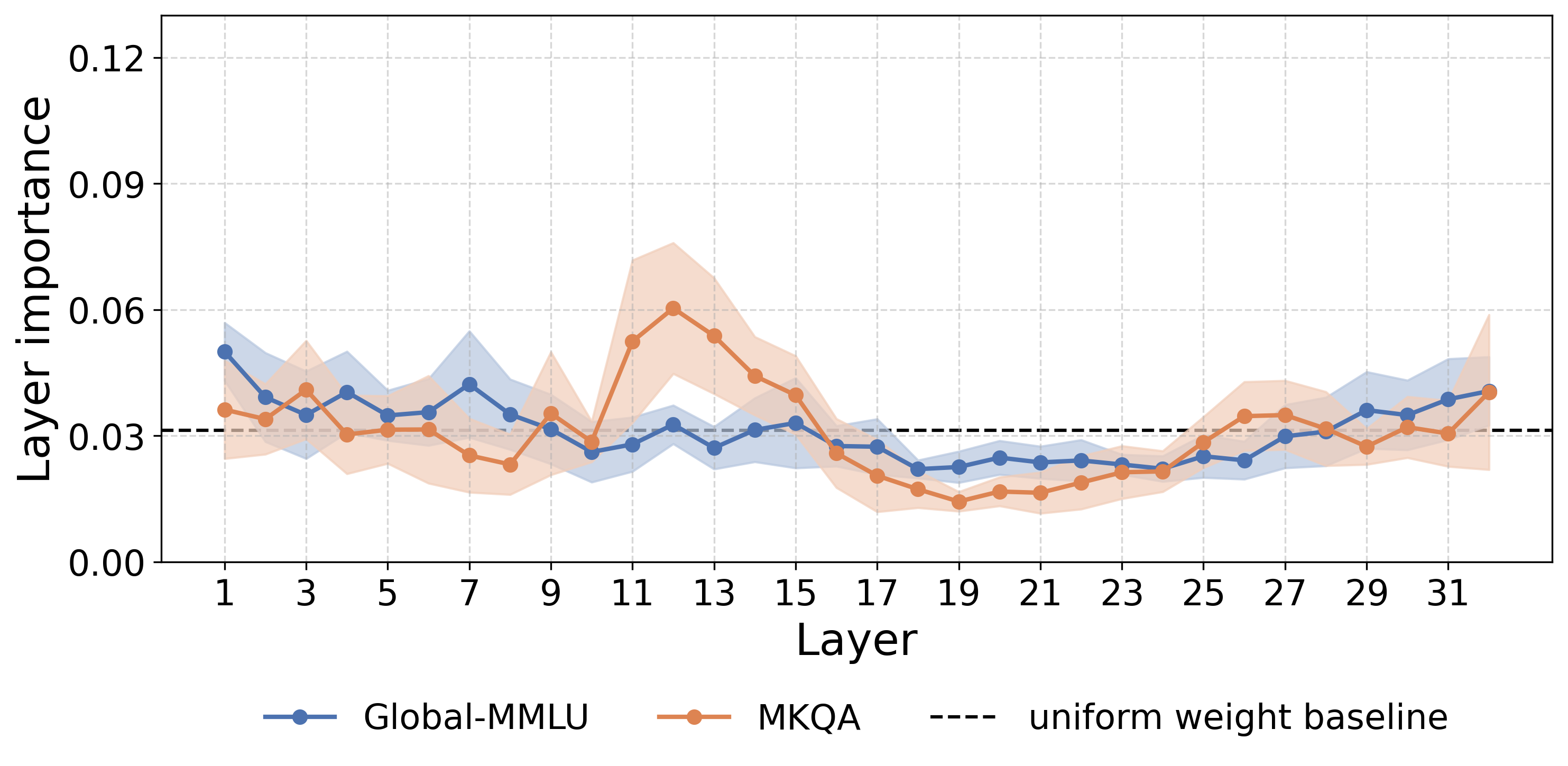}
        \caption{Layer weights for last query token.}
    \end{subfigure}
    \begin{subfigure}[t]{0.458\linewidth}
    \centering
        \includegraphics[width=\linewidth]{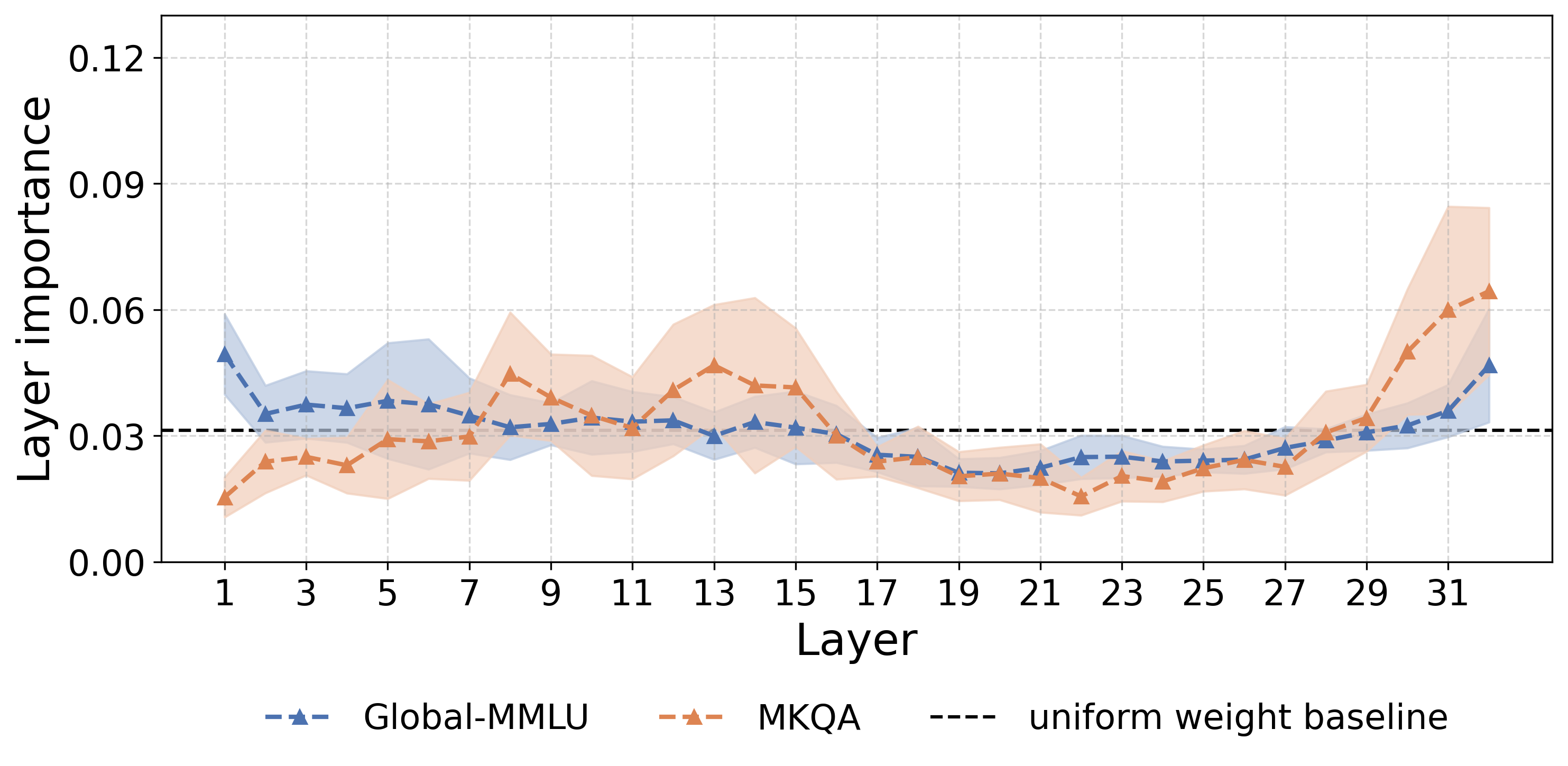}
        \caption{Layer weights for last answer token.}
    \end{subfigure}
    \caption{Distribution of learned layer weights from probes trained with different random seeds on the hidden states of the last (a) query token and (b) answer token for Llama 3.1 8B on Global-MMLU and MKQA. Shaded areas indicate ±1 standard deviation, dashed line shows uniform weight.}\label{fig:layer_importance_llama}
\end{figure*}

The corresponding layer weight plots for \cref{fig:layer_importance_qwen} in \cref{sec:ablations} for Llama 3.1 are given in \cref{fig:layer_importance_llama}.

\subsubsection{All Source Languages}
\label{app:layer-weights-all-sources}

We quantify how much the learned layer-weight profiles differ across source languages ($\sigma_{\text{lang}}$), using the variation of a single probe's profile across random seeds ($\sigma_{\text{seed}}$) as a reference scale. \cref{tab:layer_weight_variation} reports both, their ratio, and the pairwise Spearman rank correlation between the profiles of different source languages. 

\begin{table*}[htb]
\small
\centering
\setlength{\tabcolsep}{6pt}
\renewcommand{\arraystretch}{1.2}
\begin{tabular}{@{}llcccccc@{}}
\toprule
& & & & & & \multicolumn{2}{c}{Spearman $\rho$} \\
\cmidrule(lr){7-8}
Dataset & LLM & $L$ & $\sigma_{\text{lang}}$ & $\sigma_{\text{seed}}$ & Ratio & Mean & Min \\
\midrule
\multirow{2}{*}{MKQA}
& Llama 3.1 8B & 32 & .0083 & .0118 & .70 & .63 & .46 \\
& Qwen3 8B     & 36 & .0045 & .0086 & .52 & .61 & .25 \\
\midrule
\multirow{2}{*}{Global-MMLU}
& Llama 3.1 8B & 32 & .0041 & .0082 & .49 & .73 & .62 \\
& Qwen3 8B     & 36 & .0102 & .0121 & .84 & .79 & .66 \\
\bottomrule
\end{tabular}
\caption{Variation of the learned layer-weight profiles across source languages. $\sigma_{\text{lang}}$ and $\sigma_{\text{seed}}$ are the standard deviations of the per-layer weights across source languages and across random seeds, respectively. Spearman $\rho$ is computed over the 15 source-language pairs.}
\label{tab:layer_weight_variation}
\end{table*}

\subsection{Ablation Results}\label{app:ablation-results}
This section details the ablation methodology of \cref{sec:ablations} and provides additional results.

\paragraph{Weight Ablation.} 
We selectively make layers unavailable to the probe and quantify the impact of this change on the uncertainty estimation task using a sliding window approach. Recall the computation of layer weights in \cref{eq:layer_imp}. 
To ablate a window $W$ of $k$ contiguous layers, we intervene on the probe’s layer-mixing logits by setting $\{w_\ell : \ell \in W\}$ to $-\infty$: 
\\
\begin{equation}
    \alpha_\ell = \mathrm{softmax}(\mathbf{w'})_\ell = \frac{\exp(w_\ell')}{\sum_{k=1}^L \exp(w_k')},
\end{equation}
\noindent where
\begin{equation}
    w_\ell' =
    \begin{cases}
        -\infty, & \ell \in W \\[6pt]
        w_\ell, & \ell \notin W.
    \end{cases}
\end{equation}

\paragraph{Representation Ablation.}

Recall that the probe computes a logit as $z_i = \sum_{\ell=1}^L \alpha_\ell s_\ell + b$ (\cref{eq:repr_pooling,eq:logit}), where $s_{i, \ell} = \mathbf{v}^\top \mathbf{h}_{i}^{(\ell)}$ denoted here. To ablate a window $W$ of $k$ contiguous layers, for each example $i$, we switch its score $s_{i,\ell}$ with the one of another example $s_{j,\ell}$, selected by independently sampling it from the test set. 
This is done to remove useful information for a specific test sample, but to not introduce corrupting information that might unfairly distort probe performance.
The logit of an example $i$ after ablation of window $W$ is therefore:
\\
\begin{equation}
    z_i^{(W)} \;=\; \sum_{\ell \notin W} \alpha_\ell\, s_{i,\ell}
    \;+\; \sum_{\ell \in W} \alpha_\ell\, \tilde{s}_{i,\ell}
    \;+\; b,
\end{equation}
\noindent with
\begin{equation}
    s_{i,\ell} = \mathbf{v}^\top \mathbf{h}_{i}^{(\ell)}, \quad
    \tilde{s}_{i,\ell} = {s}_{\pi(i),\ell},
\end{equation}
\noindent
where $\pi(\cdot)$ is a permutation function defined on the natural numbers up to the cardinality of the test set, mapping each index to another index sampling uniformly without replacement and independently for each layer. 

\begin{figure*}[htb]
    \centering
    \begin{subfigure}[t]{0.49\linewidth}
        \centering
        \includegraphics[width=0.95\linewidth]{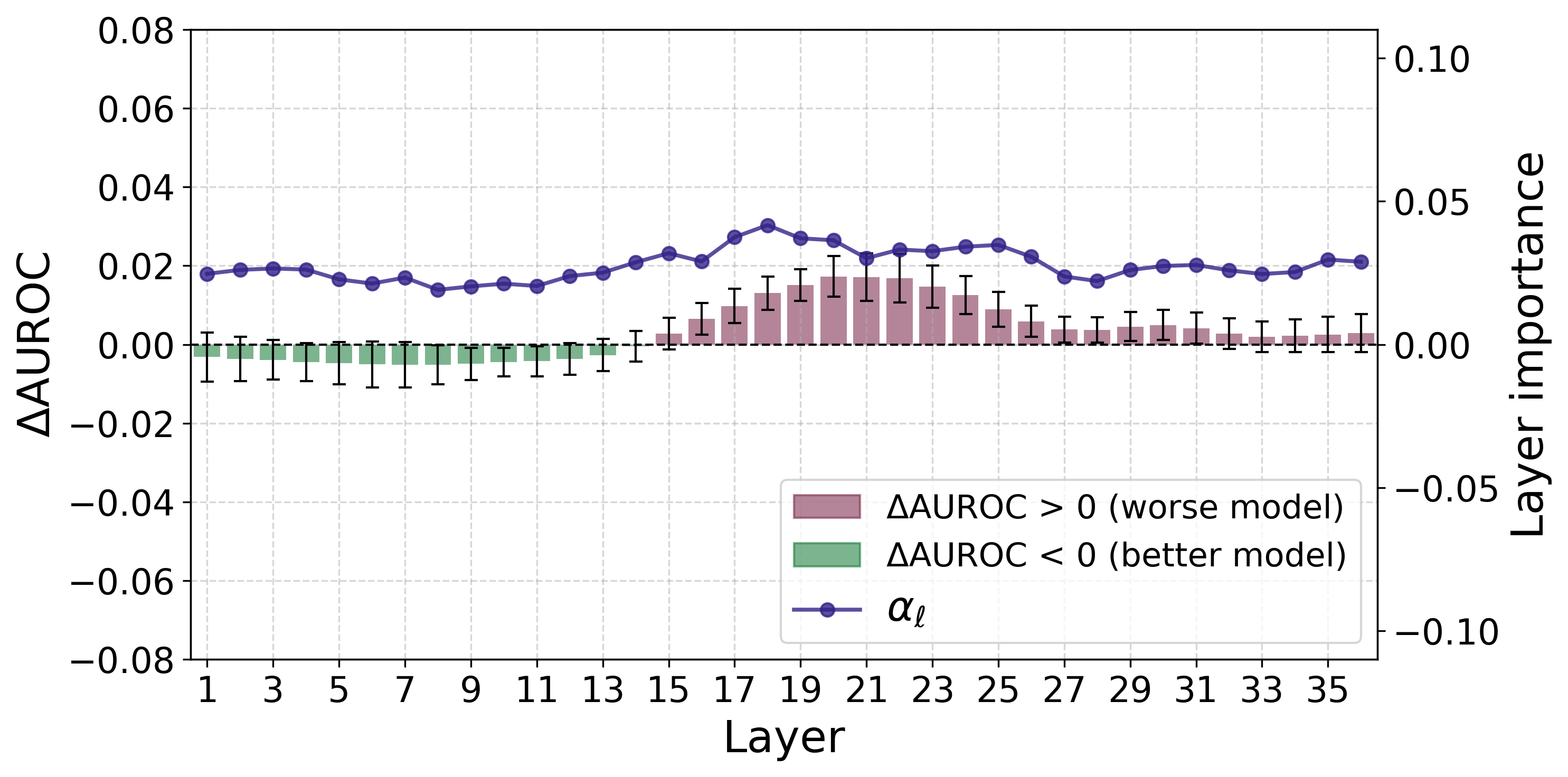}
        \caption{$\Delta$ in AUROC on MKQA.}
    \end{subfigure}
    \begin{subfigure}[t]{0.49\linewidth}
    \centering
        \includegraphics[width=0.95\linewidth]
        {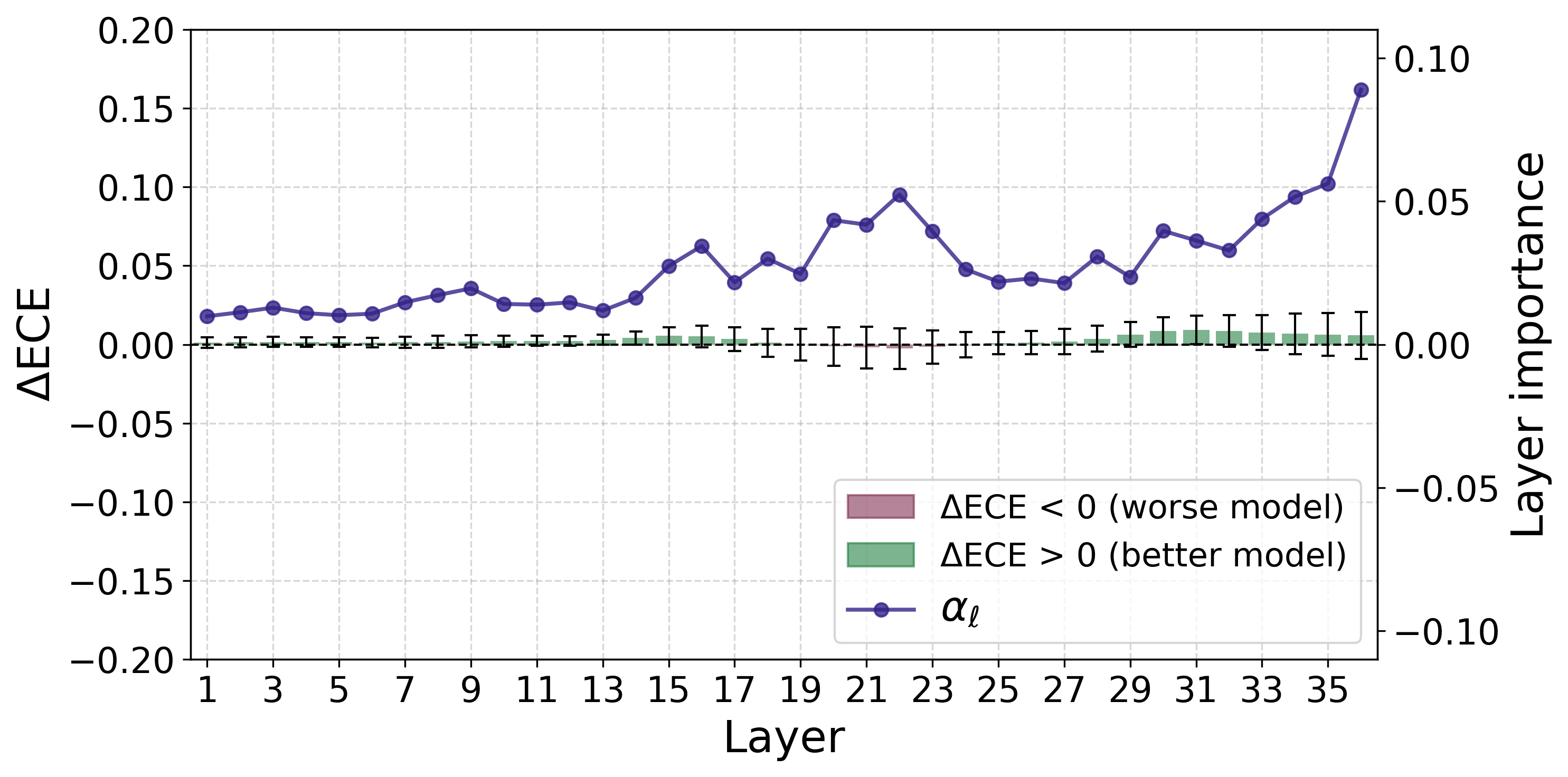}
        \caption{$\Delta$ in ECE score on Global-MMLU.}
    \end{subfigure}
    \caption{Impact of representation ablation on zero-shot AUROC and ECE (avg.\@ over all languages) for Qwen3 8B on the last answer token hidden states. Bars show mean change in metric value over 10 random seeds. Whiskers denote $\pm$1 standard deviation. Red indicates that ablation degrades performance, i.e.\ the layer is beneficial, green that ablation improves performance, i.e.\ the layer is not beneficial for zero-shot transfer.}\label{fig:ablation_examples_representation}
\end{figure*}

\begin{figure*}[htb]
    \centering
    \begin{subfigure}[t]{0.49\linewidth}
        \centering
        \includegraphics[width=0.95\linewidth]{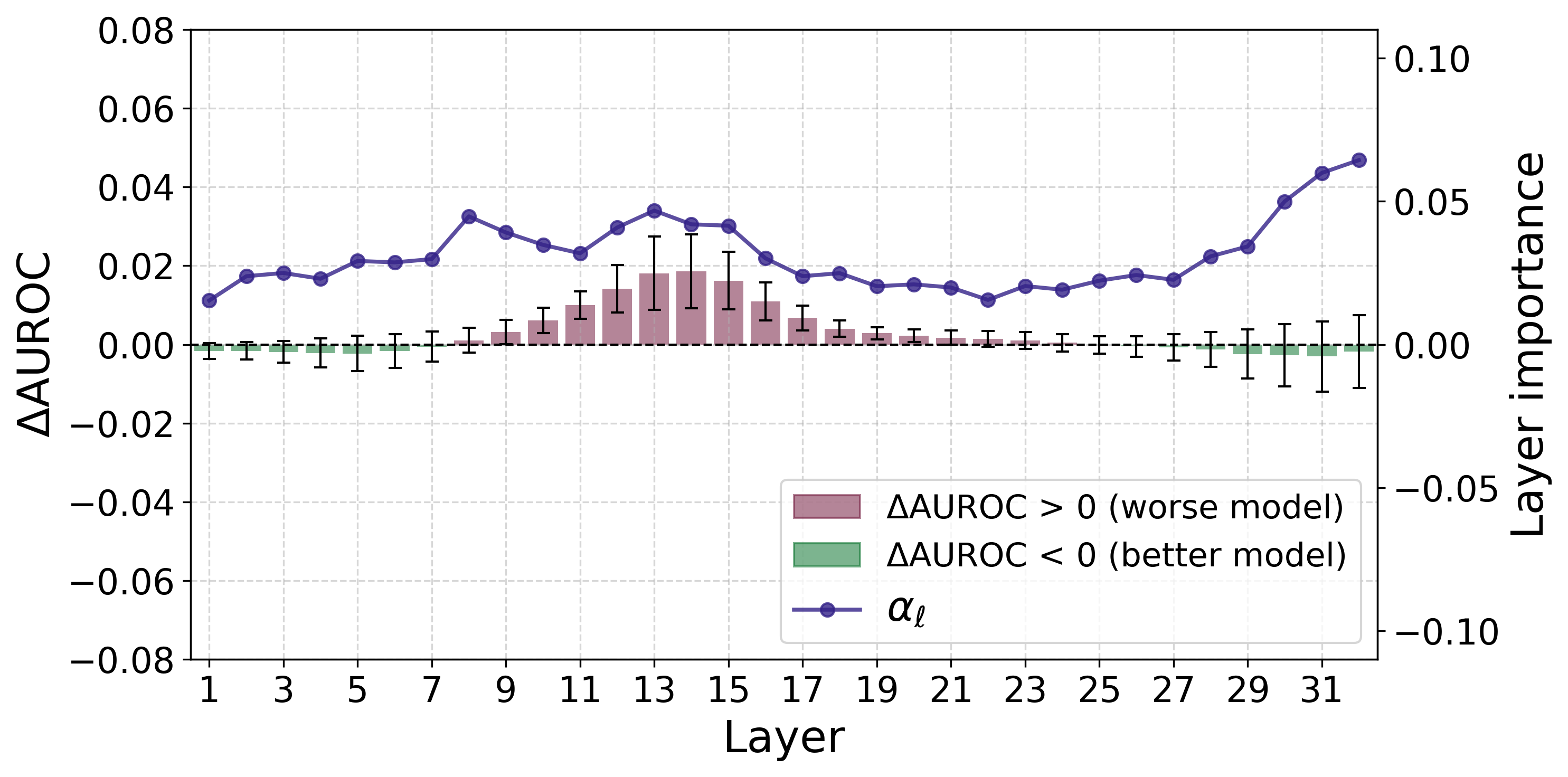}
        \caption{Weight ablation $\Delta$ in AUROC on MKQA.}
    \end{subfigure}
    \begin{subfigure}[t]{0.49\linewidth}
    \centering
        \includegraphics[width=0.95\linewidth]{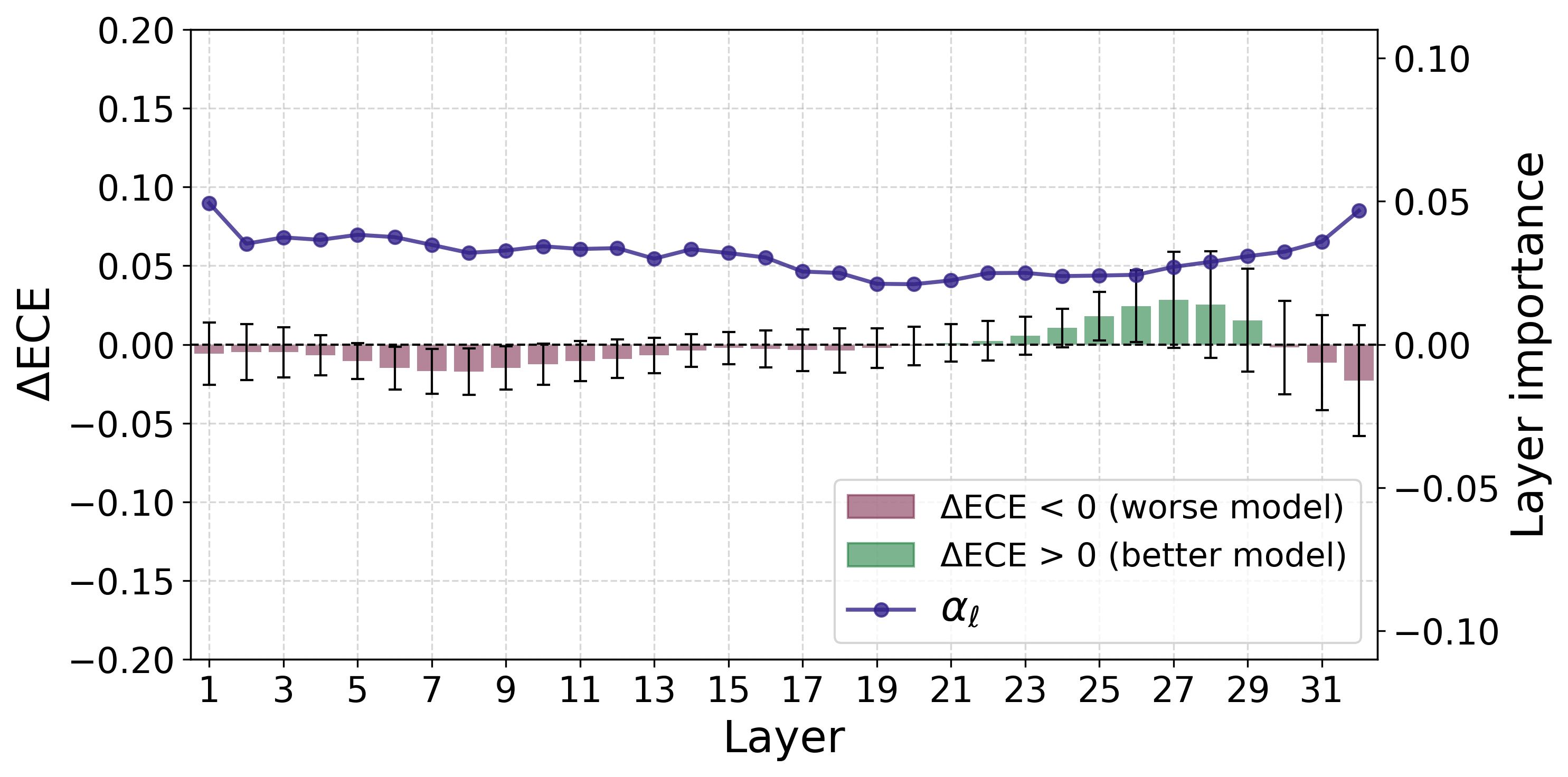}
        \caption{Weight ablation $\Delta$ in ECE score on Global-MMLU.}
    \end{subfigure}
    ~
    \begin{subfigure}[t]{0.49\linewidth}
        \centering
        \includegraphics[width=0.95\linewidth]{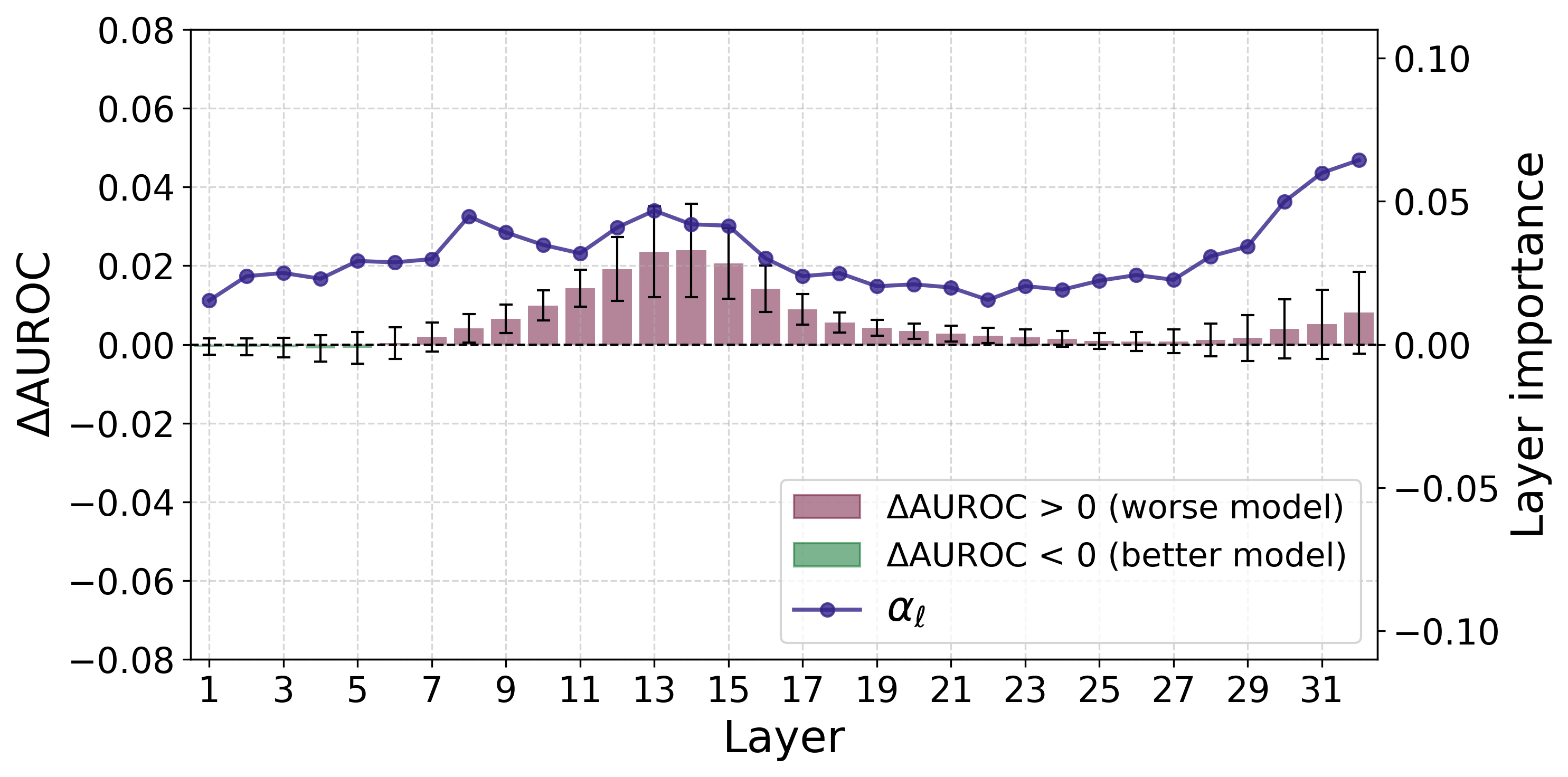}
        \caption{Representation ablation $\Delta$ in AUROC on MKQA.}
    \end{subfigure}
    \begin{subfigure}[t]{0.49\linewidth}
    \centering
        \includegraphics[width=0.95\linewidth]
        {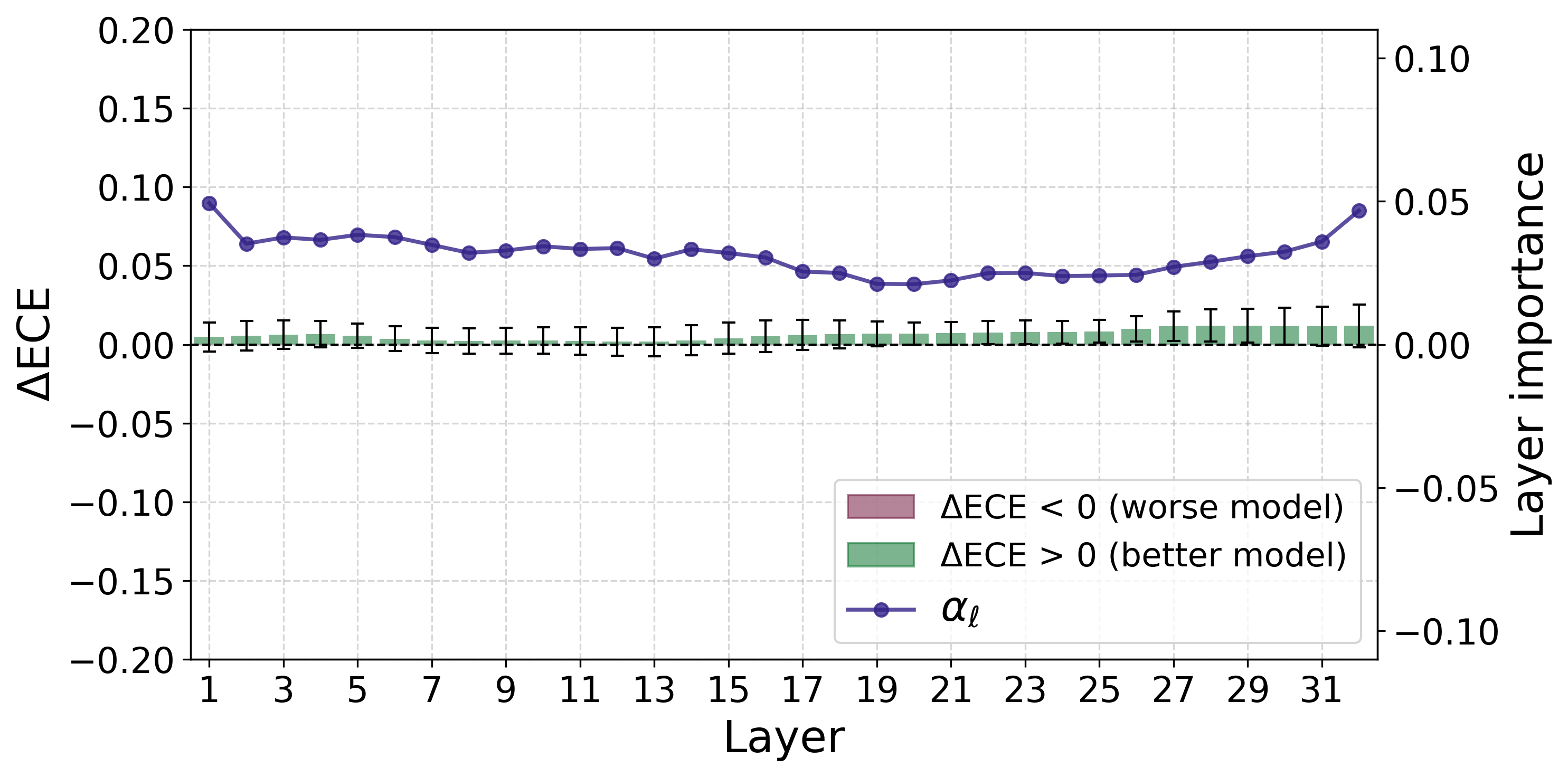}
        \caption{Representation ablation $\Delta$ in ECE score on Global-MMLU.}
    \end{subfigure}
    \caption{Impact of layer ablations on zero-shot AUROC and ECE (avg.\@ over all languages) for Llama 3.1 8B on the last answer token hidden states. Bars show mean change in metric value over 10 random seeds. Whiskers denote $\pm$1 standard deviation. Red indicates that ablation degrades performance, i.e.\ the layer is beneficial, green that ablation improves performance, i.e.\ the layer is not beneficial for zero-shot transfer.}\label{fig:ablation_examples_llama_both}
\end{figure*}

\paragraph{Results.}
Given the volume of results, we show only a selection here: the representation-ablation counterpart to \cref{fig:ablation_examples} in \cref{fig:ablation_examples_representation}, and complementary Llama 3.1 results in \cref{fig:ablation_examples_llama_both}.

\subsection{Uniform Probe Weights}\label{app:uniform-probe-weights}
\begin{figure*}[htb]
    \centering
    \begin{subfigure}{0.49\textwidth}
        \centering
        \includegraphics[width=\linewidth]{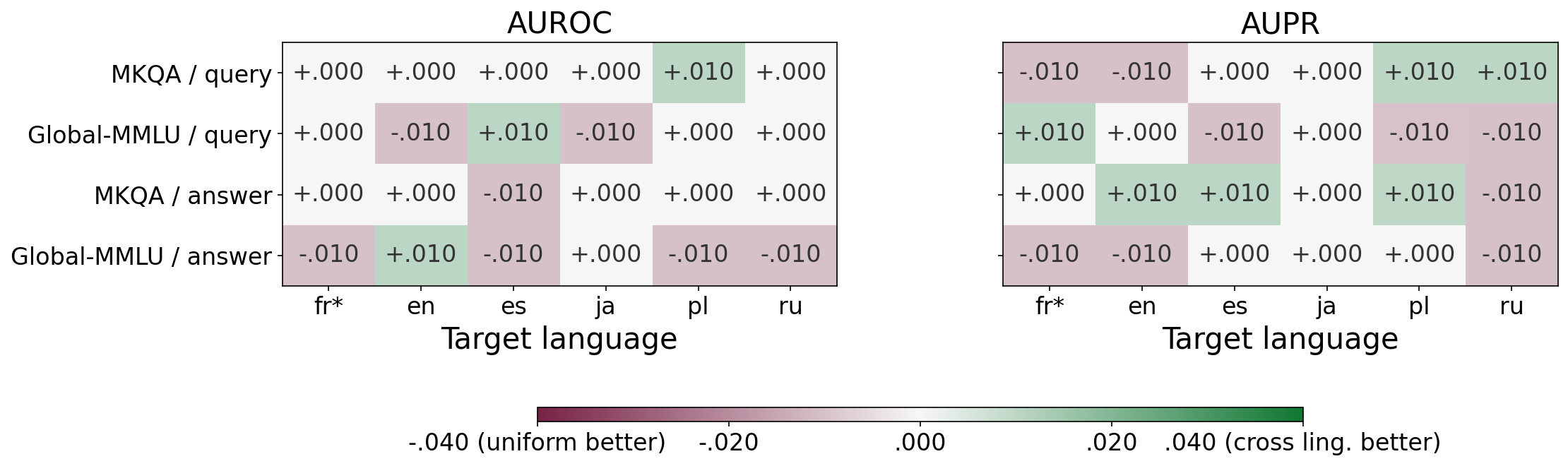}
        \includegraphics[width=\linewidth]{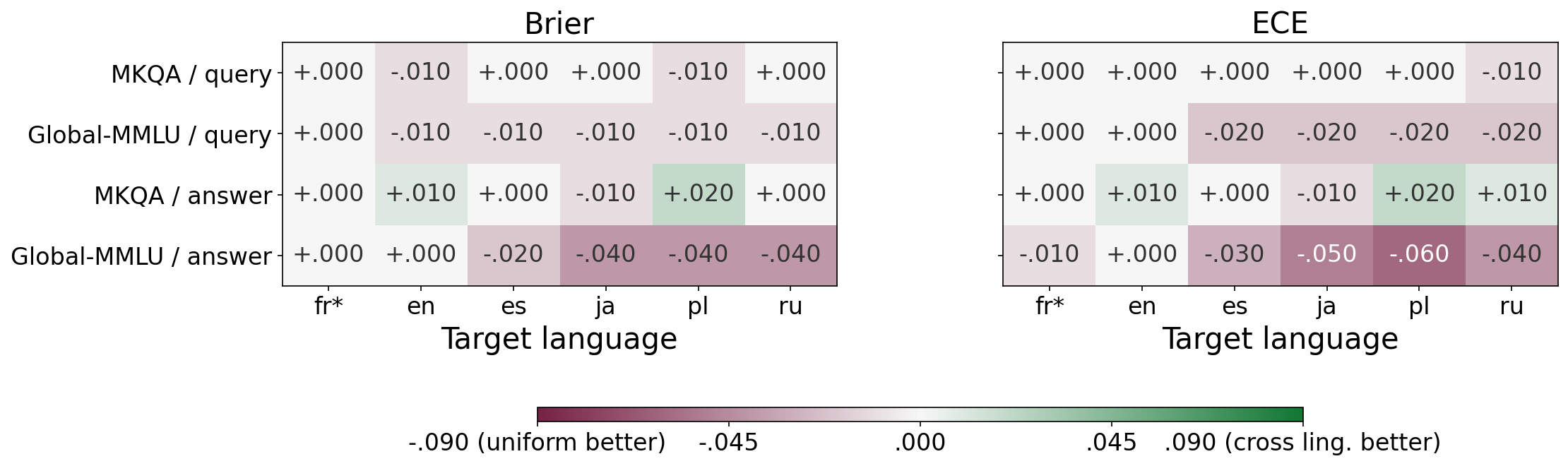}
        \caption{Qwen3 8B.}
        \label{fig:comp-uniform-qwen}
    \end{subfigure}
    \hfill
    \begin{subfigure}{0.49\textwidth}
        \centering
        \includegraphics[width=\linewidth]{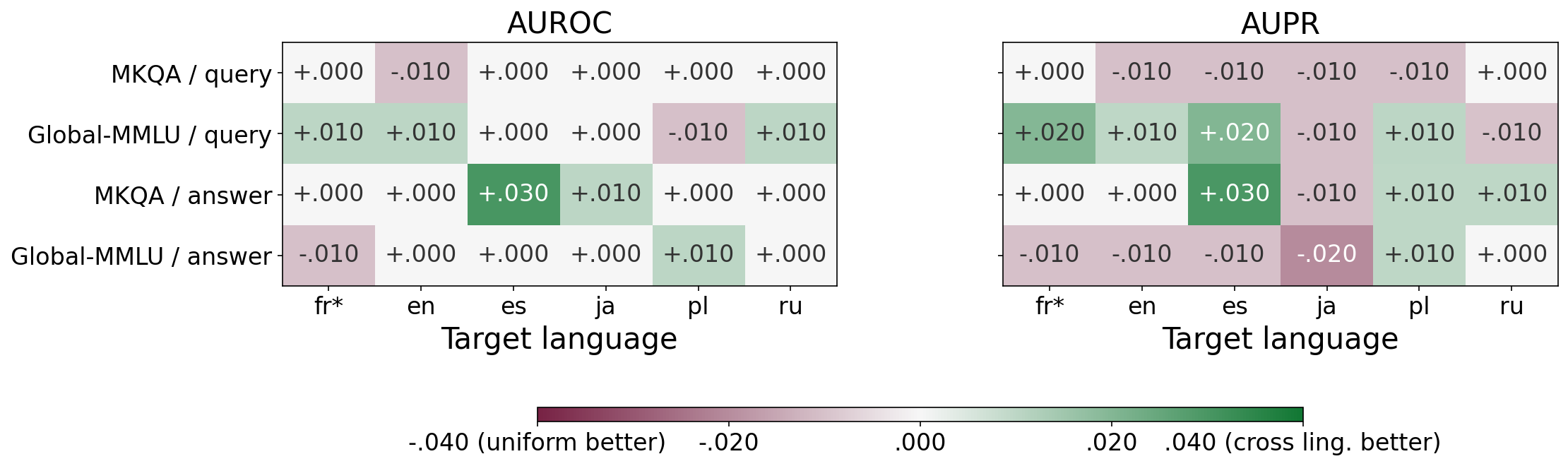}
        \includegraphics[width=\linewidth]{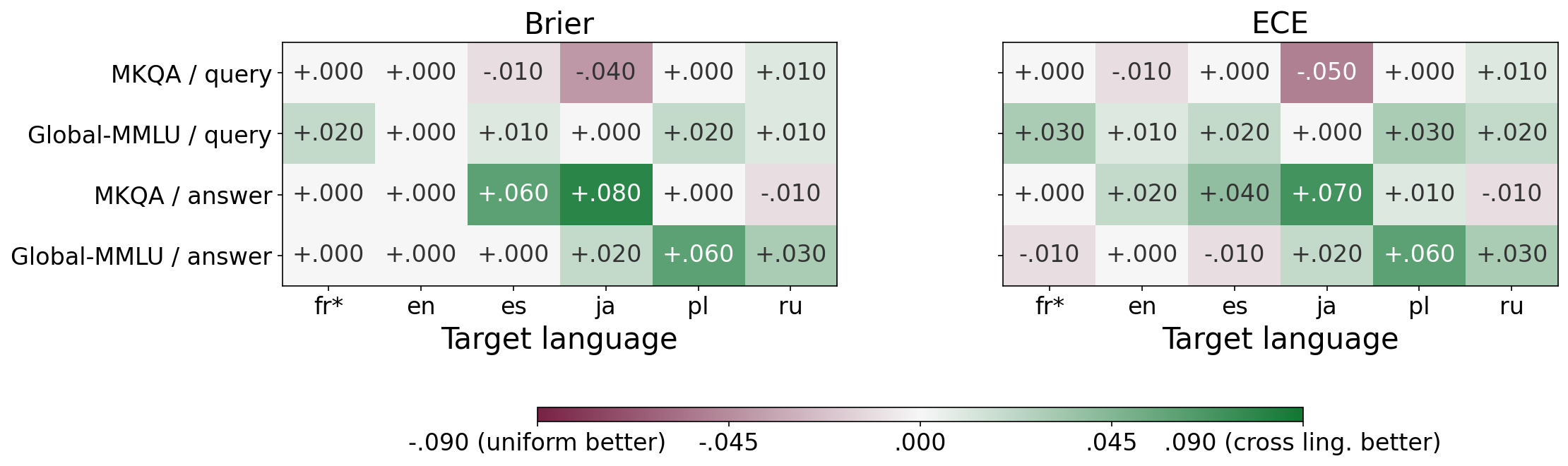}
        \caption{Llama 3.1 8B.}
        \label{fig:comp-uniform-llama}
    \end{subfigure}
    \caption{Comparison of the cross-lingual to the uniform probe (both probes trained in fr). Top row within each panel: discrimination; bottom row: calibration.}
    \label{fig:comp-uniform}
\end{figure*}

We conduct additional experiments fixing the layer weights in \cref{eq:layer_imp} to uniform and only learning the linear layer parameters. 
We show the performance differences for both models in \cref{fig:comp-uniform-qwen,fig:comp-uniform-llama}.
The comparison remains overall inconsistent, but we attribute the fact that the uniform probe is able to achieve comparable results to two factors:
First of all, the informational redundancy in transformer layers helps to preserve information in the pooled representation even when weights are uniform, and additionally the large amount of parameters in the linear layer ($4096$ given the size of hidden representations) helps to learn useful features for the task. Nevertheless, we see our results and works such as \citet{zhou2025beyond} as evidence that future, more robust works will likely rely on specific layers.



\FloatBarrier
\raggedbottom

\clearpage

\section{Reproducibility Details}

\subsection{Data Pre-Processing}
\label{sec:data_pre_appendix}

In the following, we describe the exact preprocessing used for MKQA and Global-MMLU. The final number of examples per dataset, model and language can be found in \cref{tab:total_examples}.

\begin{table} 
\centering
\renewcommand{\arraystretch}{1.4}
\resizebox{0.495\textwidth}{!}{
    \begin{tabular}{@{}llcccccc@{}}
    \toprule
    Dataset & LLM & en & es & fr & ja & pl & ru \\
    \midrule
    \multirow{2}{*}{MKQA}
    & \makecell[tl]{Llama 3.1 8B} & 5216 & 5219 & 5205 & 5176 & 5092 & 5106 \\
    & Qwen3 8B                    & 5225 & 5226 & 5222 & 5178 & 5186 & 5162 \\
    \hline
    \multirow{2}{*}{Global-MMLU}
    & \makecell[tl]{Llama 3.1 8B} & 4077 & 4148 & 4067 & 4178 & 4111 & 4111 \\
    & Qwen3 8B                    & 4212 & 4198 & 4186 & 4195 & 4120 & 4178 \\
    \bottomrule
    \end{tabular}
}
\caption{Total number of examples per dataset, model, and language.}
\label{tab:total_examples}
\end{table}

\subsubsection{MKQA}
\label{sec:data_pre_appendix_mkqa}

The MKQA dataset \cite{mkqa} (\href{https://github.com/apple/ml-mkqa}{GitHub repository}) comprises 10000 questions parallel across 26 languages, each annotated with one or more answer types (binary, date, entity, long answer, number, number with unit, short phrase, and unanswerable). We retain only questions of type date, entity, number, number with unit, and short phrase. Binary questions are excluded due to limited token generation, while long-answer and unanswerable questions are excluded because no ground-truth answers are provided in the dataset. Duplicate questions are also removed.

We further exclude questions with time-sensitive ground-truth answers that are no longer reliable. To identify such cases, we employ an LLM-as-a-judge approach using GPT-4.1 with the prompt shown in Figure \ref{fig:gpt4.1-prompt-mkqa-timesensitive}. We use default parameters, except for a temperature of 0 and a maximum of 5 generated tokens. Filtering is performed on English questions and then propagated to other languages. As a result, the final dataset contains 5237 questions per language (date: 943, entity: 3347, number: 258, number with unit: 254, short phrase: 435).

\subsubsection{Global-MMLU}
\label{sec:data_pre_appendix_gmmlu}

For our experiments, we use the test split of the Global-MMLU dataset \cite{globalmmlu}.\footnote{\url{https://huggingface.co/datasets/CohereLabs/Global-MMLU}} The dataset spans 42 languages with 14042 multiple-choice questions across 57 subjects. Following \citet{chandak-2025}, we identify and retain only those questions that are sufficiently self-contained to be posed in an open-ended format, using the following procedure applied to 
the English questions.

We first apply a RegEx-based filtering step to remove questions that are inherently tied to the multiple-choice format. Specifically, we remove questions that (1) contain keywords indicative of multiple-choice framing, such as "following", "which of these combinations", or "options", (2) contain fill-in-the-blank placeholders (e.g., three or more consecutive underscores), (3) reference "Statement 1" and "Statement 2" simultaneously, as these questions require evaluating paired assertions, or (4) ask the model to identify a true or false statement, as indicated by truth-value keywords ("true", "false", "correct", "incorrect"), unless the question also contains the word "following", in which case it is already caught by criterion (1).

We further use GPT-4.1 to score how well each question can be answered without the multiple-choice options, on a scale from 1 to 10 (see \cref{fig:gpt4.1-answerability} for the full prompt).  \cref{tab:gmmlu_gpt_score_distribution} outlines the score distribution across the questions. We retain only questions with a score of 8 or above, as these are sufficiently self-contained to be posed as open-ended questions. After these preprocessing steps we are left with 4215 questions. For all other languages, we retain the translations corresponding to these questions. Note that with this procedure we do not rephrase any question, ensuring that no additional wording bias is introduced during conversion to open-ended format.

\begin{figure*}[t]
\begin{promptbox}{P\textsubscript{1} - MKQA Time-Sensitivity Filtering Prompt (English only)}
\footnotesize
Your task is to determine whether the following question is TIME-SENSITIVE relative to when the dataset was collected (around 2019). Assume the question was asked in or before 2019.

\medskip
A question is TIME-SENSITIVE if its correct answer depends on the real-world time at which the question is asked. In particular, a question is time-sensitive if its answer depends on:
\begin{itemize}\setlength\itemsep{0pt}
  \item The current date, year, or recent events,
  \item Relative time expressions (e.g., ``this year'', ``currently'', ``latest'', ``now'', ``recent''),
  \item The current real-world status, role, ranking, statistic, population, record, or version,
  \item Ongoing or changing real-world situations.
\end{itemize}
A question is NOT time-sensitive if its answer is fixed by historical facts, completed events,
or already-released creative works (such as movies, TV series, books, or games), and does not
change depending on when the question is asked.

\medskip
\textbf{Output Instructions:} Answer with exactly one of the following:
\begin{itemize}\setlength\itemsep{0pt}
  \item YES - The question is time-sensitive (its answer would differ depending on when it is asked)
  \item NO - The question is not time-sensitive (its answer would be the same regardless of whether it is asked in 2015, 2019, or 2024)
  \item UNCERTAIN - The question cannot be reasonably judged
\end{itemize}
Do not provide any explanation.

\medskip
\textbf{Examples}

\medskip
Time-sensitive (YES): 
\begin{itemize}\setlength\itemsep{0pt}
    \item who is the prime minister of japan
    \item what is the population of india
    \item who is the ceo of apple
    \item what is the latest version of android
    \item what was the winner of last year's champions league
    \item how old is the president of france
\end{itemize}

\medskip
Not time-sensitive (NO): 
\begin{itemize}\setlength\itemsep{0pt}
    \item who wrote the novel 1984
    \item when was the berlin wall built
    \item who discovered penicillin
    \item what is the capital of argentina
    \item when does prentis come back to criminal minds
    \item who painted the sistine chapel ceiling
\end{itemize}

\medskip
Question: \verb|{question}|
\end{promptbox}
\caption{Prompt used by the LLM-as-a-judge to filter time-sensitive questions from MKQA, whose ground-truth answers may no longer be reliable. Applied to the English questions only.}
\label{fig:gpt4.1-prompt-mkqa-timesensitive}
\end{figure*}

\begin{figure*}[t]
\begin{promptbox}{P\textsubscript{2} - Answerability Scoring Prompt (English only)}
\footnotesize
Your task is to evaluate how well a multiple-choice question or incomplete statement (stem) can be answered without seeing the answer choices.

\medskip
Assign a score from 1 to 10 based only on whether the stem can be answered or completed without the choices.

\medskip
\textbf{Scoring Scale (1-10):}

\medskip
Score 1-2: The stem does not define a clear task and cannot be answered without choices.
Examples: \\
\hspace*{0.1em}\textbullet~"Oxygen is used:"\\
\hspace*{0.1em}\textbullet~"The element (4, 2) of"

\medskip
Score 3-4: The stem defines a task, but answering it requires choosing from an unspecified list or comparison that is only provided by the choices.
Examples: \\
\hspace*{0.1em}\textbullet~"Which set of integers is in order from least to greatest?"\\
\hspace*{0.1em}\textbullet~"The most accurate statement is"

\medskip
Score 5: The stem can be answered without choices, but allows many reasonable completions or answers.
Examples: \\
\hspace*{0.1em}\textbullet~"Under which circumstances would you not use a catheter valve?"\\
\hspace*{0.1em}\textbullet~"The advantages of X-rays include"

\medskip
Score 6-7: The stem can be answered without choices and limits the topic, but allows variation in the expected answer.
Examples: \\
\hspace*{0.1em}\textbullet~"What is one function of the liver?"\\
\hspace*{0.1em}\textbullet~"As temperature increases, reaction rate generally"

\medskip
Score 8-9: The stem can be answered without choices and strongly limits the expected answer.
Examples: \\
\hspace*{0.1em}\textbullet~"What gas do plants primarily absorb during photosynthesis?"\\
\hspace*{0.1em}\textbullet~"The enzymes of glycolysis are located in the"

\medskip
Score 10: The stem alone specifies a single, exact expected answer.
Examples: \\
\hspace*{0.1em}\textbullet~"What is the first law of thermodynamics in physics?" \\
\hspace*{0.1em}\textbullet~"Calculate 15 \texttimes{} 12."

\medskip
\textbf{Output Instructions:}\\
You must output exactly one line containing only one integer from 1 to 10, without any additional text.

\medskip
\textbf{Response Format (must match exactly):}\\
SCORE: X
\end{promptbox}

\caption{System prompt used by the LLM-as-a-judge to assess stem answerability on a 1-10 scale.}
\label{fig:gpt4.1-answerability}
\end{figure*}

\begin{table}[htb]
\centering
\ra{1.2}
\resizebox{0.26\textwidth}{!}{%
\begin{tabular}{ccc}
\toprule
\textbf{Score} & \textbf{Count} & \textbf{Percentage} \\
\midrule
    1  & 43   & 0.58\% \\
    2  & 50   & 0.67\% \\
    3  & 342  & 4.61\% \\
    4  & 287  & 3.87\% \\
    5  & 1171 & 15.79\% \\
    6  & 1004 & 13.54\% \\
    7  & 303  & 4.09\% \\
    8  & 2291 & 30.90\% \\
    9  & 851  & 11.48\% \\
    10 & 1073 & 14.47\% \\
\bottomrule
\end{tabular}%
}
\caption{GPT-4.1 score distribution for question answerability without multiple-choice options.}
\label{tab:gmmlu_gpt_score_distribution}
\end{table}



\subsubsection{Normalization \& Splitting}
\citet{sun-2025} showed that in transformer-based LLMs using pre-layer normalization (e.g., Llama), hidden-state variance grows exponentially with depth, hindering deeper-layer training. To address this and ensure that probe weights capture informative signal rather than magnitude differences, we apply per-layer z-score standardization with scikit-learn’s \texttt{StandardScaler}. 

For each feature $x$, we compute the standardized value $x' = (x - \mu)/\sigma$, where $\mu$ and $\sigma$ are the training-set mean and standard deviation, centering features to zero mean and unit variance. The fitted scaler is applied unchanged to the validation and test splits of the training language, and to the test sets of all zero-shot languages. We do not further normalize, as residual magnitude differences may be informative.

Given that both datasets are parallel and class ratios vary by language, we stratify splits to prevent information leakage and ensure balanced evaluation. For each training language, hidden states are divided into train/validation/test sets (0.7:0.15:0.15) while preserving correct/incorrect ratios. Evaluation samples are drawn from the test sets so that these ratios match across all languages.

\subsection{LLM Prompting}
\label{sec:llm_parameters_appendix}

The system instructions used to prompt the considered LLMs across languages are shown in \cref{fig:prompts}. \cref{tab:prompt-mapping} maps each evaluation split to its prompt template.

\begin{table}[t]
\centering
\small
\begin{tabular}{llc}
\toprule
\textbf{Dataset} & \textbf{Question Type} & \textbf{Prompt} \\
\midrule
\multirow{5}{*}{MKQA}
  & \texttt{short phrase} & $P_3$ \\
  & \texttt{entity} & $P_3$ \\
  & \texttt{date} & $P_4$ \\
  & \texttt{number} & $P_5$ \\
  & \texttt{number with unit} & $P_6$ \\
\midrule
Global MMLU & (all) & $P_7$ \\
\bottomrule
\end{tabular}
\caption{Mapping of evaluation splits to prompt templates.}
\label{tab:prompt-mapping}
\end{table}

\subsection{Answer Correctness Assessment \& Judge Validation}
\label{sec:corr_ass_appendix}

We use GPT-4.1 to assess answer correctness, with the prompt shown in \cref{fig:gpt4.1-prompt-corr-sys}. For MKQA, GPT-4.1 was used across all languages and both models. For Global-MMLU, due to resource constraints we used GPT-4.1 for French and GPT-4.1-mini for the remaining languages (for both Llama 3.1 8B and Qwen3 8B). To verify that this substitution does not compromise label quality, we measured inter-judge agreement between GPT-4.1 and GPT-4.1-mini on Global-MMLU, for Llama 3.1 8B, in French, obtaining $94.4\%$ agreement and a Cohen's $\kappa$ of $0.849$.

We have further conducted a small human validation study of judge labels. We sampled 50 question-answer pairs, stratified evenly across datasets and LLMs with a top-up of randomly sampled pairs, and used the same set of questions across all six languages. We then had the generated answers independently annotated for correctness by two native speakers per language. Specifically, for each question, we presented the annotators with the LLM's answer and the ground-truth answer from the dataset. We then asked them to evaluate whether the LLM's answer is correct according to the ground-truth. Gold labels retain items where both annotators agree (2-5 items discarded per language; $n = 279$ total).
In \cref{tab:judge_validation} we observe that human inter-annotator agreement is high in all languages (Cohen's $\kappa$ = .80-.92, raw agreement .90-.96). The judge closely matches human gold labels with 90.3\% overall agreement, .87-.92 per-language accuracy, and .75-.83 chance-corrected agreement for all languages except Japanese (.65). The Japanese gap is driven by false negatives on correct answers (recall .62, with only a single false positive).

\begin{table*}
\centering
\small
\begin{tabular}{@{}lcccccccc@{}}
\toprule
& & \multicolumn{1}{c}{Annotators} & \multicolumn{3}{c}{Judge vs.\ gold} & \multicolumn{3}{c}{Correct class} \\
\cmidrule(lr){3-3} \cmidrule(lr){4-6} \cmidrule(lr){7-9}
Language & $n$ & Cohen's $\kappa$ & Accuracy & 95\% CI & Cohen's $\kappa$ & Precision & Recall & F1 \\
\midrule
en & 48 & .92 & .92 & [.80, .97] & .83 & .95 & .86 & .90 \\
es & 45 & .80 & .91 & [.79, .96] & .81 & 1.00 & .79 & .88 \\
fr & 46 & .86 & .89 & [.77, .95] & .75 & .86 & .80 & .83 \\
ja & 47 & .85 & .87 & [.75, .94] & .65 & .89 & .62 & .73 \\
pl & 46 & .83 & .91 & [.80, .97] & .81 & 1.00 & .78 & .88 \\
ru & 47 & .87 & .91 & [.80, .97] & .80 & 1.00 & .75 & .86 \\
\bottomrule
\end{tabular}
\caption{Validation of LLM-as-a-judge correctness labels. $n$ is the number of items retained after discarding annotator disagreements (50 annotated per language). Precision, recall and F1 are for the \emph{correct} class.}
\label{tab:judge_validation}
\end{table*}

\subsection{Confidence Probe Implementation Details}\label{app:training}

The layer weights $\mathbf{w}$ were initialized to zero (uniform weighting across layers), the hidden-dimension weights $\mathbf{v}$ were sampled from the normal distribution $\mathcal{N}(0,1)$ (to break symmetry and enable diverse feature weighting), and the bias $b$ was set to zero (no offset prior to training). Probes were trained with binary cross-entropy loss with logits and $L_2$ regularization for $v$:
\begin{equation}
\label{eq:loss_function}
\displaystyle
\resizebox{0.95\columnwidth}{!}{
\ensuremath{
\begin{aligned}
& \mathcal{L}(\mathbf{y}, \mathbf{z}) =  \lambda \lVert \mathbf{v} \rVert_2^2 \\
& -\frac{1}{N} \sum_{i=1}^N \Big[
w_{+} \, y_i \log \sigma(z_i) 
+ (1 - y_i)\log\!\left(1 - \sigma(z_i)\right)
\Big]
\end{aligned}%
}
}
\end{equation}
\noindent where $N$ is the number of training examples, $y_i \in \{0, 1\}$ is the ground-truth correctness label, $z_i \in \mathbb{R}$ is the logit produced by the probe, and $\sigma(\cdot)$ is the sigmoid function. The parameter $w_{+} \in \mathbb{R}^+$ is the correct-class weight to address class imbalance, computed from the ratio of negative to positive examples in the training set.\footnote{For more information, see the \texttt{pos\_weight} parameter of PyTorch’s \texttt{BCEWithLogitsLoss} class.} $\lambda \in \mathbb{R}^+$ is the $L_2$ regularization applied only to $\mathbf{v}$, with no regularization applied to $\mathbf{w}$ or $b$.

\subsection{Hyperparameter Tuning}

We tuned hyperparameters with \texttt{Ray Tune} using the \texttt{ASHA} scheduler for inter-trial early stopping, optimizing the scalar objective \texttt{val\_loss} (minimized). Each trial sampled from a log-uniform search space: learning rate $\texttt{lr} \in [10^{-5}, 10^{-1}]$ and L2 regularization on the neuron weights $\mathbf{v}$, $\texttt{l2\_lambda} \in [10^{-4}, 3 \times 10^{-1}]$. The batch size and maximum epochs were fixed at $4$ and $1000$, respectively. During each trial, we trained with AdamW using selective weight decay (L2 applied only to $\mathbf{v}$; none for layer weights $\mathbf{w}$ or bias terms), and reported the mean training loss and validation loss at every epoch to Ray Tune. Validation loss was computed on a fixed held-out validation split. To address class imbalance, we used \texttt{BCEWithLogitsLoss} with a positive-class weight proportional to the ratio of negative to positive examples in the training set. Trials were stopped early by ASHA (\texttt{max\_t}=1000, \texttt{grace\_period}=30, \texttt{reduction\_factor}=2) when underperforming, while in-trial early stopping with patience $=20$ halted training if \texttt{val\_loss} failed to improve. After $50$ trials, Ray Tune selected the best configuration according to the minimum validation loss. The selected hyperparameters are reported in Table~\ref{tab:hyperparams}.

\begin{table}[htb]
\centering
\renewcommand{\arraystretch}{1.4}
\resizebox{0.495\textwidth}{!}{
\begin{tabular}{@{}lllccc@{}}
\toprule
Dataset & LLM & Probe & Learning rate & Epochs & \makecell{L2 regularization\\for $v$} \\
\midrule
\multirow{4}{*}{MKQA}
& \multirow{2}{*}{\makecell[tl]{Llama 3.1 8B}}
& Query  & 0.0008 & 139 & 0.2861 \\
& & Answer & 0.0005 & 210 & 0.2728 \\
& \multirow{2}{*}{Qwen3 8B}
& Query  & 0.0009 & 98  & 0.2953 \\
& & Answer & 0.0009 & 111 & 0.2225 \\
\midrule
\multirow{4}{*}{Global-MMLU}
& \multirow{2}{*}{\makecell[tl]{Llama 3.1 8B}}
& Query  & 0.0011 & 94  & 0.2826 \\
& & Answer & 0.0020 & 60  & 0.2557 \\
& \multirow{2}{*}{Qwen3 8B}
& Query  & 0.0010 & 103 & 0.2687 \\
& & Answer & 0.0010 & 165 & 0.2115 \\
\bottomrule
\end{tabular}
}
\caption{Selected hyperparameters for probes trained on last-token query and answer representations for MKQA and Global-MMLU using Llama 3.1 8B and Qwen3 8B models.}
\label{tab:hyperparams}
\end{table}

\vfill

\FloatBarrier
\clearpage

\begin{figure*}[t]

\begin{promptbox}{P\textsubscript{3} - MKQA: short phrase / entity questions (6 languages)}
\footnotesize
\begin{tabular}{@{}lp{0.92\textwidth}@{}}
\textbf{en} & You are a helpful assistant. Answer the user's question in one to ten words. End your output immediately after your answer to the question. \\[2pt]
\textbf{fr} & Tu es un assistant serviable. Réponds à la question de l'utilisateur en un à dix mots. Arrête la génération immédiatement après ta réponse à la question. \\[2pt]
\textbf{es} & Eres un asistente útil. Responde la pregunta del usuario en una a diez palabras. Detén la generación inmediatamente después de responder la pregunta. \\[2pt]
\textbf{pl} & Jesteś użytecznym asystentem. Odpowiadaj na pytania użytkownika, używając od jednego do dziesięciu słów. Przestań generować natychmiast po udzieleniu odpowiedzi. \\[2pt]
\textbf{ru} & \foreignlanguage{russian}{Ты полезный помощник. Отвечай на вопросы пользователя, используя от одного до десяти слов. Останавливай генерацию сразу после ответа. } \\[2pt]
\textbf{ja} & \begin{CJK}{UTF8}{min} あなたは役に立つアシスタントです。ユーザーの質問に1〜10語で答えてください。質問に答えたらすぐに出力を終了してください。\end{CJK}\\
\end{tabular}
\end{promptbox}

\smallskip

\begin{promptbox}{P\textsubscript{4} - MKQA: date questions (6 languages)}
\footnotesize
\begin{tabular}{@{}lp{0.92\textwidth}@{}}
\textbf{en} & You are a helpful assistant. Answer the user's question using only the date requested. Do not include explanations or extra text. End your output immediately after your answer to the question. \\[2pt]
\textbf{fr} & Tu es un assistant serviable. Réponds à la question de l'utilisateur en utilisant uniquement la date demandée. N'ajoute ni explications ni texte supplémentaire. Arrête la génération immédiatement après ta réponse à la question. \\[2pt]
\textbf{es} & Eres un asistente útil. Responde la pregunta del usuario usando solo la fecha solicitada. No incluyas explicaciones ni texto adicional. Detén la generación inmediatamente después de responder la pregunta. \\[2pt]
\textbf{pl} & Jesteś użytecznym asystentem. Odpowiadaj na pytania użytkownika, używając wyłącznie żądanej daty. Nie dodawaj wyjaśnień ani dodatkowego tekstu. Przestań generować natychmiast po udzieleniu odpowiedzi. \\[2pt]
\textbf{ru} & \foreignlanguage{russian}{Ты полезный помощник. Отвечай на вопросы пользователя, используя только указанную дату. Не добавляй пояснений или дополнительного текста. Останавливай генерацию сразу после ответа.} \\[2pt]
\textbf{ja} & \begin{CJK}{UTF8}{min} あなたは役に立つアシスタントです。ユーザーの質問には、要求された日付のみを使って回答してください。説明や余分なテキストは含めないでください。質問への回答後、すぐに出力を終了してください。\end{CJK}\\
\end{tabular}
\end{promptbox}

\smallskip

\begin{promptbox}{P\textsubscript{5} - MKQA: number questions (6 languages)}
\footnotesize
\begin{tabular}{@{}lp{0.92\textwidth}@{}}
\textbf{en} & You are a helpful assistant. Answer the user's question using only the number needed to answer it. Do not include explanations or extra text. End your output immediately after your answer to the question. \\[2pt]
\textbf{fr} & Tu es un assistant serviable. Réponds à la question de l'utilisateur en utilisant uniquement le nombre nécessaire. N'ajoute ni explications ni texte supplémentaire. Arrête la génération immédiatement après ta réponse à la question. \\[2pt]
\textbf{es} & Eres un asistente útil. Responde la pregunta del usuario usando solo el número necesario. No incluyas explicaciones ni texto adicional. Detén la generación inmediatamente después de responder la pregunta. \\[2pt]
\textbf{pl} & Jesteś użytecznym asystentem. Odpowiadaj na pytania użytkownika, używając jedynie liczby, która jest potrzebna. Nie dodawaj wyjaśnień ani dodatkowego tekstu. Przestań generować natychmiast po udzieleniu odpowiedzi. \\[2pt]
\textbf{ru} & \foreignlanguage{russian}{Ты полезный помощник. Отвечай на вопросы пользователя, используя только необходимое число. Не добавляй пояснений или дополнительного текста. Останавливай генерацию сразу после ответа.} \\[2pt]
\textbf{ja} & \begin{CJK}{UTF8}{min}あなたは役に立つアシスタントです。ユーザーの質問には、回答に必要な数字だけを使って答えてください。説明や余分なテキストは含めないでください。質問に答えたらすぐに出力を終了してください。\end{CJK} \\
\end{tabular}
\end{promptbox}

\caption*{} 
\label{fig:prompts-part1}
\end{figure*}

\begin{figure*}[t]

\begin{promptbox}{P\textsubscript{6} - MKQA: number with unit questions (6 languages)}
\footnotesize
\begin{tabular}{@{}lp{0.92\textwidth}@{}}
\textbf{en} & You are a helpful assistant. Answer the user's question using only the required number and its unit. Do not include explanations or extra text. End your output immediately after your answer to the question. \\[2pt]
\textbf{fr} & Tu es un assistant serviable. Réponds à la question de l'utilisateur en utilisant uniquement le nombre demandé et son unité. N'ajoute ni explications ni texte supplémentaire. Arrête la génération immédiatement après ta réponse à la question. \\[2pt]
\textbf{es} & Eres un asistente útil. Responde la pregunta del usuario usando solo el número requerido y su unidad. No incluyas explicaciones ni texto adicional. Detén la generación inmediatamente después de responder la pregunta. \\[2pt]
\textbf{pl} & Jesteś użytecznym asystentem. Odpowiadaj na pytania użytkownika, używając tylko wymaganej liczby i jej jednostki. Nie dodawaj wyjaśnień ani dodatkowego tekstu. Przestań generować natychmiast po udzieleniu odpowiedzi. \\[2pt]
\textbf{ru} & \foreignlanguage{russian}{Ты полезный помощник. Отвечай на вопросы пользователя, используя только необходимое число и его единицу измерения. Не добавляй пояснений или дополнительного текста. Останавливай генерацию сразу после ответа.} \\[2pt]
\textbf{ja} & \begin{CJK}{UTF8}{min}あなたは役に立つアシスタントです。ユーザーの質問には、必要な数値とその単位のみを使って答えてください。説明や余分なテキストは含めないでください。質問に答えたらすぐに出力を終了してください。\end{CJK} \\
\end{tabular}
\end{promptbox}

\smallskip

\begin{promptbox}{P\textsubscript{7} - Global MMLU (6 languages)}
\footnotesize
\begin{tabular}{@{}lp{0.92\textwidth}@{}}
\textbf{en} & You are a helpful assistant. Answer the user's question in one to ten words. If the answer is a number, respond with only that number. Do not include explanations or extra text. End your output immediately after your answer to the question. \\[2pt]
\textbf{fr} & Tu es un assistant serviable. Réponds à la question de l'utilisateur en un à dix mots. Si la réponse est un nombre, indique uniquement ce nombre. N'ajoute ni explications ni texte supplémentaire. Arrête la génération immédiatement après ta réponse à la question. \\[2pt]
\textbf{es} & Eres un asistente útil. Responde la pregunta del usuario en una a diez palabras. Si la respuesta es un número, responde solo con ese número. No incluyas explicaciones ni texto adicional. Detén la generación inmediatamente después de responder la pregunta. \\[2pt]
\textbf{pl} & Jesteś asystentem pomocniczym. Odpowiadaj na pytania użytkownika, używając od jednego do dziesięciu słów. Jeśli odpowiedź jest liczbą, podaj tylko tę liczbę. Nie dodawaj wyjaśnień ani dodatkowego tekstu. Przestań generować natychmiast po udzieleniu odpowiedzi. \\[2pt]
\textbf{ru} & \foreignlanguage{russian}{Ты полезный помощник. Отвечай на вопросы пользователя, используя от одного до десяти слов. Если ответ число, укажи только это число. Не добавляй пояснений или дополнительного текста. Останавливай генерацию сразу после ответа. } \\[2pt]
\textbf{ja} & \begin{CJK}{UTF8}{min}あなたは役に立つアシスタントです。ユーザーの質問に1〜10語で答えてください。答えが数字の場合は、その数字のみを答えてください。説明や余分なテキストは含めないでください。質問に答えたらすぐに応答を終了してください。\end{CJK} \\
\end{tabular}
\end{promptbox}

\caption{System prompts used for answer extraction, organized by dataset and by question type (P\textsubscript{3}-P\textsubscript{7}). All prompts cover six languages: English (en), French (fr), Spanish (es), Polish (pl), Russian (ru), and Japanese (ja).}
\label{fig:prompts}
\end{figure*}

\begin{figure*}[t]
\begin{promptbox}{P\textsubscript{8} - Correctness Assessment System Prompts (6 languages)}
\footnotesize
\begin{tabular}{@{}lp{0.95\textwidth}@{}}
\textbf{en} & You are an expert answer evaluator assessing whether a predicted answer conveys the key information as any of the correct answers, regardless of phrasing. Reply with only YES or NO. \\[2pt]
\textbf{fr} & Tu es un expert en évaluation des réponses, chargé de déterminer si une réponse proposée transmet les informations clés au même titre que n'importe quelle réponse correcte, indépendamment de sa formulation. Réponds par OUI ou NON uniquement. \\[2pt]
\textbf{es} & Usted es un evaluador experto de respuestas que debe determinar si una respuesta predicha transmite la misma información clave que alguna de las respuestas correctas, independientemente de la redacción. Responda solo con SÍ o NO. \\[2pt]
\textbf{pl} & Jesteś ekspertem w ocenie odpowiedzi i oceniasz, czy przewidywana odpowiedź przekazuje kluczowe informacje, jako którakolwiek z poprawnych odpowiedzi, niezależnie od sformułowania. Odpowiadaj tylko TAK lub NIE. \\[2pt]
\textbf{ru} & \foreignlanguage{russian}{Вы являетесь экспертом по оценке ответов. Определите, передает ли предсказанный ответ ключевую информацию так же, как и любой из правильных ответов, независимо от формулировки. Отвечайте только «ДА» или «НЕТ».} \\[2pt]
\textbf{ja} & \begin{CJK}{UTF8}{min}あなたは専門の回答評価者です。予測された回答が、表現方法に関わらず、いずれかの正解と同等に重要な情報を伝えられているかを評価しています。「はい」または「いいえ」のみで回答してください。\end{CJK} \\
\end{tabular}
\end{promptbox}
\caption{System prompt for LLM judge used to assess answer correctness. 
All prompts cover 6 languages: English (en), French (fr), Spanish (es), Polish (pl), Russian (ru), and Japanese (ja).}
\label{fig:gpt4.1-prompt-corr-sys}
\end{figure*}
\begin{figure*}[t]
\begin{promptbox}{P\textsubscript{9} - P(True) Prompts (6 languages)}
\footnotesize
\begin{tabular}{@{}lp{0.95\textwidth}@{}}
\textbf{en} & Question: \{question\}\newline
            Possible answer: \{answer\}\newline
            Is the possible answer:\newline
            (A) True\newline
            (B) False\newline
            You must respond with only (A) or (B). The possible answer is:\\[2pt]
\textbf{fr} & Question : \{question\}\newline
            Réponse possible : \{answer\}\newline
            La réponse possible est-elle :\newline
            (A) Vrai\newline
            (B) Faux\newline
            Vous devez répondre uniquement par (A) ou (B). La réponse possible est :\\[2pt]
\textbf{es} & Pregunta: \{question\}\newline
            Posible respuesta: \{answer\}\newline
            La posible respuesta es:\newline
            (A) Verdadero\newline
            (B) Falso\newline
            Debes responder únicamente con (A) o (B). La posible respuesta es:\\[2pt]
\textbf{pl} & Pytanie: \{question\}\newline
            Możliwa odpowiedź: \{answer\}\newline
            Czy możliwa odpowiedź to:\newline
            (A) Prawda\newline
            (B) Fałsz\newline
            Musisz odpowiedzieć wyłącznie (A) lub (B). Możliwa odpowiedź to: \\[2pt]
\textbf{ru} & \foreignlanguage{russian}{Вопрос: \{question\}\newline
            Возможный ответ: \{answer\}\newline
            Верен ли возможный ответ:\newline
            (A) Верно\newline
            (B) Неверно\newline
            Необходимо ответить только (A) или (B). Возможный ответ верен::} \\[2pt]
\textbf{ja} & \begin{CJK}{UTF8}{min}質問: \{question\}\newline
            考えられる回答: \{answer\}\newline
            考えられる回答は:\newline
            (A) 真\newline
            (B) 偽\newline
            (A)または(B)のみで答えてください。考えられる回答は:\end{CJK} \\
\end{tabular}
\end{promptbox}
\caption{Prompts used for P(True) \citep{kadavath2022language}. 
All prompts cover 6 languages: English (en), French (fr), Spanish (es), Polish (pl), Russian (ru), and Japanese (ja).}
\label{fig:ptrue-prompts}
\end{figure*}
\begin{figure*}[t]
\begin{promptbox}{P\textsubscript{10} - Verbalized Confidence Prompts (6 languages)}
\footnotesize
\begin{tabular}{@{}lp{0.95\textwidth}@{}}
\textbf{en} & Considering the question and your provided answer, estimate the probability the answer is correct as a percentage from 0 to 100, where 0 = certainly incorrect and 100 = absolutely correct. 
Return only Confidence: <0-100>.
\\[2pt]
\textbf{fr} & Compte tenu de la question et de votre réponse, estimez la probabilité que la réponse soit correcte, en pourcentage, entre 0 et 100, où 0 signifie « certainement incorrect » et 100 « absolument correct ».
Indiquez uniquement le niveau de confiance : <0-100>. \\[2pt]
\textbf{es} & Considerando la pregunta y la respuesta proporcionada, estime la probabilidad de que la respuesta sea correcta en forma de porcentaje de 0 a 100, donde 0 = definitivamente incorrecta y 100 = absolutamente correcta.
Devuelva solo el nivel de confianza: <0-100>. \\[2pt]
\textbf{pl} & Biorąc pod uwagę pytanie i udzieloną odpowiedź, oszacuj prawdopodobieństwo, że odpowiedź jest poprawna, w procentach od 0 do 100, gdzie 0 oznacza odpowiedź zdecydowanie błędną, a 100 – odpowiedź całkowicie poprawną.
Zwróć tylko poziom pewności: <0-100>. \\[2pt]
\textbf{ru} & \foreignlanguage{russian}{Учитывая вопрос и предоставленный вами ответ, оцените вероятность правильности ответа в процентах от 0 до 100, где 0 = заведомо неверный ответ, а 100 = абсолютно верный ответ.
Верните только Уверенность: <0-100>.} \\[2pt]
\textbf{ja} & \begin{CJK}{UTF8}{min}質問とあなたの回答を考慮し、回答が正しい確率を0～100の範囲で推定してください。0は確実に間違っている、100は絶対に正しいとします。
信頼度 <0-100> のみを返してください。 \end{CJK} \\
\end{tabular}
\end{promptbox}
\caption{Prompts to ask for verbalized confidence. 
All prompts cover 6 languages: English (en), French (fr), Spanish (es), Polish (pl), Russian (ru), and Japanese (ja).}
\label{fig:verbalized-confidence-prompts}
\end{figure*}

\end{document}